\documentclass[10pt,twocolumn,letterpaper]{article}

\usepackage{iccv}
\usepackage{times}
\usepackage{epsfig}
\usepackage{graphicx}
\usepackage{amsmath}
\usepackage{amssymb}
\usepackage{bm}
\usepackage{booktabs}
\usepackage{multirow}
\usepackage{rotating}
\usepackage{xcolor}
\usepackage{colortbl}
\usepackage{array}
\usepackage{caption}
\usepackage{subcaption} 
\usepackage{setspace}
\usepackage{soul}
\usepackage{cite} 
\usepackage{graphicx} 
\usepackage{stfloats} 
\usepackage{mathtools} 
\usepackage{eso-pic} 
\usepackage{pifont} 
\usepackage[ruled,vlined,,linesnumbered]{algorithm2e}
\usepackage{dsfont}
\usepackage{balance}
\usepackage{float}
\usepackage{changepage}

\definecolor{iccvblue}{rgb}{0.21,0.49,0.74}
\usepackage[pagebackref=true,breaklinks=true,colorlinks,bookmarks=false, allcolors=iccvblue]{hyperref}
\usepackage{cleveref} 

\Crefname{section}{Sec.}{Secs.} 
\Crefname{figure}{Fig.}{Figs.} 
\Crefname{table}{Table}{Tables} 
\crefname{equation}{}{}
\Crefname{equation}{}{}

\iccvfinalcopy 

\def\iccvPaperID{10698} 

\ificcvfinal\fi

\newcommand{\bdmath}{\begin{dmath}}
\newcommand{\edmath}{\end{dmath}}
\newcommand{\beq}{\begin{equation}}
\newcommand{\eeq}{\end{equation}}
\newcommand{\bdm}{\begin{displaymath}}
\newcommand{\edm}{\end{displaymath}}
\newcommand{\bea}{\begin{eqnarray}}
\newcommand{\eea}{\end{eqnarray}}
\newcommand{\beal}{\beq \begin{array}{ll}}
\newcommand{\eeal}{\end{array} \eeq}
\newcommand{\beas}{\begin{eqnarray*}}
\newcommand{\eeas}{\end{eqnarray*}}
\newcommand{\ba}{\begin{array}}
\newcommand{\ea}{\end{array}}
\newcommand{\bit}{\begin{itemize}}
\newcommand{\eit}{\end{itemize}}
\newcommand{\ben}{\begin{enumerate}}
\newcommand{\een}{\end{enumerate}}

\newcommand{\calS}{{\cal S}}

\newcommand{\setal}{~\emph{et~al.}\xspace}
\renewcommand{\eg}{\emph{e.g.,}\xspace}
\renewcommand{\ie}{\emph{i.e.,}\xspace}

\def\etalcite#1{\setal~\cite{#1}}

\newcommand{\rom}[1]{\uppercase\expandafter{\romannumeral #1\relax}}

\newcommand{\M}[1]{{\bm #1}} 
\renewcommand{\boldsymbol}[1]{{\bm #1}}

\newcommand{\hide}[1]{}

\newcommand{\hiddenText}{{\color{gray} hidden text.}}
\newcommand{\hideWithText}[1]{\hiddenText}

\DeclareMathOperator*{\argmin}{arg\,min}

\newcommand{\twonorm}[1]{\|#1\|_{2}}

\newcommand{\SOthree}{\ensuremath{\mathrm{SO}(3)}\xspace}

\newcommand{\MC}{\M{C}}

\newcommand{\MR}{\M{R}}

\newcommand{\MI}{\M{I}}
\newcommand{\MV}{\M{V}}

\newcommand{\vc}{\boldsymbol{c}}

\newcommand{\vn}{\boldsymbol{n}}

\newcommand{\vp}{\boldsymbol{p}}
\newcommand{\vq}{\boldsymbol{q}}

\newcommand{\vv}{\boldsymbol{v}}
\newcommand{\vt}{\boldsymbol{t}}

\newcommand{\blue}[1]{{\color{blue}#1}}

\newcommand{\linkToPdf}[1]{\href{#1}{\blue{(pdf)}}}
\newcommand{\linkToPpt}[1]{\href{#1}{\blue{(ppt)}}}
\newcommand{\linkToCode}[1]{\href{#1}{\blue{(code)}}}
\newcommand{\linkToWeb}[1]{\href{#1}{\blue{(web)}}}
\newcommand{\linkToVideo}[1]{\href{#1}{\blue{(video)}}}
\newcommand{\linkToMedia}[1]{\href{#1}{\blue{(media)}}}
\newcommand{\award}[1]{\xspace} 

\newcommand{\vz}{\boldsymbol{z}}

\newcommand{\oursname}{{BUFFER-X}}

\newcommand{\ThreeDMatch}{\texttt{3DMatch}}
\newcommand{\ThreeDLoMatch}{\texttt{3DLoMatch}}
\newcommand{\ScanNetppi}{\texttt{ScanNet++i}}
\newcommand{\ScanNetppF}{\texttt{ScanNet++F}}
\newcommand{\TIERS}{\texttt{TIERS}}
\newcommand{\KITTI}{\texttt{KITTI}}
\newcommand{\ETH}{\texttt{ETH}}
\newcommand{\MIT}{\texttt{MIT}}
\newcommand{\KAIST}{\texttt{KAIST}}
\newcommand{\Oxford}{\texttt{Oxford}}
\newcommand{\WOD}{\texttt{WOD}}

\newcommand{\omitted}[1]{}

\newcommand{\bmat}{\left[ \begin{array}}
\newcommand{\emat}{\end{array}\right]}

\newcommand{\subMeas}[1]{\calS} %

\begin{document}



\title{\oursname: Towards Zero-Shot Point Cloud Registration in Diverse Scenes}


\author{
  Minkyun Seo$^{1}$\thanks{\noindent These authors contributed equally to this work. \hfill \break \indent \, $^\dagger$Corresponding author.}\qquad Hyungtae Lim$^{3*}$\qquad Kanghee Lee$^1$ \qquad Luca Carlone$^3$ \qquad  Jaesik Park$^{1, 2\dagger}$\\
  \normalsize{$^1$Computer Science Engineering and $^2$Interdisciplinary Program of AI, Seoul National University}\\
  \normalsize{$^3$Laboratory for Information \& Decision Systems, Massachusetts Institute of Technology}\\
{\tt\small minkyunseo00@gmail.com, shapelim@mit.edu, kanghee.lee@snu.ac.kr,}\\
{\tt\small lcarlone@mit.edu, jaesik.park@snu.ac.kr}
}

\maketitle
\ificcvfinal\thispagestyle{empty}\fi

\begin{abstract}

Recent advances in deep learning-based point cloud registration have improved generalization, yet most methods still require retraining or manual parameter tuning for each new environment.
In this paper, we identify three key factors limiting generalization:
(a)~reliance on environment-specific voxel size and search radius,
(b)~poor out-of-domain robustness of learning-based keypoint detectors,
and (c)~raw coordinate usage, which exacerbates scale discrepancies.
To address these issues, we present a zero-shot registration pipeline called \textit{BUFFER-X} by (a)~adaptively determining voxel size/search radii, (b)~using farthest point sampling to bypass learned detectors, and (c)~leveraging patch-wise scale normalization for consistent coordinate bounds.
In particular, we present a multi-scale patch-based descriptor generation and a hierarchical inlier search across scales to improve robustness in diverse scenes.
We also propose a novel generalizability benchmark using 11 datasets that cover various indoor/outdoor scenarios and sensor modalities, demonstrating that BUFFER-X achieves substantial generalization without prior information or manual parameter tuning for the test datasets.
Our code is available at \href{https://github.com/MIT-SPARK/BUFFER-X}{\texttt{https://github.com/MIT-SPARK/BUFFER-X}}.
\end{abstract}

\section{Introduction}

The field of deep learning-based point cloud registration has made steady and remarkable progress, including enhancing feature distinctiveness~\cite{Ao20ietcv-SGHs,Ao20pr-Repeatable,Ao23CVPR-BUFFER,Ao21cvpr-Spinnet,Chen24cpvr-DCATr,Mu24cvpr-ColorPCR, Huang21cvpr-PREDATORRegistration,Chen24icra-Tree-based-Transformer}, improving data association strategies~\cite{Huang21cvpr-PREDATORRegistration,Zhang24cvpr-FastMAC, Zhang23cvpr-MaxCliques,Fathian24ral-Clipperplus,Liu24tgrs-DeepSemanticMatching,Liu24tpami-NCMNet,Yuan24cvpr-InlierConfidence,Liu24cvpr-Extend}, and developing more robust pose estimation solvers~\cite{Shi24ral-RANSAC,Huang24cvpr-Scalable,Yang24tpami-MAC,Zhou16eccv-FastGlobalRegistration}. 
Consequently, existing approaches achieve strong performance on test sequences within the same dataset used for training, successfully estimating the relative pose between two partially overlapping point clouds~\cite{Yang20tro-teaser, Lim24ijrr-Quatropp, Lim22icra-Quatro, Yang16pami-goicp, Bernreiter21ral-PHASER, Yin23icra-Segregator}.

More recently, there has been growing interest in tackling the generalization of these deep learning-based methods~\cite{Ao23CVPR-BUFFER,Ao21cvpr-Spinnet,Poiesi22pami-GeDi,Dosovitskiy19iclr-YOTO,Chen21nips-OnlyTrainOnce}, which is the capability of a network to perform well across diverse real-world scenarios.

While these approaches have demonstrated excellent generalization performance, in practice, most existing methods still require the user to provide optimal parameters, such
as voxel size for downsampling cloud points and search radius for descriptor generation, when dealing with unseen domain datasets.
In this paper, we refer to this manual tuning as an \textit{oracle}.
Therefore, it is still desirable to develop zero-shot registration approaches for better usability and practical deployment.

\begin{figure}[t!]
 	\centering
 	\begin{subfigure}[b]{0.45\textwidth}
 		\includegraphics[width=1.0\textwidth]{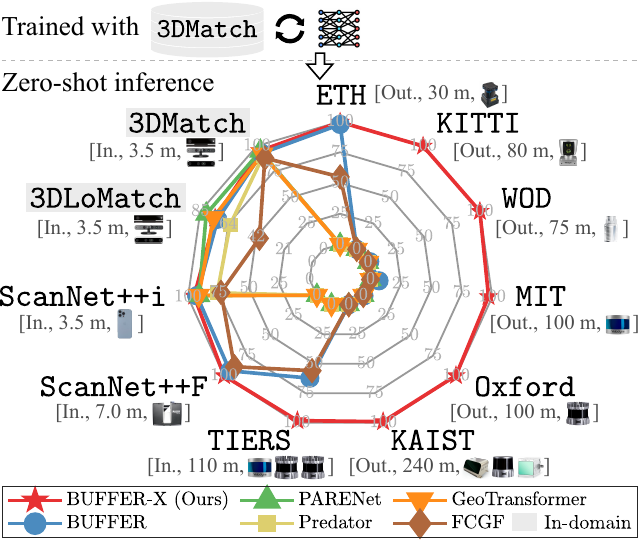}
 	\end{subfigure}
 	\captionsetup{font=small}
  \caption{Success rate (unit: \%) of \textit{zero-shot point cloud registration} with state-of-the-art approaches on 11 datasets~\cite{Geiger13ijrr-KITTI,Zeng17cvpr-3dmatch,Jung23ijrr-HeLiPR,Huang21cvpr-PREDATORRegistration,Yeshwanth23iccv-Scannet++,Qingqing22iros-TIERS,Ramezani20iros-NewerCollege,Sun20cvpr-WaymoDataset, Tian23iros-KimeraMultiExperiments,Pomerleau12ijrr-ETH}. Without any prior information or manual parameter tuning for the test datasets, our \textit{BUFFER-X} shows robust generalization capability across diverse scenes even though the network is only trained on the {\ThreeDMatch} dataset~\cite{Zeng17cvpr-3dmatch}.} 
    \label{fig:fig1}
\end{figure}

Furthermore, despite nearly a decade of research in deep learning-based registration, most studies remain confined to specific scenarios, 
primarily conducting experiments using omnidirectional LiDAR point clouds for outdoor environments~\cite{Yew18eccv-3dfeatnet, Geiger13ijrr-KITTI} and RGB-D depth clouds for indoor settings~\cite{Zeng17cvpr-3dmatch}.
For this reason, domain generalization experiments on LiDAR point clouds in indoors~\cite{Qingqing22iros-TIERS} or with different LiDAR scan patterns in outdoor environments~\cite{Qingqing22iros-TIERS,Jung23ijrr-HeLiPR,Lim24iros-HeLiMOS} are less explored.
This underscores the need for a new benchmark that better reflects real-world sensor variations to evaluate generalizability across unseen environments and diverse scanning patterns.

In this context, the main contribution of this paper is addressing two key issues above and propose: a)~a \textit{zero-shot registration} architecture and b)~a novel benchmark to help evaluate the generalization capability of deep learning-based registration approaches, as shown in \Cref{fig:fig1}.
First, inspired by the remarkable generalization of BUFFER~\cite{Ao23CVPR-BUFFER} as long as a user manually tunes the voxel size and search radius,
we first thoroughly analyze the architectural principles that underpin its generalization. 
Then, we identify three factors that hinder the zero-shot capabilities of existing methods in \Cref{sec:our_preliminaries}.
Building on these insights, we introduce a self-adaptive mechanism to determine the optimal voxel size for each test scene and streamline the pipeline of BUFFER, ultimately presenting a robust multi-scale patch-wise approach. 
We name our approach \textit{\oursname} to signify that it is an extension of BUFFER.

Second, we establish a comprehensive benchmark that encompasses both indoor and outdoor settings,
ensuring that outdoor settings include culturally and geographically diverse locations~(\eg captured in Europe, Asia, and the USA), various environmental scales (ranging from meters to kilometers),
and different LiDAR scanning patterns, while indoor settings also incorporate LiDAR-captured data. 
Subsequently, we demonstrate that our method achieves promising generalization capability without any prior information or manual parameter tuning during the evaluation; see \Cref{fig:fig1}.

In summary, we make three key claims: 
(i)~we thoroughly analyze the limitations of existing approaches and identify the key factors that have hindered zero-shot generalization,
(ii)~we present an improved approach, named {\oursname}, that addresses the generalization issues of state-of-the-art methods,
and (iii)~we introduce a benchmark to evaluate zero-shot generalization performance comprehensively. 

\section{Related Work}
\label{sec:related}

3D point cloud registration, which estimates the relative pose between two partially overlapping point clouds, is a fundamental problem in the fields of robotics and computer vision~\cite{Aoki24icra-3DBBS, Yin24ijcv-LiDARLocSurvey, Chen22ar-Overlapnet, Cattaneo22tro-LCDNet, Lee22arxiv-LearningReg}.
Overall, point cloud registration methods are classified into two categories based on whether their performance relies on the availability of an initial guess for registration: a)~\emph{local} registration~\cite{Besl92pami, Segal09rss-GeneralizedICP,Pomerleau13auro-ICPcomparison,Koide21icra-VGICP,Oomerleau12ijrr-ethpc,Vizzo23ral-KISSICP} and b)~\emph{global} registration~\cite{Fischler81,Dong17isprsremotesensing-GHICP,Yang16pami-goicp,Zhou16eccv-FastGlobalRegistration,Bernreiter21ral-PHASER,Yang19rss-teaser,Yang20tro-teaser,Lim22icra-Quatro,Lim24ijrr-Quatropp}.
Global registration methods can be further classified into two types: a)~\emph{correspondence-free}~\cite{Rouhani11iccv-CorrespondenceFreeReg,Brown19pr-AFamiliyofBnB,Bernreiter21ral-PHASER, Papazov12ijrr-Rigid3DGeometryMatching,Chum03jprs-LocallyOptimizedRANSAC,Choi97jcv-RANSAC,Schnabel07cgf-EfficientRANSAC,Olsson09pami-bnbRegistration, Hartley09ijcv-globalRotationRegistration, Pan19robotbiomim-MultiViewBnB,Zhao24cvpr-CorrespondenceFree} and b)~\emph{correspondence-based}~\cite{Fischler81,Yang16pami-goicp,Zhou16eccv-FastGlobalRegistration,Dong17isprsremotesensing-GHICP,Lei17tip-FastDescriptors,Yuan24cvpr-InlierConfidence,Liu24cvpr-Extend,Yu24cvpr-InstanceAware} approaches.
In this study, we focus on the latter and particularly place more emphasis on deep learning-based registration methods.

Since Qi\etalcite{Qi17cvpr-pointnet} demonstrated that learning-based techniques in 2D images can also be applied to 3D point clouds, a wide range of learning-based point cloud registration approaches have been proposed.  
Building on these advances, novel network architectures with increased capacity have continuously emerged, ranging from MinkUNet~\cite{Choi19iccv-FCGF,Choy20cvpr-deepGlobalRegistration,choy19cvpr-4DSTConv}, cylindrical convolutional network~\cite{Ao21cvpr-Spinnet,Ao23CVPR-BUFFER,Zhu21cvpr-Cylindrical3D}, KPConv~\cite{Huang21cvpr-PREDATORRegistration,Thomas19iccv-kpconv, Bai20cvpr-D3Feat} to Transformers~\cite{Qin23tpami-GeoTransformer,Wu22nips-PointTransformerV2,Wu24cvpr-PointTransformerV3,Chen24cpvr-DCATr}.

While these advancements have led to improved registration performance, some of these methods often exhibit limited generalization capability,
leading to performance degradation when applied to point clouds captured by different sensor configurations or in unseen environments. 
To tackle the generalization problem, Ao\etalcite{Ao21cvpr-Spinnet, ao22tpami-YOTO} introduced SpinNet, a patch-based method that normalizes the range of local point coordinates within a fixed-radius neighborhood to $[-1, 1]$.
This makes us come to realize that patch-wise scale normalization is key to achieving a data-agnostic registration pipeline.

Further, Ao\etalcite{Ao23CVPR-BUFFER} proposed BUFFER to enhance efficiency by combining point-wise feature extraction with patch-wise descriptor generation.
However, we found that such learning-based keypoint detectors can hinder robust generalization, as their failure in out-of-domain distributions may trigger a cascading failure in subsequent steps; see \Cref{sec:probpt_detector}.
In addition, despite the high generalizability of BUFFER,
we observed that during cross-domain testing, users had to manually specify the optimal voxel size for the test data, which hinders fully zero-shot inference.

Under these circumstances, we revisit the generalization problem in point cloud registration and explore how to achieve zero-shot registration while preserving the key benefits of BUFFER’s scale normalization strategy.
In addition, we remove certain modules that hinder robustness and introduce an adaptive mechanism that determines the voxel size and search radii depending on the given cloud points pair. 
To the best of our knowledge, this is the first approach to evaluate the zero-shot generalization across diverse scenes covering various environments, geographic regions, scales, sensor types, and acquisition setups.

\section{Preliminaries}\label{sec:our_preliminaries}

\subsection{Problem statement}\label{sec:problem}
\newcommand{\corr}{\mathcal{A}}
\newcommand{\estoutliers}{\hat{\mathcal{O}}}
\newcommand{\srcpt}{\srcpoint_\srcidx}
\newcommand{\tgtpt}{\tgtpoint_\tgtidx}
\newcommand{\voxel}{v}
\newcommand{\voxelfunc}[1]{f_\voxel({#1})}
\newcommand\srcpoint{\vp}
\newcommand\tgtpoint{\vq}
\newcommand\srcidx{i}
\newcommand\tgtidx{j}
\newcommand{\kth}{k}
\newcommand{\lth}{l}
\newcommand\srccloud{\mathcal{P}}
\newcommand\tgtcloud{\mathcal{Q}}

\newcommand{\search}{\mathcal{S}}
\newcommand{\Pqvalid}{\mathcal{P}_\text{valid}}
\newcommand{\normalvec}{\mathbf{n}}
\newcommand{\plin}{p_\text{lin}}
\newcommand{\numthr}{\tau_\text{num}}
\newcommand{\neighboring}{k}
\newcommand{\neighboringIdxSet}{\mathcal{I}}
\newcommand{\neighboringValidIdxSet}{\mathcal{I}_\text{valid}}
\newcommand{\searchfpfh}{\search_{\rfpfh}}
\newcommand{\validQueryIndices}{\mathcal{J}}

The goal of point cloud registration is to estimate the relative 3D rotation matrix $\MR \in \SOthree$ and translation vector $\vt \in \mathbb{R}^{3}$ between two unordered 3D point clouds $\mathcal{P}$ and $\mathcal{Q}$.
To this end, most approaches~\cite{Huang21cvpr-PREDATORRegistration, Lim25icra-KISSMatcher} follow three steps:
a)~apply voxel sampling $\voxelfunc{\cdot}$ to the point cloud with voxel size $\voxel$ as preprocessing,
b)~establish associations (or \textit{correspondences})~$\corr$,
and c)~estimate $\MR$ and $\vt$. 

Formally, by denoting corresponding points for a correspondence $(i, j)$ in $\mathcal{A}$ as $\vp_i \in \voxelfunc{\mathcal{P}}$ and  $\vq_j \in \voxelfunc{\mathcal{Q}}$, respectively,
the objective function used for pose estimation can be defined as:
\begin{equation}
  \hat{\MR}, \hat{\vt} =\argmin_{\MR \in \mathrm{SO}(3), \vt \in \mathbb{R}^{3}}  \sum_{(i,j) \in  \corr} \rho\Big( \twonorm{\tgtpoint_\tgtidx -\MR \srcpoint_\srcidx -\vt} \Big),
 \label{eq:final_goal}
\end{equation}
\noindent where $\rho(\cdot)$ represents a nonlinear kernel function that mitigates the effect of spurious correspondences in $\corr$.

\begin{figure}[t!]
    \centering
    \begin{subfigure}[b]{.23\textwidth}
        \includegraphics[width=1\textwidth]{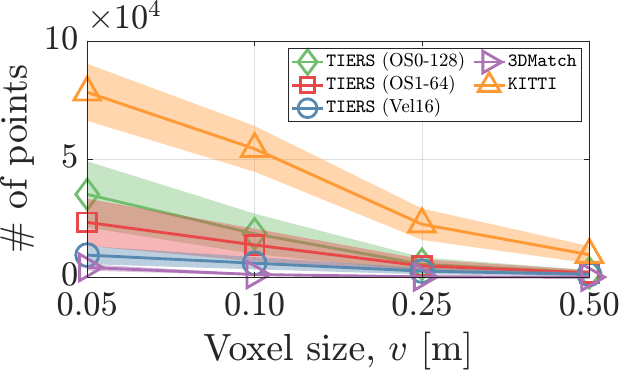}
        \caption{}
    \end{subfigure}
    \begin{subfigure}[b]{.215\textwidth}
        \includegraphics[width=1\textwidth]{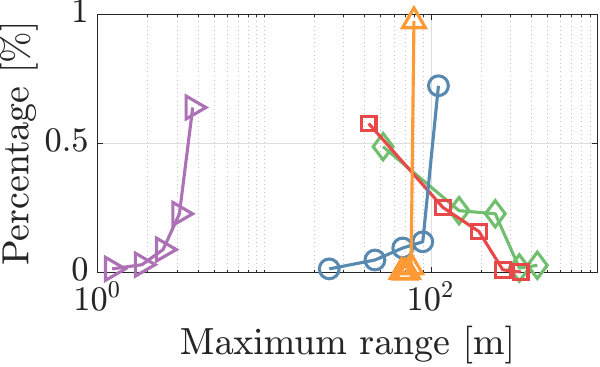}
        \caption{}
    \end{subfigure}
    \captionsetup{font=small}
    \vspace{-2mm}
    \caption{
    (a) Variation in the number of points after voxelization with different voxel sizes $v$ across datasets.
    Even in indoor scenes, point counts vary significantly depending on the sensor type~(\ie {\TIERS}~\cite{Qingqing22iros-TIERS} vs. {\ThreeDMatch}~\cite{Zeng17cvpr-3dmatch}).
    Notably, {\TIERS} and {\KITTI}~\cite{Geiger13ijrr-KITTI}, both using omnidirectional LiDARs, yield different point densities due to indoor vs. outdoor environments.
    (b) Empirical distribution of the datasets’ maximum range.}
    \label{fig:key_elements}
    \vspace{-2mm}
\end{figure}

\subsection{Key observations}\label{sec:factors}

If $\corr$ in \Cref{eq:final_goal} is accurate, solving \cref{eq:final_goal} is easy.
However, we have observed there exist three factors that cause learning-based registration to struggle in estimating $\corr$ when given out-of-domain data.

\vspace{-2mm}
\subsubsection{Voxel size and search radius}\label{sec:prob_user_defined}

First, dependencies of optimal search radius $r$ for local descriptors and voxel size $\voxel$ for each dataset are problematic.
The optimal parameters vary significantly across datasets due to differences in scale and point density~(\eg small indoor scenes vs.\ large outdoor spaces~\cite{Ao23CVPR-BUFFER})
Consequently, improper $r$ or $v$ can severely degrade registration performance by failing to account for specific scale and density characteristics of a given environment or sensor; see \Cref{sec:analyses}.
For instance, in \Cref{fig:key_elements}(a), as $\voxel$ controls the maximum number of points that can be fed into the network, 
a too-small $\voxel$ can trigger out-of-memory errors when outdoor data processed with parameters optimized for indoor environments are taken as input to the network.

In particular, most methods heavily depend on manual tuning, which hinders generalization.
Therefore, we employ a \textit{geometric bootstrapping} to adaptively determine $v$ and $r$ at test time based on the scale and point density of the given input clouds; see \Cref{sec:geometric}. 

\subsubsection{Input scale normalization}\label{sec:prob_scale_norm}

Next, directly feeding raw $x$, $y$, and $z$ values into the network leads to strong in-domain dependency~\cite{Huang21cvpr-PREDATORRegistration, Qin23tpami-GeoTransformer}.
That is, when a model fits to the training distribution, large scale discrepancies between training and unseen data can cause catastrophic failure~(see \Cref{fig:key_elements}(b) for an example of maximum range discrepancy).
For this reason, we conclude that normalizing input points within local neighborhoods~(or~\textit{patches)} is necessary to achieve generalizability, ensuring that their coordinates lie within a bounded range~(\eg $[-1, 1]$)~\cite{Ao21cvpr-Spinnet,Ao23CVPR-BUFFER}.

Based on these insights, we adopt patch-based descriptor generation as our pipeline for descriptor matching; see \Cref{sec:tri-scale-patch}.

\subsubsection{Keypoint detection}\label{sec:probpt_detector}

Following \Cref{sec:prob_scale_norm}, we observed that point-wise feature extractor modules in existing methods~\cite{Huang21cvpr-PREDATORRegistration, Qin23tpami-GeoTransformer, Yao24iccv-PARENet} are empirically brittle to out-of-domain data.
Because failed keypoint detection leads to the selection of unreliable and non-repeatable points as keypoints, it results in low-quality descriptors and ultimately degrades the quality of~$\mathcal{A}$~\cite{Harris88avc-HarrisCorner}.

An interesting observation is that replacing the learning-based detector with the farthest point sampling~(FPS) preserves registration performance.
For this reason, we adopt FPS over a learning-based module~(see \Cref{sec:ablation}).
Specifically, we apply it separately at local, middle, and global scales to account for multi-scale variations.


\section{BUFFER-X}
\newcommand{\spinnetout}{\mathcal{S}}
Building upon our key observations in \Cref{sec:factors}, we present our multi-scale zero-shot registration pipeline; see \Cref{fig:pipeline}.
First, the appropriate voxel size and radii for each cloud pair are predicted by geometric bootstrapping~(\Cref{sec:geometric}), considering the overall distribution of cloud points and the density of neighboring points, respectively.
Then, we extract Mini-SpinNet-based features~\cite{Ao23CVPR-BUFFER} for the sampled points via FPS at multiple scales~(\Cref{sec:tri-scale-patch}).
Finally, at the intra- and cross-scale levels, refined correspondences are estimated based on consensus maximization~\cite{Sun22ral-TriVoC,Shi24ral-RANSAC,Zhang24tpami-AcceleratingGloballyCM}~(\Cref{sec:hierarchical}) and serve as input for the final relative pose estimation using a solver.

\begin{figure*}[t!]
 	\centering
        \begin{subfigure}[b]{1.0\textwidth}
 		\includegraphics[width=1.0\textwidth]
 {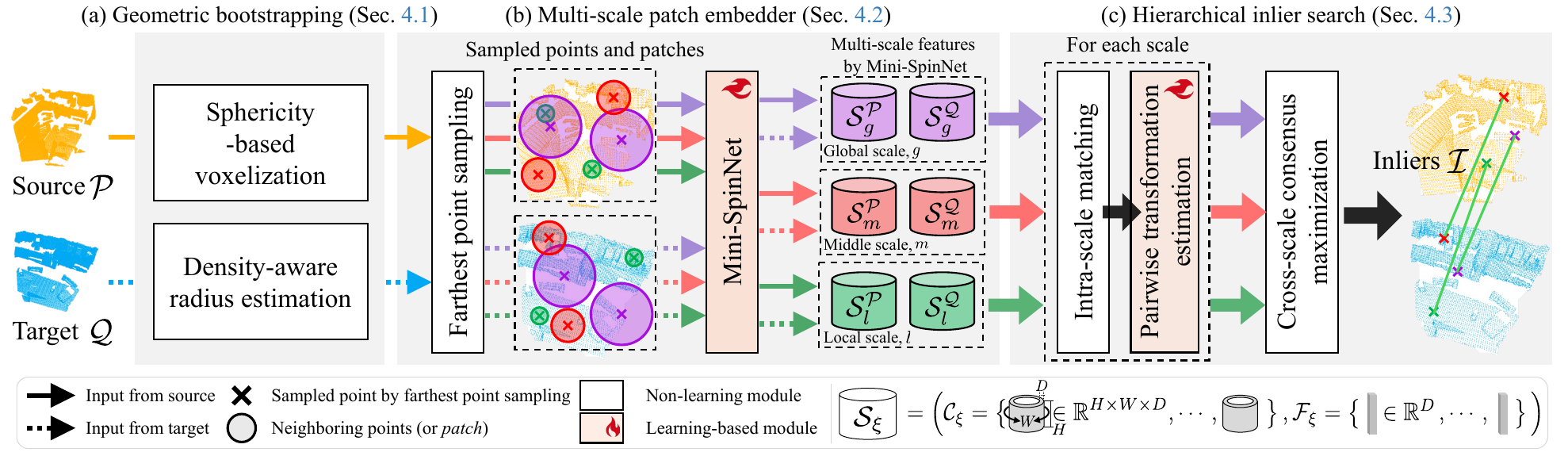}        
        \end{subfigure}
 	\captionsetup{font=small}
  \vspace{-0.2cm}
  \caption{Overview of our \textit{\oursname}, which mainly consists of three steps. (a) Geometric bootstrapping~(\Cref{sec:geometric}) to determine the appropriate voxel size and radii for the given source~$\mathcal{P}$ and target $\mathcal{Q}$ clouds. 
    (b)~Multi-scale patch embedder~(\Cref{sec:tri-scale-patch}) to generate patch-wise descriptor $\spinnetout_\xi$ for multiple scale~$\xi \in \{l, m, g\}$, where  $l$, $m$, and $g$ represent local, middle, and global scales, respectively. 
    Specifically, Mini-SpinNet~\cite{Ao23CVPR-BUFFER} outputs cylindrical feature maps $\mathcal{C}_\xi$ and vector feature set $\mathcal{F}_\xi$.
    (c)~Hierarchical inlier search~(\Cref{sec:hierarchical}), which first performs nearest neighbor-based intra-scale matching using $\mathcal{F}^\mathcal{P}_\xi$ and $\mathcal{F}^\mathcal{Q}_\xi$ at each scale, followed by pairwise transformation estimation.
    Finally, it identifies globally consistent inliers $\mathcal{I}$ across all scales to refine correspondences based on consensus maximization~\cite{Sun22ral-TriVoC,Zhang24tpami-AcceleratingGloballyCM}.} 	
    \label{fig:pipeline}
  \vspace{-2mm}
\end{figure*}

\subsection{Geometric bootstrapping}\label{sec:geometric}
\newcommand{\covsampledpoint}{\MC}
\newcommand{\Psampled}{\mathcal{P}_\text{sampled}}
\newcommand{\eigenidx}{a}



\vspace{2mm}
\noindent \textbf{Sphericity-based voxelization.} 
First, we determine the proper voxel size $v$ by leveraging sphericity, quantified using eigenvalues~\cite{Hansen21remotesensing-Classification,Alexiou24jivp-PointPCA}, to reflect how the cloud points are dispersed in space.
To this end, we apply principal component analysis (PCA)~\cite{Lim21ral-Patchwork} to the covariance of sampled points, which can efficiently capture point dispersion by analyzing eigenvalues while remaining computationally lightweight.

Formally, let $h(\mathcal{P}, \mathcal{Q})$ be a function that selects the larger point cloud based on cardinality, let $g(\mathcal{P}, \delta)$ be a function that samples $\delta\%$ of points from a given point cloud,
and let $\covsampledpoint \in \mathbb{R}^{3 \times 3}$ be the covariance of $g(h(\mathcal{P}, \mathcal{Q}), \delta_v)$, where $\delta_v$ is a user-defined sampling ratio.
Then, using PCA, three eigenvalues $\lambda_\eigenidx$ and their corresponding eigenvectors $\vv_\eigenidx$ are calculated as follows:
\begin{equation}
\covsampledpoint \vv_\eigenidx = \lambda_\eigenidx \vv_\eigenidx, \quad \eigenidx \in \{1,2,3\},
\label{eq:pca}
\end{equation}
\noindent which are assumed to be $\lambda_1 \geq \lambda_2 \geq \lambda_3$. Then, using these properties, we can compute the \textit{sphericity} $\frac{\lambda_3}{\lambda_1}$~\cite{Alexiou24jivp-PointPCA}, which quantifies how evenly a point cloud is distributed in space.
Since LiDAR points are primarily distributed along the sensor's horizontal plane (\ie forming a disc-like shape), $\frac{\lambda_3}{\lambda_1}$ tends to be low compared to RGB-D point clouds.

In addition, as observed in \Cref{fig:key_elements}(a), LiDAR point clouds require a larger voxel size; thus, we set $v$ as follows:
\newcommand{\coefficient}{\kappa}
\begin{equation}
    v =
    \begin{cases}
        \coefficient_{\text{spheric}} \sqrt{s}, & \text{if} \; \frac{\lambda_3}{\lambda_1} \geq \tau_v, \\
        \coefficient_{\text{disc}} \sqrt{s}, & \text{otherwise},
    \end{cases}
    \label{eq:voxel_size}
\end{equation}
\noindent where $\coefficient_{\text{spheric}}$ and $\coefficient_{\text{disc}}$ are constant user-defined coefficients across all datasets, satisfying $\coefficient_{\text{spheric}} < \coefficient_{\text{disc}}$, $\tau_v$ is a user-defined threshold,
and $s$ is the length that represents the spread of points along the eigenvector corresponding to the smallest eigenvalue $\vv_3$~(\ie $s = \max(\Psampled \cdot \vv_3) - \min(\Psampled \cdot \vv_3)$). 
Consequently, as $\frac{\lambda_3}{\lambda_1}$ and $s$ adapt based on the environment (\ie indoor or outdoor) and the field of view of the sensor type (\ie RGB-D or LiDAR point cloud),
\Cref{eq:voxel_size} enables the adaptive setting of $v$.

Hereafter, for brevity, we denote $f_v(\mathcal{P})$ and $f_v(\mathcal{Q})$ simply as $\mathcal{P}$ and $\mathcal{Q}$, respectively.

\newcommand{\rxi}{r_\xi}
\newcommand{\rxithres}{\tau_{\xi}}
\vspace{2mm}
\noindent \textbf{Density-aware radius estimation.}
Next, in contrast to some state-of-the-art approaches \cite{Ao23CVPR-BUFFER, Ao21cvpr-Spinnet} that use a single fixed user-defined search radius,
we determine $r$ at local, middle, and global scales, respectively, by considering the input point densities. 
Let neighboring search function within the radius $r$ around a query point $\srcpoint_q$ be:
\begin{equation}
\mathcal{N}\big(\srcpoint_q, \mathcal{P}, r\big) = \big\{ \srcpoint \in \mathcal{P} | \; \twonorm{\srcpoint - \srcpoint_q} \leq r \big\}.
\label{eq:search}
\end{equation}

\noindent Then, as presented in \Cref{fig:radius_search}(a), the radius for patch-wise descriptor generation for each scale $\rxi$ is defined as follows:
\begin{equation}
  \rxi = \argmin_{r} \left| \frac{1}{N} \sum_{\srcpoint_q \in \mathcal{P}_r} \text{card}\Big(\mathcal{N}\big(\srcpoint_q, \mathcal{P}_r, r\big) \Big) - \tau_{\xi} \right|,
  \label{eq:density_radius}
\end{equation}
where $\xi \in \{l, m, g\}$ denotes the scale level~(\ie local, middle, and global scale, respectively), 
$\tau_{\xi}$ denotes the user-defined threshold, which represents the desired neighborhood density~(\ie~average proportion of neighboring points relative to the total number of points), satisfying $\tau_l \leq \tau_m \leq \tau_g$ (accordingly, $r_l \leq r_m \leq r_g$ as presented in \Cref{fig:radius_search}(a)),
and $\mathcal{P}_r$ is a set of $N_r$ points sampled from $h(\mathcal{P}, \mathcal{Q})$, where $N_r$ is a user-defined parameter for radius estimation.
To account for cases where the points are too sparse, we set the maximum truncation radius $r_\text{max}$ as 
$\rxi \leftarrow \text{max}(\rxi, r_\text{max})$.

\subsection{Multi-scale patch embedder}\label{sec:tri-scale-patch}
\newcommand{\Pxi}{{\mathcal{P}_\xi}}
\newcommand{\Qxi}{{\mathcal{Q}_\xi}}

Next, with the voxelized point clouds $\mathcal{P}$ and $\mathcal{Q}$ and radii estimated by \Cref{eq:density_radius}, patch-wise descriptors are generated at each scale.

\vspace{2mm}
\noindent \textbf{Farthest point sampling.} As discussed in \Cref{sec:probpt_detector}, 
we sample $\Pxi$ from $\mathcal{P}$ at each scale using FPS to be free from a learning-based keypoint detector~(resp. $\Qxi$ from $\mathcal{Q}$).
Note that instead of extracting local, middle, and global-scale descriptors for the same sampled point~\cite{Yu21nips-CofiNet}, we independently sample separate points for each scale, as illustrated in \Cref{fig:pipeline}(b).
This is because we empirically found that different regions may require distinct scales for optimal feature extraction; see \Cref{sec:ablation}.

\newcommand{\patchidx}{b}
\newcommand{\ppatchidx}{\mathcal{P}_\xi}
\newcommand{\qpatchidx}{\mathcal{Q}_\xi}
\newcommand{\SPxi}{\mathcal{S}^{\mathcal{P}}_\xi}
\newcommand{\SQxi}{\mathcal{S}^{\mathcal{Q}}_\xi}
\newcommand{\FPxi}{\mathcal{F}^{\mathcal{P}}_\xi}
\newcommand{\FQxi}{\mathcal{F}^{\mathcal{Q}}_\xi}
\newcommand{\CPxi}{\mathcal{C}^{\mathcal{P}}_\xi}
\newcommand{\CQxi}{\mathcal{C}^{\mathcal{Q}}_\xi}

\vspace{2mm}
\noindent \textbf{Mini-SpinNet-based descriptor generation.}
Using multiple radii $\rxi$, 
we sample patches at three distinct scales, providing a more comprehensive multi-scale representation.
Then, we use Mini-SpinNet\cite{Ao23CVPR-BUFFER} for descriptor generation, which is a lightweight version of SpinNet~\cite{Ao21cvpr-Spinnet}.

\begin{figure}[t!]
 	\centering
 	\includegraphics[trim={15mm 0mm 0mm 13mm}, clip, width=1.0\columnwidth]{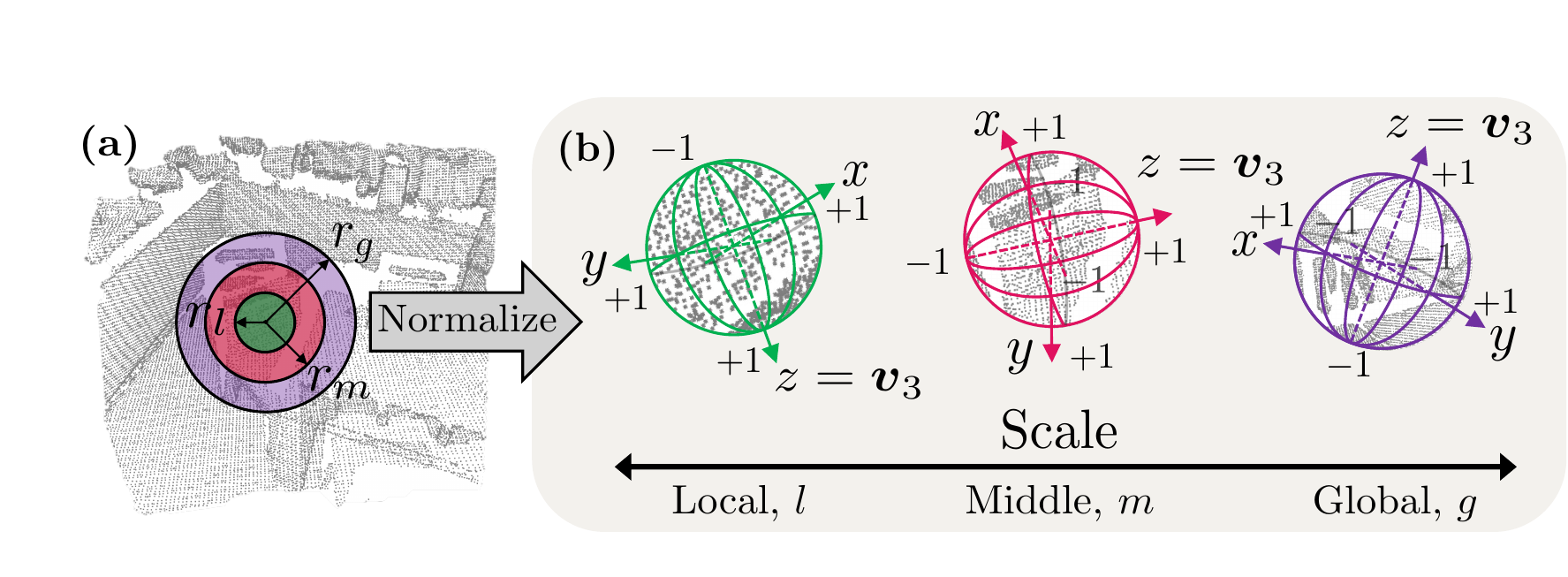}
    \vspace{-4mm}
 	\captionsetup{font=small}
 	\caption{(a)~Visual description of local~($r_l$), middle~($r_m$), and global~($r_g$) radii for the same point to illustrate scale differences and (b) normalized patches ranging from $[-1, 1]$. Note that their reference frames follow the eigenvectors obtained from principal component analysis~(PCA)~\cite{Lim21ral-Patchwork,Alexiou24jivp-PointPCA}. The $z$-axis is assigned to the eigenvector $\vv_3$, which corresponds to the smallest eigenvalue.} 	
    \label{fig:radius_search}
    \vspace{-5mm}
\end{figure}

In particular, building on the insights from \Cref{sec:prob_scale_norm}, we ensure the scale of points in each patch is normalized to a bounded range of $[-1, 1]$ by dividing by $\rxi$; see \Cref{fig:radius_search}(b). 
By doing so, we can resolve the dependency on the in-domain scale.
To maintain consistency across patches at all scales, we fix the patch size to $N_\text{patch}$ and randomly sample when a patch exceeds this size, ensuring a consistent number of points regardless of scale variations.




Finally, taking these normalized patches as inputs, Mini-SpinNet outputs a superset $\SPxi$ consisting of $D$-dimensional feature vectors~$\FPxi$ and cylindrical feature maps~$\CPxi$, where corresponds to $\Pxi$~(resp. $\SQxi$ consisting of $\FQxi$ and $\CQxi$ from $\Qxi$), as described in~\Cref{fig:pipeline}.
Note that while BUFFER~\cite{Ao23CVPR-BUFFER} utilizes learned reference axes to extract cylindrical coordinates, our approach defines the reference axes for each patch by applying PCA to the covariance of points within the patch, 
setting the $z$-direction as $\vv_3$ (as in \Cref{eq:pca} and illustrated by $z=\vv_3$ in \Cref{fig:radius_search}(b)), to eliminate potential dataset-specific inductive biases.


\subsection{Hierarchical inlier search}\label{sec:hierarchical}
\newcommand{\corrinit}{\mathcal{A}_\xi}
\newcommand{\CPximatched}{\widehat{\mathcal{C}}^{\mathcal{P}}_\xi}
\newcommand{\CQximatched}{\widehat{\mathcal{C}}^{\mathcal{Q}}_\xi}
\newcommand{\Pximatched}{\widehat{\mathcal{P}}_\xi}
\newcommand{\Qximatched}{\widehat{\mathcal{Q}}_\xi}

Here, we first perform inter-scale matching to get initial correspondences $\corrinit$ at each scale and then establish cross-scale consistent correspondences in a consensus maximization manner. 

\vspace{2mm}
\noindent \textbf{Intra-scale matching.} First, we perform nearest neighbor-based mutual matching~\cite{Lowe04ijcv} between $\FPxi$ and $\FQxi$, yielding matched correspondences $\corrinit$ at each scale. 
Using $\corrinit$, we extract the corresponding elements from $\CPxi$ and $\CQxi$, denoted as $\CPximatched$ and $\CQximatched$, and the sampled keypoints from $\Pxi$ and $\Qxi$ as $\Pximatched$ and $\Qximatched$, respectively~(\ie~$| \corrinit| = |\CPximatched | = |\CQximatched| = |\Pximatched | = |\Qximatched|$).

\vspace{2mm}
\noindent \textbf{Pairwise transformation estimation.} Next, using each cylindrical feature pair $\vc^{\vp} \in \CPximatched$ and $\vc^{\vq} \in \CQximatched$ at each scale, each of whose size is $\mathbb{R}^{H \times W \times D}$, we calculate pairwise 3D relative transformation between two patches. 
Here, $H$, $W$, and $D$ denote the height, sector size for the yaw direction along the $z$-axis of the reference axes, and feature dimensionality of a cylindrical feature, respectively.

As mentioned earlier, since the cylindrical feature is aligned with the local reference axes via PCA, 
the relative 3D rotation between $\vv^\vp_{3}$ (resp. $\vv^\vq_{3}$) and the unit $z$-axis, $\vz = [0 \; 0 \; 1]^\intercal$, can be calculated using Rodrigues' rotation formula~\cite{Mebius07arxiv-Derivation} as follows:
\begin{equation}
\MR^{\vp} = \MI + \sin(\theta^{\vp}) [\vn^{\vp}]_{\times} + \big(1 - \cos(\theta^{\vp})\big) [\vn^{\vp}]_{\times}^2,
\end{equation}
\noindent where $\vn^{\vp} = \vv^{\vp}_{3} \times \vz$, $\theta^{\vp} = \cos^{-1} \left( \vv^{\vp}_{3} \cdot \vz \right)$, and $[\cdot]_\times$ denotes the skew operator (resp. $\MR^\vq$).
Thus, once the yaw rotation between the two patches $\MR_{\text{yaw}}$ is determined, the full 3D rotation can be obtained as $\MR = \left(\MR^\vq\right)^\intercal \MR_{\text{yaw}} \MR^\vp$.

As explained by Ao~\etalcite{Ao23CVPR-BUFFER}, $\vc^{\vp}$ and $\vc^{\vq}$ follow discretized SO(2)-equivariant representation; 
thus, by finding the yaw rotation that maximizes circular cross-correlation between $\vc^{\vp}$ and $\vc^{\vq}$, we can estimate the relative SO(2) rotation $\MR_\text{yaw}$. 
To this end, a 4D matching cost volume $\MV \in \mathbb{R}^{H\times W \times W \times D}$ is constructed to represent the sector-wise differences between $\vc^{\vp}$ and $\vc^{\vq}$.
Then, $\MV$ is processed by a 3D cylindrical convolutional network~(3DCCN)~\cite{Ao21cvpr-Spinnet}, mapping $\MV$ to a score vector $\boldsymbol{\beta}$ of size~$W$.  

By applying the softmax operation $\sigma(\cdot)$ to $\boldsymbol{\beta}$, we obtain $\sigma(\boldsymbol{\beta})$, 
where the $w$-th element $\sigma_w(\boldsymbol{\beta}) \in [0, 1]$ represents the probability mass assigned to the discrete yaw rotation index~$w$.
Using this distribution, the discrete rotation offset~$d$ is computed as follows:
\begin{equation}
d = \sum_{w=1}^{W}  \sigma_w(\boldsymbol{\beta}) \times w.
\label{eq:calc_d}
\end{equation}

Finally, $\MR_\text{yaw}$ is calculated as follows:
\begin{equation}
\MR_\text{yaw} = 
\begin{bmatrix}
\cos \left(\frac{2\pi d}{W} \right) & -\sin \left(\frac{2\pi d}{W} \right) & 0 \\ 
\sin \left(\frac{2\pi d}{W} \right) & \cos \left(\frac{2\pi d}{W} \right) & 0 \\ 
0 & 0 & 1
\end{bmatrix}.
\end{equation}
 
Subsequently, the translation vector is given by $\vt = \vq - \MR \vp$, where $\vp \in \Pximatched$ and $\vq \in \Qximatched$ are a matched point pair.

\vspace{2mm}
\noindent \textbf{Cross-scale consensus maximization.}
Then, using per-pair $(\MR, \vt)$ estimates from all scales, the 3D point pairs with the largest cardinality across scales should be selected as the final inlier correspondences $\mathcal{I}$, ensuring cross-scale consistency.
To achieve this, we formulate the cross-scale inlier selection as \emph{consensus maximization} problem~\cite{Sun22ral-TriVoC,Zhang24tpami-AcceleratingGloballyCM}.

Formally, by denoting $N=\sum_{\xi} |\corrinit|$, let $(\MR, \vt) \in \mathcal{T}$ be a candidate transformation  set of size $N$, 
and let $(\vp_n, \vq_n) \in \mathcal{D}$ be the set of matched point pairs, where $n \in \{1, \dots, N\}$, $\vp_n \in \bigcup_\xi \Pximatched$ and $\vq_n \in \bigcup_\xi \Qximatched$.
Then, $\mathcal{I}$ is estimated as follows:
\begin{gather}
\max _{(\MR, \vt) \in \mathcal{T}, \; \mathcal{I}} |\mathcal{I}|  \label{eq:consensus_max} \\
\text {s.t.} ~~ \twonorm{\MR \vp_n  + \vt - \vq_n} <  \epsilon, ~~ \forall (\vp_n, \vq_n) \in \mathcal{I} \subseteq \mathcal{D},\nonumber
\end{gather}
\noindent where $\epsilon$ is an inlier threshold.


Finally, $\mathcal{I}$ is given as input to a solver, such as RANSAC~\cite{Fischler81} or TEASER++~\cite{Yang20tro-teaser} to estimate $\hat{\MR}$ and $\hat{\vt}$.
 For a fair comparison with existing approaches, we use RANSAC.

\subsection{Loss function and training}\label{sec:loss}
\noindent \textbf{Loss functions.}
Unlike BUFFER, which was trained in four stages, our network follows a relatively simpler two-stage training process thanks to its detector-free nature.
First, we train the feature discriminability of Mini-SpinNet descriptors using contrastive learning~\cite{Yew18eccv-3dfeatnet}, followed by training $d$ in \Cref{eq:calc_d} to improve transformation estimation accuracy.

In particular, we employ the Huber loss~\cite{Zhang97ivc-Parameter} $\rho_\text{Huber}(\cdot)$ for training $d$ to remain robust to outliers\cite{Barron19-generaladaptiverobustloss}, while balancing sensitivity to small errors, which is formulated as follows:

\begin{equation}
\rho_\text{Huber}(r) =
\begin{cases}
\frac{1}{2} r^2, & \text{if } |r| \leq \delta \\
\delta (|r| - \frac{1}{2} \delta), & \text{otherwise},
\end{cases}
\end{equation}

\noindent where $r$ denotes the residual and $\delta$ denotes the user-defined truncation threshold. 
Then, denoting the total number of data pairs by ${N_d}$, the $\gamma$-th predicted offset by $d_\gamma$, and the corresponding ground-truth offset by $d^*_\gamma$, the loss function $\mathcal{L}_{d}$ is defined as follows:

\begin{equation}
  \mathcal{L}_{d} = \frac{1}{N_d} \sum_{\gamma=1}^{N_d} \rho_\text{Huber}(d_\gamma - d^*_\gamma).
\end{equation}

\vspace{2mm}
\noindent \textbf{Patch distribution augmentation.} 
Furthermore, we propose an inter-patch point distribution augmentation to allow Mini-SpinNet to experience a wider variety of patch distribution patterns.
Specifically, we empirically sample the radius within $[\frac{2}{3}r, \frac{4}{3}r]$ based on a uniform probability. 
As mentioned in \Cref{sec:tri-scale-patch}, since $N_\text{patch}$ points within the radius are randomly selected as an input, a diverse set of patterns can be provided as $r$ varies.

Notably, training is conducted using only a single scale. This leverages the scale normalization characteristic of BUFFER-X, making it unnecessary to train with multi-scale separately.



\newcommand{\oracle}{\raisebox{-0.6ex}{\includegraphics[width=0.30cm]{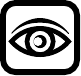}}}
\newcommand{\subsampling}{\raisebox{-0.6ex}{\includegraphics[width=0.30cm]{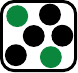}}}
\newcommand{\scalealign}{\raisebox{-0.6ex}{\includegraphics[width=0.30cm]{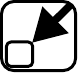}}}

\section{Experiments}


\noindent {\bf Datasets.}  As presented in \Cref{tab:overview_of_datasets}, we designed our generalizability benchmark using eleven different datasets~\cite{Zeng17cvpr-3dmatch,Huang21cvpr-PREDATORRegistration, Yeshwanth23iccv-Scannet++,Qingqing22iros-TIERS, Sun20cvpr-WaymoDataset, Geiger13ijrr-KITTI, Pomerleau12ijrr-ETH, Jung23ijrr-HeLiPR, Tian23iros-KimeraMultiExperiments, Ramezani20iros-NewerCollege} to ensure balanced consideration of the following aspects: a)~variation in environmental scales~(\ie~indoor and outdoor environments), b)~different scanning patterns with different sensor types, c)~acquisition setups, 
and d)~diversity of geographic and cultural environments as the data was collected across Europe, Asia, and the USA~(\ie \Oxford, \KAIST, and {\MIT} campuses, respectively). 
More details can be found in Appendices \ref{app:dataset_details} and \ref{app:dataset_rationale}.

\vspace{2mm}
\noindent{\bf Training settings.} Then, we only train the network on a single dataset, such as \ThreeDMatch~\cite{Zeng17cvpr-3dmatch} or \KITTI~\cite{Geiger13ijrr-KITTI}.
Using the same hyperparameters of BUFFER~\cite{Ao23CVPR-BUFFER}, we conducted a two-stage optimization~(\ie~Mini-SpinNet is first trained, followed by training the 3DCCN) and we used Adam optimizer~\cite{Kingma2014arxiv-Adam} with a learning rate of 0.001, a weight decay of 1e-6, and a learning rate decay of 0.5.
We used NVIDIA GeForce RTX 3090 with AMD EPYC 7763 64-Core.

\vspace{2mm}
\noindent \textbf{Testing settings.} In the case of the existing datasets~\cite{Zeng17cvpr-3dmatch,Huang21cvpr-PREDATORRegistration,Sun20cvpr-WaymoDataset, Geiger13ijrr-KITTI, Pomerleau12ijrr-ETH}, we follow the conventional given pairs.
The description of the newly employed datasets for evaluation can be found in Appendix~\ref{app:dataset_details}. 

\vspace{2mm}
\noindent \textbf{Evaluation Metrics.} As a key metric, we use the success rate, which directly assesses the robustness of global registration~\cite{Lim24ijrr-Quatropp}.
Specifically, a registration is deemed successful if the translation and rotation errors are within $\tau_\text{trans}$ and $\tau_\text{rot}$, respectively~\cite{Yew18eccv-3dfeatnet}. 
For successful cases, we evaluated the performance using relative translation error (RTE) and relative rotation error (RRE), which are defined as follows:

\newcommand{\numsuc}{N_\text{success}}
{\footnotesize
\begin{itemize}
	\item $\text{RTE}= \sum_{n=1}^{\numsuc} (\vt_{n, \text{GT}}-{\hat{\vt}}_{n})^{2} / \numsuc$,
	\item $\text{RRE}= \frac{180}{\pi} \sum_{n=1}^{\numsuc} | \cos^{-1} (\frac{\operatorname{Tr}\left({\hat{\MR}}_{n}^{\intercal} \MR_{n, \text{GT}}\right)-1}{2}) | / \numsuc $
\end{itemize}
}

\noindent where $\vt_{n, \text{GT}}$ and $\MR_{n, \text{GT}}$ denote the $n$-th ground truth translation and rotation, respectively; $\numsuc$ represents the number of successful registration. 
The more detailed criteria for determining a successful registration are provided in \cref{tab:overview_of_datasets}.

\newcommand{\subline}{ \cmidrule(lr){1-2} \cmidrule(lr){3-5} \cmidrule(lr){6-8} \cmidrule(lr){9-11}}
\newcommand{\errormetrics}{ RTE [cm] \; $\downarrow$ & RRE [$^\circ$] \; $\downarrow$ & Succ. [\%] \; $\uparrow$ }

\definecolor{myemerald}{rgb}{0.753, 0.898, 0.804}
\definecolor{mylightgreen}{rgb}{0.894, 0.933, 0.745}
\definecolor{myyellow}{rgb}{0.996, 0.972, 0.780}
\newcommand{\cg}{\cellcolor{gray!15}} 
\newcommand{\firstc}{\cellcolor{myemerald!100}}
\newcommand{\secondc}{\cellcolor{mylightgreen!100}}
\newcommand{\thirdc}{\cellcolor{myyellow!100}}
\newcommand{\ssp}{+ \subsampling}
\newcommand{\ora}{+ \oracle}
\newcommand{\osa}{+ {\oracle} + {\scalealign}}
\begingroup
\begin{table*}[t!]
        \setlength{\tabcolsep}{2pt}
        \centering
	{\scriptsize
		\begin{tabular}{l|l|cccccccccccc}
			\toprule \midrule
			& Env. & \multicolumn{5}{c}{Indoor} & \multicolumn{6}{c}{Outdoor} & \multirow{2}{*}{\begin{tabular}{@{}c@{}}Average \\ rank\end{tabular}} \\  \cmidrule(lr){3-7} \cmidrule(lr){8-13}
			& Dataset & \ThreeDMatch & \ThreeDLoMatch & \ScanNetppi & \ScanNetppF & \TIERS & \KITTI & \WOD & \KAIST &  \MIT & \ETH & \Oxford \\ \midrule
            \parbox[t]{5mm}{\multirow{3}{*}
            {\rotatebox[origin=c]{90}{\begin{tabular}{@{}c@{}}Conven- \\tional \end{tabular}}}}
       & FPFH~\cite{Rusu09icra-fast3Dkeypoints} + FGR~\cite{Zhou16eccv-FastGlobalRegistration} \ora & 62.53 & 15.42 & 77.68 & 92.31 & 80.60 & 98.74 & \firstc \textbf{100.00} & 89.80 & 74.78 & 91.87 & 99.00  & 9.55 \\
       & FPFH~\cite{Rusu09icra-fast3Dkeypoints} + Quatro~\cite{Lim22icra-Quatro} \ora & 8.22 & 1.74 & 9.88 & 97.27 & 86.57 & 99.10 & \firstc \textbf{100.00} & 91.46 & 79.57 & 51.05 & 91.03 & 10.73\\
       & FPFH~\cite{Rusu09icra-fast3Dkeypoints} + TEASER++~\cite{Yang20tro-teaser}  \ora & 52.00 & 13.25 & 66.15 & 97.22 & 73.13 & 98.92 & \firstc \textbf{100.00} & 89.20 & 71.30 & 93.69 & \secondc 99.34 & 10.00 \\ \midrule
			\parbox[t]{2mm}{\multirow{17}{*}{\rotatebox[origin=c]{90}{Deep learning-based}}}
       & FCGF~\cite{Choi19iccv-FCGF} & \cg 88.18 & \cg 40.09 & 72.90 &  88.69 & 55.96 & 0.00 & 0.00 & 0.00 & 0.00 & 54.98 & 0.00 & 15.00 \\
       & \ora & \cg 88.18 & \cg 40.09 & 85.87 & 88.69 & 78.62 & 90.27 & 97.69 & 92.91 & 92.61 & 54.98 & 93.68 & 10.18 \\
       & \osa & \cg 88.18 & \cg 40.09 & 85.87 & 88.69 & 80.11 & 94.41 & 97.69 & 93.55 & 93.04 & 55.53 & 95.68 & 9.55 \\
       & Predator~\cite{Huang21cvpr-PREDATORRegistration} & \cg 90.60 & \cg 62.40 & 75.94 & N/A & N/A & N/A & N/A & N/A & N/A &N/A & N/A & 15.73 \\
       &  \ora & \cg 90.60 & \cg 62.40 & 75.94 & 29.81 & 56.44 & 0.00 & 0.00 & 0.95 & 0.00 & 0.14 & 0.33 & 14.55 \\
       &  \osa & \cg 90.60 & \cg 62.40 & 75.94 & 86.01 & 75.74 & 77.29 & 86.92 & 87.09 & 79.56 & 54.42 & 93.68 & 11.82 \\
       & GeoTransformer~\cite{Qin23tpami-GeoTransformer} & \cg 92.00 & \cg 75.00 & 91.18 & N/A & N/A & N/A & N/A & N/A & N/A & N/A & N/A & 14.00 \\
       &  \ora & \cg 92.00 & \cg 75.00 & 91.18 & 7.54 & 5.06 & 0.36 & 0.77 & 0.25 & 0.87 & 0.00 & 0.33 & 13.09 \\
       &  \osa & \cg 92.00 & \cg 75.00 & 92.72  & \thirdc 97.02 & \secondc 92.99 & 92.43 & 89.23 & 91.86 & \secondc 95.65 & 71.53 & 97.01 & 6.27 \\
       & BUFFER~\cite{Ao23CVPR-BUFFER} & \cg 92.90 & \cg 71.80 & 92.72 & 93.75 & 62.30 & 0.00 & 1.54 & 0.50 & 6.96 & 97.62 & 0.66 & 10.45 \\
	&  \ora & \cg 92.90 & \cg 71.80 & \thirdc 93.01 & 94.69 & 88.96 & \thirdc 99.46 & \firstc \textbf{100.00} & \thirdc 97.24 & \secondc 95.65 & \secondc 99.30 & 99.00  & \thirdc 3.82  \\
       &  \osa & \cg 92.90 & \cg 71.80 & \thirdc 93.01 & 94.69 & 88.96 & \thirdc  99.46 & \firstc \textbf{100.00} & \thirdc 97.24 & \secondc 95.65 & \secondc 99.30 & 99.00 & \thirdc 3.82 \\
	& PARENet~\cite{Yao24iccv-PARENet} & \cg 95.00 & \cg \textbf{80.50} & 90.84 & N/A & N/A & N/A & N/A & N/A & N/A & N/A & N/A & 13.27 \\
      & \ora & \cg 95.00 & \cg \textbf{80.50} & 90.84 & 43.75 & 6.21 & 0.18 & 0.77 & 0.75 & 1.30 & 1.40 & 1.66 & 11.55 \\
      & \osa & \cg 95.00 & \cg \textbf{80.50} & 90.84 & 87.95 & 75.06  & 84.86 & 92.31 & 86.44 & 84.78 & 69.42 & 93.36 & 8.82 \\ \cmidrule(lr){2-14}
    & Ours with only $r_m$ & \cg 93.38 & \cg 71.69 & \secondc 93.10 &\secondc 99.60 & \thirdc 90.80 & \firstc \textbf{99.82} & \firstc \textbf{100.00} & \secondc 99.05 & \secondc 95.65 & \secondc 99.30 & \secondc 99.34 & \secondc 3.00 \\ 
    & Ours & \cg \textbf{95.58} & \cg 74.18 & \firstc \firstc \textbf{94.99} & \firstc \textbf{99.90} & \firstc \textbf{93.45} & \firstc \textbf{99.82} & \firstc \textbf{100.00} & \firstc \textbf{99.15} & \firstc \textbf{97.39} & \firstc \textbf{99.72} & \firstc \textbf{99.67} & \firstc \textbf{1.55} \\ \midrule \bottomrule
		\end{tabular}
	}
  \captionsetup{font=small}
  \vspace{-2mm}
  \caption{Quantitative comparison of generalization performance in terms of success rate (\%).
  Deep learning-based models were trained only on {\ThreeDMatch}~\cite{Zeng17cvpr-3dmatch} and RANSAC was used with a maximum iteration of 50K. 
The icons represent oracle tuning~(\oracle) for voxel size and radius, and scale alignment~(\scalealign) to normalize dataset scales to be similar to that of {\ThreeDMatch} data~(\eg the scale of {\KITTI}, which typically uses a voxel size of 0.3\,m, is adjusted to match the scale of {\ThreeDMatch}, where 0.025\,m is commonly used, by dividing by $\frac{0.3}{0.025}$).}
	\label{table:success_rates}
  \vspace{-4mm}
\end{table*}
\endgroup
 
\subsection{Analyses on the generalization}\label{sec:analyses}

First, we demonstrate that existing methods struggle in achieving out-of-the-box generalization, leading to performance degradation due to the issues explained in \Cref{sec:our_preliminaries}.
In this experiment, we mainly used renowned learning-based approaches: FCGF~\cite{Choi19iccv-FCGF}, Predator~\cite{Huang21cvpr-PREDATORRegistration}, GeoTransformer~(\textit{GeoT} for brevity)~\cite{Qin23tpami-GeoTransformer}, BUFFER~\cite{Ao23CVPR-BUFFER}, and PARENet~\cite{Yao24iccv-PARENet}.

As shown in \Cref{table:success_rates}, models trained with small voxel sizes and search radii for indoor datasets exhibited substantial performance degradation in outdoor scenarios.
In particular, as explained in \Cref{sec:prob_user_defined}, we observed that some approaches did not even work due to out-of-memory issues~(\ie N/A in \Cref{table:success_rates}) caused by an excessively large number of input points.

Once a properly user-tuned voxel size and search radius were provided (referred to as \textit{oracle tuning},  \raisebox{-0.3ex}{\includegraphics[width=0.30cm]{pics/symbols_oracle.pdf}}), BUFFER showed a remarkable performance increase owing to its patch-wise input scale normalization characteristics.
In contrast, other approaches still showed relatively lower success rates 
because the networks received point clouds with magnitudes not encountered during training.
This potential limitation is further evidenced by the performance improvement of Predator after scale alignment, supporting our claim that scale normalization is a key factor in achieving generalizability.

\subsection{Performance comparison}\label{sec:sota_comparison}

\begin{figure}[t!]
    \centering
    \includegraphics[width=1\columnwidth]{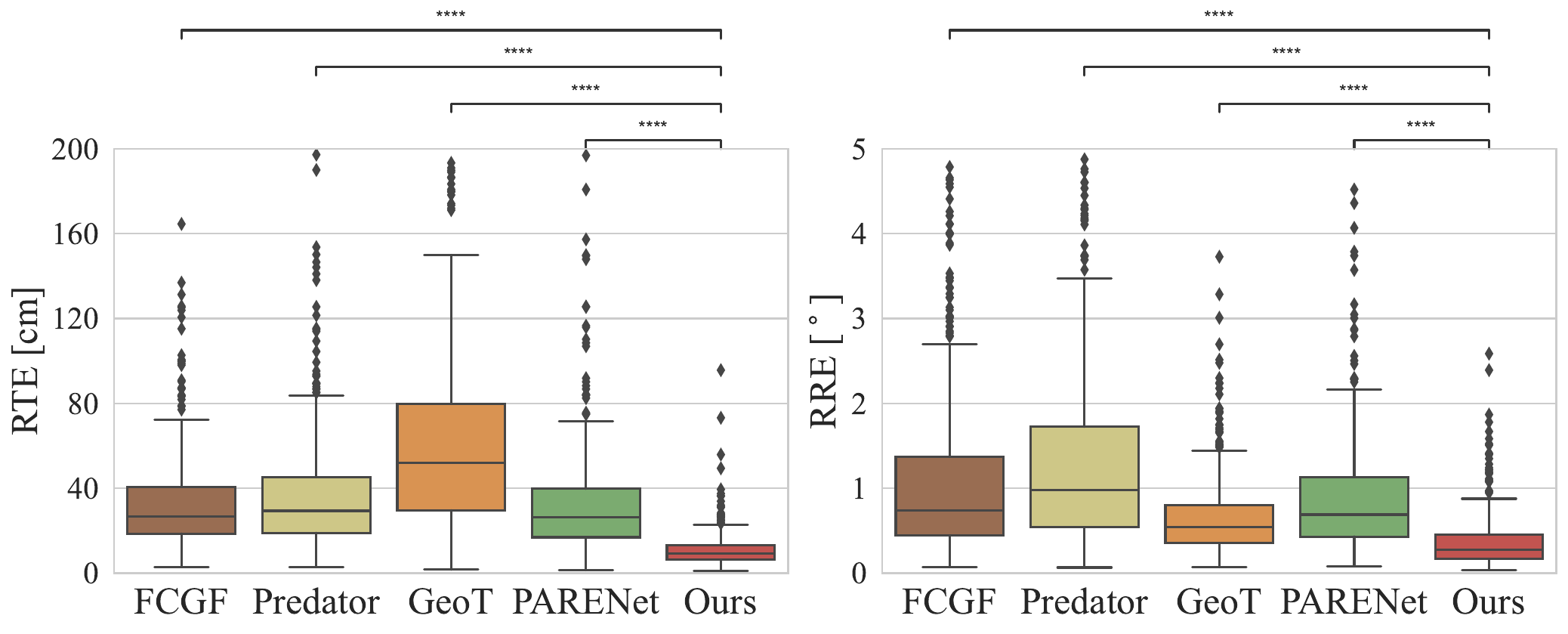}
    \vspace{-4mm}
    \captionsetup{font=small}
    \caption{Relative translation error~(RTE) and relative rotation error~(RRE) of our approach to state-of-the-art methods, all trained on {\ThreeDMatch} and tested on \KITTI, with oracle tuning and scale alignment, corresponding to those in \Cref{table:success_rates} under the + \raisebox{-0.4ex}{\includegraphics[width=0.30cm]{pics/symbols_oracle.pdf}} + \raisebox{-0.4ex}{\includegraphics[width=0.30cm]{pics/symbols_scale_align.pdf}} setting.
    The **** annotations indicate measurements with a $p$-value $< 10^{-4}$ after a paired $t$-test.
    }
    \label{fig:poor_alignment}
    \vspace{-4mm}
\end{figure}

Second, we demonstrate that our proposed algorithm, inspired by these key observations, achieves substantial out-of-the-box generalization; see \Cref{table:success_rates}.
In particular, our approach achieved lower RTE and RRE than state-of-the-art approaches even with oracle tuning and scale alignment~(\Cref{fig:poor_alignment}), while maintaining competitive in-domain performance~(\Cref{table:kitti_results}).
Therefore, this experimental evidence supports our claim that our algorithm achieves a high generalization capability and successfully performs in-domain scenarios.

\newcommand{\graysetup}{\color{gray!70}}
\begin{table}[t!]
    \captionsetup{font=small}
    \centering    
    \setlength{\tabcolsep}{2pt}
    {\scriptsize
        \begin{tabular}{l|lccc}
            \toprule \midrule
            &{Method} & {RTE [cm]\,$\downarrow$} & {RRE [°]\,$\downarrow$} & {Succ. rate [\%]\,$\uparrow$} \\
            \midrule
            \parbox[t]{2mm}{\multirow{4}{*}{\rotatebox[origin=c]{90}{Conventional.}}}
            & G-ICP~\cite{Segal09rss-GeneralizedICP}  & 8.56  & \firstc \textbf{0.22} & 37.95 \\            
            & FPFH~\cite{Rusu09icra-fast3Dkeypoints} + FGR~\cite{Zhou16eccv-FastGlobalRegistration} & 18.75  & 0.38  & 98.74  \\
            & FPFH~\cite{Rusu09icra-fast3Dkeypoints} + Quatro~\cite{Lim22icra-Quatro}  & 18.56  & 0.93  & 99.10 \\
            & FPFH~\cite{Rusu09icra-fast3Dkeypoints} + TEASER++~\cite{Yang20tro-teaser}  & 15.35  & 0.68  & 98.92  \\ 
            & KISS-Matcher~\cite{Lim25icra-KISSMatcher}  & 21.33  & 0.96  & 99.46  \\ 
            \midrule
            \parbox[t]{2mm}{\multirow{11}{*}{\rotatebox[origin=c]{90}{Learning-based}}} & 3DFeat-Net~\cite{Yew18eccv-3dfeatnet} & 25.90 &  0.57  & 95.97 \\
            & FCGF~\cite{Choi19iccv-FCGF} & \thirdc 6.47 &  \secondc {0.23} & 98.92 \\
            & DIP~\cite{Poiesi21icpr-DIP}  & 8.69 & 0.44  & 97.30 \\
            & Predator~\cite{Huang21cvpr-PREDATORRegistration} & \secondc {5.60} & 0.24  & \firstc \textbf{99.82} \\
            & SpinNet~\cite{Ao21cvpr-Spinnet} & 9.88  & 0.47  & 99.10 \\
            & CoFiNet~\cite{Yu21nips-CofiNet} & 8.20  & 0.41  & \firstc \textbf{99.82} \\ 
            & D3Feat~\cite{Bai20cvpr-D3Feat} & 11.00  & 0.24  & \firstc \textbf{99.82} \\%
            & GeDi~\cite{Poiesi22pami-GeDi} & 7.55 & 0.33 & \firstc \textbf{99.82} \\   
            & GeoTransformer~\cite{Qin23tpami-GeoTransformer} & 7.40 & 0.27 & \firstc \textbf{99.82} \\ 
            & BUFFER~\cite{Ao23CVPR-BUFFER} & 7.46 & 0.26 & 99.64 \\ 
            & PARE-Net~\cite{Yao24iccv-PARENet} & \firstc \textbf{4.90} & \secondc 0.23 & \firstc \textbf{99.82} \\ 
            & Ours & 7.74 & 0.27 & \firstc \textbf{99.82} \\            
            \midrule \bottomrule
        \end{tabular}
    }
    \vspace{-2mm}
    \caption{In-domain quantitative results in terms of relative translation error~(RTE), relative rotation error~(RRE), and success rate on \KITTI~\cite{Yew18eccv-3dfeatnet}.}
    \label{table:kitti_results}
    \vspace{-2mm}
\end{table}

\subsection{Ablation study}\label{sec:ablation}

\begin{figure}[t!]
    \centering
    \begin{subfigure}[b]{.27\textwidth}
        \includegraphics[width=1\textwidth]{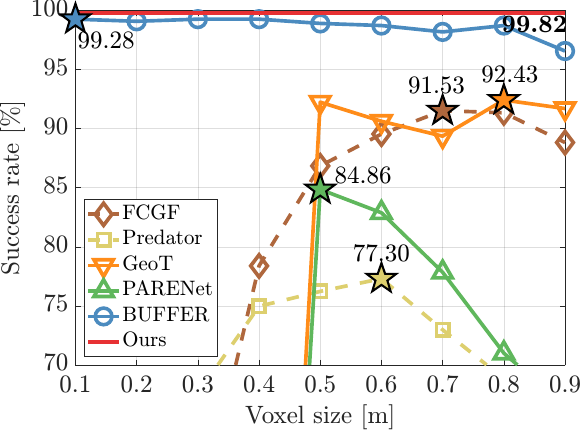}
        \caption{}
    \end{subfigure}
    \begin{subfigure}[b]{.19\textwidth}
        \includegraphics[width=1\textwidth]{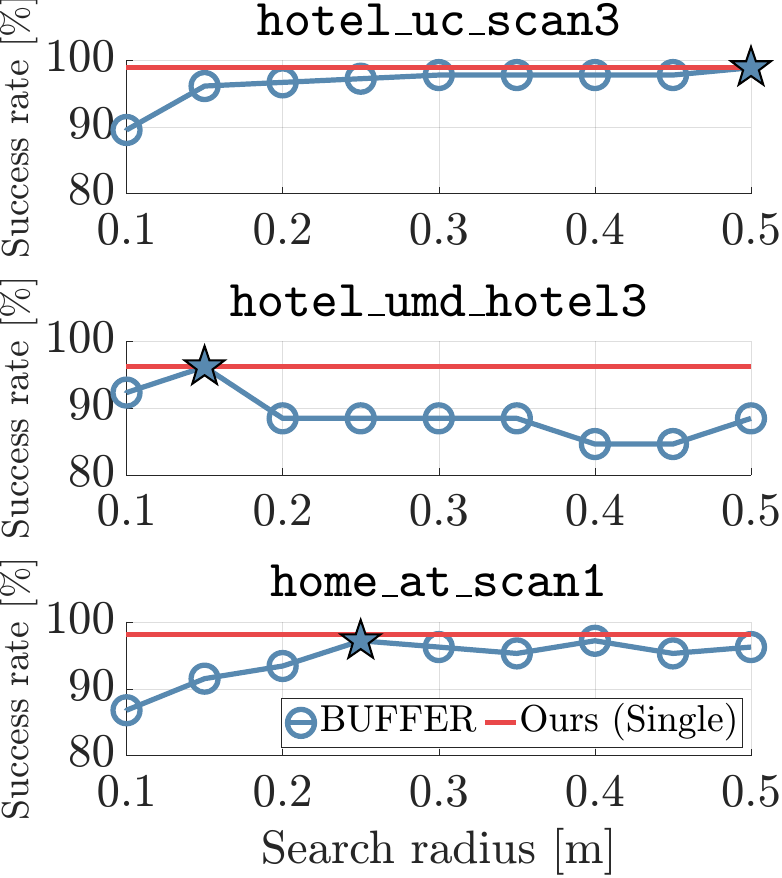}
        \caption{}
    \end{subfigure}
    \vspace{-2mm}
    \captionsetup{font=small}
    \caption{Effect of voxel size and search radius on success rates, where \ding{72} indicates the best performance after tuning. 
    (a)~Impact of voxel size (with \scalealign) on models trained on {\ThreeDMatch}, evaluated on \KITTI. 
    (b)~BUFFER-X vs. BUFFER across different radii, showing that optimal radius may vary within the same dataset.}
    \label{fig:ablations}
    \vspace{-3mm}
\end{figure}

\noindent \textbf{Impact of geometric bootstrapping.}
\Cref{fig:ablations} supports our claim that voxel size and radius have a more direct impact on performance than expected.
Unlike BUFFER, which relies on manual tuning, our method automatically estimates optimal $v$ and $r$, adapting to different scenes and maintaining consistency across varying dataset densities~(\Cref{fig:ablations}(b)).

\vspace{2mm}
\noindent \textbf{Learning-based detector vs. Farthest point sampling}.
Interestingly, FPS rather showed better performance than the learning-based keypoint detector in BUFFER\cite{Ao23CVPR-BUFFER}, even in its training domain~(\ie in {\ThreeDMatch} and \ThreeDLoMatch). 
In particular, the performance gap sometimes becomes more pronounced in the out-of-domain scenes, demonstrating that robust cross-domain generalization can be achieved without the need for additional learning-based keypoint selection.

\begin{figure}[t!]
 	\centering
 	\includegraphics[width=1.0\columnwidth]{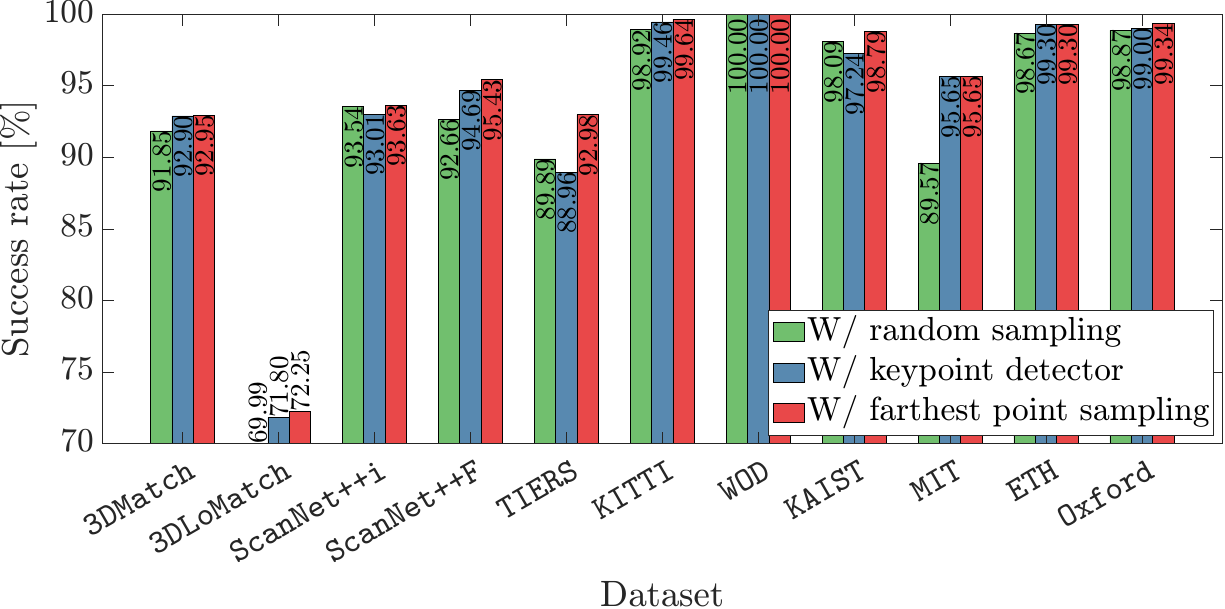}
 	\captionsetup{font=small}
        \vspace{-6mm}
 	\caption{Comparison of success rates between random sampling, learning-based keypoint detection in BUFFER~\cite{Ao23CVPR-BUFFER}, and our farthest point sampling (FPS) strategy, showing that FPS performs comparably or even better across various datasets.} 	
    \label{fig:efcnn_fps}
    \vspace{-2mm}    
\end{figure}

\begin{figure}[t!]
    \centering
    \begin{subfigure}[b]{.40\textwidth}
        \includegraphics[width=1\textwidth]{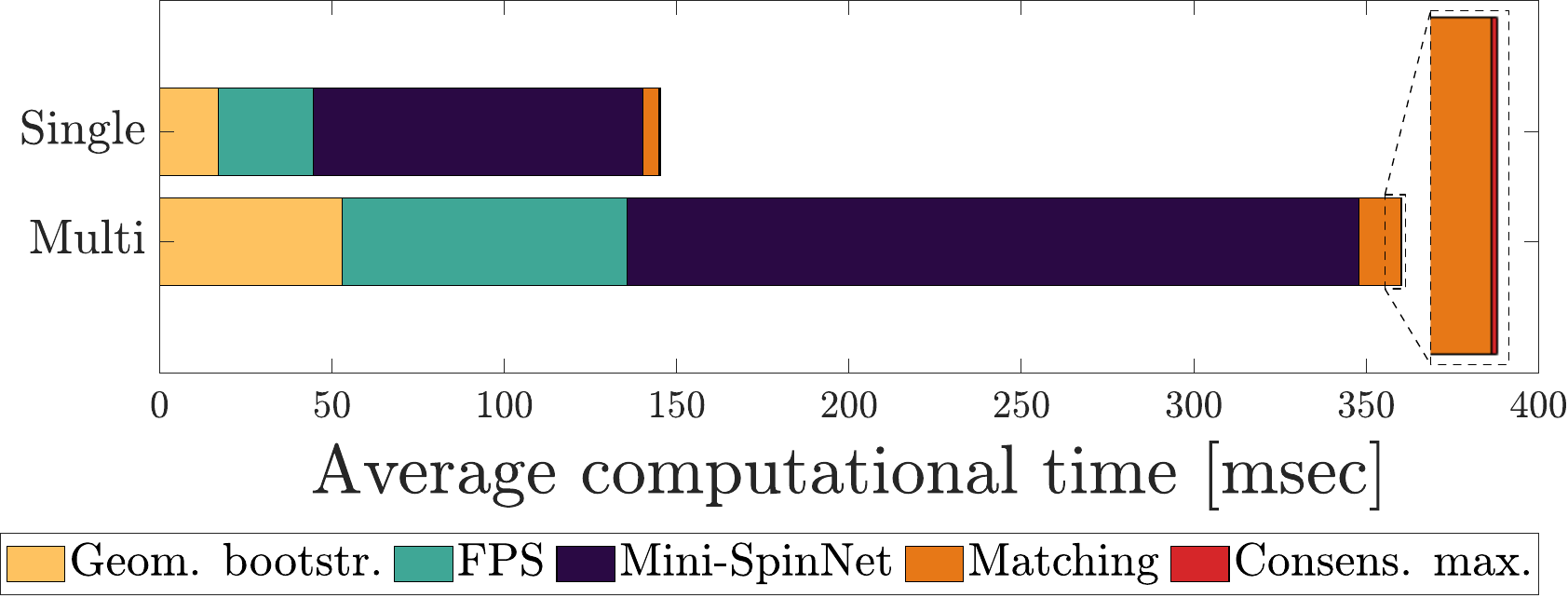}
    \end{subfigure}
    \captionsetup{font=small}
    \vspace{-2mm}
    \caption{Area plot of computation time per scale for each module on an NVIDIA GeForce RTX 3090 using \ThreeDMatch~\cite{Zeng17cvpr-3dmatch}. 
    }
    \label{fig:speed_sing_and_multi}
    \vspace{-5mm}
\end{figure}

\vspace{2mm}
\noindent \textbf{Impact of multi-scale.}
While we demonstrated that using only the middle scale in a single-scale setting is comparable to the multi-scale approach (see \Cref{table:success_rates}), \Cref{table:ablation-triscale} shows that incorporating multiple scales further increases the success rate.
This implies that correspondences across scales complement each other, leading to higher success rates.
However, increasing the number of scales introduces a trade-off between accuracy and computational cost (\Cref{fig:speed_sing_and_multi}), allowing users to balance efficiency and performance based on their specific needs.


\subsection{Limitations}\label{sec:limitation}


As seen in \Cref{table:success_rates}, our approach showed lower success rate in \ThreeDLoMatch, which only have 10-30\% overlaps.
This is because Eq.~\Cref{eq:consensus_max} selects correspondences with the largest cardinality as inliers.
However, in partial overlap scenarios, maximizing the number of correspondences might not yield the actual global optimum~(\ie there might exist $\mathcal{I}^*$ that satisfies $|\mathcal{I}^*| \leq |\mathcal{I}|$ but leads to a better relative pose estimate).
This highlights a trade-off between generalization and robustness to partial overlaps (see Appendix~\ref{app:trade_off} for further analyses).

\begin{table}[t!]
    \centering  
    \setlength{\tabcolsep}{3pt}
    {\scriptsize
    \begin{tabular}{ccc|cccc}
        \toprule \midrule
    Local & Middle  & Global  &  {RTE [cm]\,$\downarrow$}  &   {RRE [°]\,$\downarrow$} & {Succ. rate [\%]\,$\uparrow$} & Hz $\uparrow$ \\ \midrule
     \checkmark &   &        &  6.57 & 2.15 & 84.06 & \firstc \textbf{5.61}  \\ 
      &   \checkmark    &        & 5.87 & 1.85 & 93.38 & \thirdc 5.47  \\ 
      &   & \checkmark           &  6.06 & 1.91 & 93.57 & \secondc 5.49  \\ 
      \checkmark &      \checkmark & & \firstc \textbf{5.73} & \secondc 1.81 & \thirdc 94.31 & 2.35 \\  
      \checkmark &      &  \checkmark  & \secondc 5.77 & \thirdc 1.81 & 94.02 & 2.36 \\  
        & \checkmark  &  \checkmark  & 5.78 & 1.81 & \secondc 94.62 & 2.33 \\  
      \checkmark & \checkmark & \checkmark & \thirdc 5.78 & \firstc \textbf{1.79} & \firstc \textbf{95.58} & 1.81\\  \midrule \bottomrule
    \end{tabular}
    }
    \vspace{-2mm}
    \captionsetup{font=small}
    \caption{Ablation study: the impact of different scale combinations on registration performance in the \ThreeDMatch~\cite{Zeng17cvpr-3dmatch}.}
    \label{table:ablation-triscale}
    \vspace{-2mm}
\end{table}
\vspace{-1mm}

\section{Conclusion}

In this study, we addressed the generalization limitations of deep learning-based registration and analyzed key factors hindering it.
Based on these insights, we proposed a fully zero-shot pipeline, \textit{\oursname}, and introduced a comprehensive benchmark for evaluating generalization on real-world point cloud data.
In future works, we plan to study how to boost the inference speed for better usability.

\subsection*{Acknowledgments}
This work was supported by IITP grant (RS-2021-II211343:
AI Graduate School Program at Seoul National University) (5\%), and by NRF grant funded by the Korea government (MSIT) (No. 2023R1A1C200781211 (65\%) and No. RS-2024-00461409 (30\%), respectively). \\
\vspace{-5mm}
{
\small
\balance
\bibliographystyle{ieee_fullname}
\bibliography{myRefs.bib, main.bib}

@string{ar = {Autonomous Robots}}

@string{cvpr = {IEEE Conf. on Computer Vision and Pattern Recognition (CVPR)}}

@string{eccv = {European Conf. on Computer Vision (ECCV)}}

@string{fusion = {Intl. Conf. on Information Fusion, FUSION}}

@string{iccv = {Intl. Conf. on Computer Vision (ICCV)}}

@string{icpr = {Intl. Conf. on Pattern Recognition (ICPR)}}

@string{icra = {IEEE Intl. Conf. on Robotics and Automation (ICRA)}}

@string{ijcv = {Intl. J. of Computer Vision}}

@string{ijrr = {Intl. J. of Robotics Research}}

@string{iros = {IEEE/RSJ Intl. Conf. on Intelligent Robots and Systems (IROS)}}

@string{pami = {{IEEE} Trans. Pattern Anal. Machine Intell.}}

@string{rss = {Robotics: Science and Systems (RSS)}}

@string{ral = {{IEEE} Robotics and Automation Letters}}

@string{springer = {Springer Verlag}}

@string{tro = {{IEEE} Trans. Robotics}}

@article{Fischler81,
	Author = {M. Fischler and R. Bolles},
	Journal = {Commun. ACM},
	Pages = {381-395},
	Title = {Random sample consensus: a paradigm for model fitting with application to image analysis and automated cartography},
	Volume = 24,
	Year = 1981}

@inproceedings{Qi17cvpr-pointnet,
	title={{Pointnet: Deep learning on point sets for 3D classification and segmentation}},
	author={Qi, Charles R and Su, Hao and Mo, Kaichun and Guibas, Leonidas J},
	booktitle= cvpr,
	pages={652--660},
	year={2017}
}

@article{Lowe04ijcv,
	Author = {D.G. Lowe},
	Journal = IJCV,
	Number = 2,
	Pages = {91-110},
	Title = {Distinctive Image Features from Scale-Invariant Keypoints},
	Volume = 60,
	Year = 2004}

@article{Yang16pami-goicp,
	fullauthor = {Yang, Jiaolong and Li, Hongdong and Campbell, Dylan and Jia, Yunde},
	author = {J. Yang and H. Li and D. Campbell and Y. Jia},
	title = {{Go-ICP}: A Globally Optimal Solution to {3D ICP} Point-Set Registration},
	journal = pami,
	issue_date = {November 2016},
	volume = {38},
	number = {11},
	month = nov,
	year = {2016},
	issn = {0162-8828},
	pages = {2241--2254},
	numpages = {14},
	acmid = {3084750},
	publisher = {IEEE Computer Society},
	address = {Washington, DC, USA}
}

@article{Hartley09ijcv-globalRotationRegistration,
	title={Global optimization through rotation space search},
	fullauthor={Hartley, Richard I. and Kahl, Fredrik},
	author={R.I. Hartley and F. Kahl},
	journal=ijcv,
	volume={82},
	number={1},
	pages={64--79},
	year={2009},
	publisher={Springer}
}

@inproceedings{Yew18eccv-3dfeatnet,
	title={3DFeat-Net: Weakly Supervised Local 3D Features for Point Cloud Registration},
	fullauthor={Yew, Zi Jian and Lee, Gim Hee},
	author={Z.J. Yew and G.H. Lee},
	booktitle=eccv,
	year={2018}
}

@inproceedings{Choy20cvpr-deepGlobalRegistration,
	title={Deep Global Registration},
	author={Choy, Christopher and Dong, Wei and Koltun, Vladlen},
	booktitle=cvpr,
	year={2020}
}

@inproceedings{Rusu09icra-fast3Dkeypoints,
	title={Fast point feature histograms (FPFH) for 3D registration},
	fullauthor={Rusu, Radu Bogdan and Blodow, Nico and Beetz, Michael},
	author={R.B. Rusu and N. Blodow and M. Beetz},
	booktitle=icra,
	pages={3212--3217},
	year={2009},
	organization={Citeseer}
}

@article{Olsson09pami-bnbRegistration,
	title={Branch-and-bound methods for euclidean registration problems},
	author={Olsson, Carl and Kahl, Fredrik and Oskarsson, Magnus},
	shortauthor={C. Olsson and F. Kahl and M. Oskarsson},
	journal=pami,
	volume={31},
	number={5},
	pages={783--794},
	year={2009},
	publisher={IEEE}
}

@article{Oomerleau12ijrr-ethpc,
	fullauthor = {Pomerleau, Fran{\c c}ois and Liu, M. and Colas, Francis and Siegwart, Roland},
	author = {F. Pomerleau and M. Liu and F. Colas and R. Siegwart},
	title = {{Challenging data sets for point cloud registration algorithms}},
	journal = ijrr,
	year = {2012},
	volume = {31},
	number = {14},
	pages = {1705--1711}
}

@article{Pomerleau13auro-ICPcomparison,
	title={Comparing {ICP} variants on real-world data sets},
	fullauthor={Pomerleau, Fran{\c{c}}ois and Colas, Francis and Siegwart, Roland and Magnenat, St{\'e}phane},
	author = {F. Pomerleau and F. Colas and R. Siegwart and S. Magnenat},
	journal={Autonomous Robots},
	volume={34},
	number={3},
	pages={133--148},
	year={2013},
	publisher={Springer}
}

@inproceedings{Bai20cvpr-D3Feat,
	title =        {D3Feat: Joint Learning of Dense Detection and Description of
	3D Local Features},
	author =       {Bai, Xuyang and Luo, Zixin and Zhou, Lei and Fu, Hongbo and
	Quan, Long and Tai, Chiew-Lan},
	booktitle =    cvpr,
	year =         2020
}

@inproceedings{Zeng17cvpr-3dmatch,
	title={{3DMatch: Learning the matching of local 3d geometry in range scans}},
	author={Zeng, Andy and Song, Shuran and Nie{\ss}ner, Matthias and Fisher, Matthew and Xiao, Jianxiong and Funkhouser, T},
	booktitle=cvpr,
	pages={4},
	year={2017}
}

@inproceedings{Segal09rss-GeneralizedICP,
	title = {Generalized {{ICP}}},
	author = {Segal, Aleksandr and Haehnel, Dirk and Thrun, Sebastian},
	year = {2009},
	month = {Jun.},
	booktitle = rss, 
	pages = {},
	doi = {10.15607/RSS.2009.V.021}
}

@inproceedings{Huang21cvpr-PREDATORRegistration,
	title = {{{PREDATOR}}: {{Registration}} of {{3D Point Clouds}} with {{Low Overlap}}},
	shorttitle = {{{PREDATOR}}},
	booktitle = cvpr, 
	author = {Huang, Shengyu and Gojcic, Zan and Usvyatsov, Mikhail and Wieser, Andreas and Schindler, Konrad},
	year = {2021},
	month = {Jun.},
	pages = {4265--4274}
}

@inproceedings{Koide21icra-VGICP,
  title={Voxelized \text{GICP} for Fast and Accurate \text{3D} Point Cloud Registration},
  author={Koide, Kenji and Yokozuka, Masashi and Oishi, Shuji and Banno, Atsuhiko},
  booktitle=icra,
  pages={11054--11059},
  year={2021}
}

@article{Bernreiter21ral-PHASER,
	title={{PHASER: A robust and correspondence-free global point cloud} Registration},
	author={Bernreiter, Lukas and Ott, Lionel and Nieto, Juan and Siegwart, Roland and Cadena, Cesar},
	journal=ral,
	volume={6},
	number={2},
	pages={855--862},
	year={2021},
}

@article{Vizzo23ral-KISSICP,
  title={{KISS-ICP: In defense of point-to-point ICP -- Simple, accurate, and robust registration if done the right way}},
  author={Vizzo, Ignacio and Guadagnino, Tiziano and Mersch, Benedikt and Wiesmann, Louis and Behley, Jens and Stachniss, Cyrill},
  journal= ral ,
  year={2023},
  pages={1029--1036}
}

@article{Lim24ijrr-Quatropp,
  title={{Quatro++: R}obust global registration exploiting ground segmentation for loop closing in {LiDAR SLAM}},
  author={Lim, Hyungtae and Kim, Beomsoo and Kim, Daebeom and Mason Lee, Eungchang and Myung, Hyun},
  journal=ijrr,
  pages={685--715},
  year={2024},
  doi={10.1177/02783649231207654}
}

@article{Aoki24icra-3DBBS,
  title={{3D-BBS: Global localization for 3D point cloud scan matching using branch-and-bound algorithm}},
  author={Aoki, Koki and Koide, Kenji and Oishi, Shuji and Yokozuka, Masashi and Banno, Atsuhiko and Meguro, Junichi},
  journal=icra,
  pages={1796--1802},
  year={2024}
}

@article{Cattaneo22tro-LCDNet,
  title={{LCDNet: Deep loop closure detection and point cloud registration for LiDAR SLAM}},
  author={Cattaneo, Daniele and Vaghi, Matteo and Valada, Abhinav},
  journal=tro,
  volume={38},
  number={4},
  pages={2074--2093},
  year={2022}
}

@inproceedings{Choi19iccv-FCGF,
  title={Fully convolutional geometric features},
  author={Choy, Christopher and Park, Jaesik and Koltun, Vladlen},
  booktitle=iccv,
  pages={8958--8966},
  year={2019}
}

@inproceedings{Lim22icra-Quatro,
    title={A single correspondence is enough: Robust global registration to avoid degeneracy in urban environments},
    author={Lim, Hyungtae and Yeon, Suyong and Ryu, Soohyun and Lee, Yonghan and Kim, Youngji and Yun, Jaeseong and Jung, Euigon and Lee, Donghwan and Myung, Hyun},
    booktitle= icra,
    pages={8010--8017},
    year={2022}
}

@inproceedings{Rouhani11iccv-CorrespondenceFreeReg,
	title={Correspondence free registration through a point-to-model distance minimization},
	author={Rouhani, Mohammad and Sappa, Angel D},
	booktitle=iccv,
	pages={2150--2157},
	year={2011}
}

@article{Brown19pr-AFamiliyofBnB,
	title={{A family of globally optimal branch-and-bound algorithms for 2D-3D correspondence-free registration}},
	author={Brown, Mark and Windridge, David and Guillemaut, Jean-Yves},
	journal={Pattern Recognition},
	volume={93},
	pages={36--54},
	year={2019}
}

@inproceedings{Chum03jprs-LocallyOptimizedRANSAC,
	title={{Locally optimized RANSAC}},
	author={Chum, Ond{\v{r}}ej and Matas, Ji{\v{r}}{\'\i} and Kittler, Josef},
	booktitle={Joint Pattern Recognition Symposium},
	pages={236--243},
	year={2003}
}

@article{Choi97jcv-RANSAC,
	title={{Performance evaluation of RANSAC family}},
	author={Choi, Sunglok and Kim, Taemin and Yu, Wonpil},
	journal={J. of Computer Vision},
	volume={24},
	number={3},
	pages={271--300},
	year={1997}
}

@inproceedings{Schnabel07cgf-EfficientRANSAC,
	title={{Efficient RANSAC for point-cloud shape detection}},
	author={Schnabel, Ruwen and Wahl, Roland and Klein, Reinhard},
	booktitle={Computer Graphics Forum},
	pages={214--226},
	year={2007}
}

@inproceedings{Pan19robotbiomim-MultiViewBnB,
	title={{Multi-view global 2D-3D registration based on branch and bound algorithm}},
	author={Pan, Jin and Min, Zhe and Zhang, Ang and Ma, Han and Meng, Max Q-H},
	booktitle={Proc. IEEE Int. Conf. Robot. Biomim.},
	pages={3082--3087},
	year={2019}
}

@article{Dong17isprsremotesensing-GHICP,
	title={{A novel binary shape context for 3D local surface description}},
	author={Dong, Zhen and Yang, Bisheng and Liu, Yuan and Liang, Fuxun and Li, Bijun and Zang, Yufu},
	journal={ISPRS Journal of Photogrammetry and Remote Sensing},
	volume={130},
	pages={431--452},
	year={2017}
}

@article{Lei17tip-FastDescriptors,
	title={Fast descriptors and correspondence propagation for robust global point cloud registration},
	author={Lei, Huan and Jiang, Guang and Quan, Long},
	journal={IEEE Trans. Image Processing},
	volume={26},
	number={8},
	pages={3614--3623},
	year={2017}
}

@article{Papazov12ijrr-Rigid3DGeometryMatching,
	title={{Rigid 3D geometry matching for grasping of known objects in cluttered scenes}},
	author={Papazov, Chavdar and Haddadin, Sami and Parusel, Sven and Krieger, Kai and Burschka, Darius},
	journal=ijrr,
	volume={31},
	number={4},
	pages={538--553},
	year={2012}
}

@article{Geiger13ijrr-KITTI,
	title={{Vision meets robotics: The KITTI dataset}},
	author={Geiger, Andreas and Lenz, Philip and Stiller, Christoph and Urtasun, Raquel},
	journal=ijrr,
	volume={32},
	number={11},
	pages={1231--1237},
	year={2013}
}

@article{Poiesi22pami-GeDi,
  title={{Learning general and distinctive 3D local deep descriptors for point cloud registration}},
  author={Poiesi, Fabio and Boscaini, Davide},
  journal=pami,
  volume={45},
  number={3},
  pages={3979--3985},
  year={2022}
}

@inproceedings{Ao21cvpr-Spinnet,
  title={{SpinNet: Learning a general surface descriptor for 3D point cloud registration}},
  author={Ao, Sheng and Hu, Qingyong and Yang, Bo and Markham, Andrew and Guo, Yulan},
  booktitle=cvpr,
  pages={11753--11762},
  year={2021}
}

@inproceedings{Poiesi21icpr-DIP,
  title={Distinctive {3D} local deep descriptors},
  author={Poiesi, Fabio and Boscaini, Davide},
  booktitle=icpr,
  pages={5720--5727},
  year={2021}
}

@article{Chen22ar-Overlapnet,
  title={{OverlapNet: A siamese network for computing LiDAR scan similarity with applications to loop closing and localization}},
  author={Chen, Xieyuanli and L{\"a}be, Thomas and Milioto, Andres and R{\"o}hling, Timo and Behley, Jens and Stachniss, Cyrill},
  journal=ar,
  volume={46},
  number={1},
  pages={61--81},
  year={2022}
}

@article{Yin24ijcv-LiDARLocSurvey,
  title={A survey on global lidar localization: Challenges, advances and open problems},
  author={Yin, Huan and Xu, Xuecheng and Lu, Sha and Chen, Xieyuanli and Xiong, Rong and Shen, Shaojie and Stachniss, Cyrill and Wang, Yue},
  journal=ijcv,
  pages={1--33},
  year={2024}
}

@inproceedings{Ao23CVPR-BUFFER,
  title={{BUFFER: B}alancing accuracy, efficiency, and generalizability in point cloud registration},
  author={Ao, Sheng and Hu, Qingyong and Wang, Hanyun and Xu, Kai and Guo, Yulan},
  booktitle=cvpr,
  pages={1255--1264},
  year={2023}
}

@article{Sun22ral-TriVoC,
  title={{TriVoC: E}fficient voting-based consensus maximization for robust point cloud registration with extreme outlier ratios},
  author={Sun, Lei and Deng, Lu},
  journal=ral,
  volume={7},
  number={2},
  pages={4654--4661},
  year={2022}
}

@article{Lim21ral-Patchwork,
  title={{Patchwork: Concentric zone-based region-wise ground segmentation with ground likelihood estimation using a 3D LiDAR sensor}},
  author={Lim, Hyungtae and Oh, Minho and Myung, Hyun},
  journal=ral,
  volume={6},
  number={4},
  pages={6458--6465},
  year={2021}
}

@article{lee22arxiv-LearningReg,
  title={Learning to register unbalanced point pairs},
  author={Lee, Kanghee and Lee, Junha and Park, Jaesik},
  journal={arXiv preprint arXiv:2207.04221},
  year={2022}
}

@article{Besl92pami,
	Author = {P. J. Besl and N. D. McKay},
	Journal = PAMI,
	Number = 2,
	Title = {A method for registration of {3-D} shapes},
	Volume = 14,
	Year = 1992}

@inproceedings{Zhou16eccv-fastGlobalRegistration,
  title={Fast global registration},
  author={Zhou, Qian-Yi and Park, Jaesik and Koltun, Vladlen},
  booktitle={Proceedings of the European Conference on Computer Vision (ECCV)},
  pages={766--782},
  year={2016}
}

@article{Yu21nips-CofiNet,
  title={{CofiNet:} Reliable coarse-to-fine correspondences for robust pointcloud registration},
  author={Yu, Hao and Li, Fu and Saleh, Mahdi and Busam, Benjamin and Ilic, Slobodan},
  journal={Advances in Neural Information Processing Systems},
  volume={34},
  pages={23872--23884},
  year={2021}
}

@inproceedings{Yao24iccv-PARENet,
  title={PARE-Net: Position-Aware Rotation-Equivariant Networks for Robust Point Cloud Registration},
  author={Yao, Runzhao and Du, Shaoyi and Cui, Wenting and Tang, Canhui and Yang, Chengwu},
  booktitle={European Conference on Computer Vision},
  pages={287--303},
  year={2024},
  organization={Springer}
}

@article{Pomerleau12ijrr-ETH,
  title={Challenging data sets for point cloud registration algorithms},
  author={Pomerleau, Fran{\c{c}}ois and Liu, Ming and Colas, Francis and Siegwart, Roland},
  journal={The International Journal of Robotics Research},
  volume={31},
  number={14},
  pages={1705--1711},
  year={2012}
}

@inproceedings{Zhang23cvpr-MaxCliques,
  title={{3D} registration with maximal cliques},
  author={Zhang, Xiyu and Yang, Jiaqi and Zhang, Shikun and Zhang, Yanning},
  booktitle={Proceedings of the IEEE/CVF Conference on Computer Vision and Pattern Recognition},
  pages={17745--17754},
  year={2023}
}

@article{yang24tpami-MAC,
  title={{MAC: Maximal Cliques for 3D Registration}},
  author={Yang, Jiaqi and Zhang, Xiyu and Wang, Peng and Guo, Yulan and Sun, Kun and Wu, Qiao and Zhang, Shikun and Zhang, Yanning},
  journal={IEEE Transactions on Pattern Analysis and Machine Intelligence},
  year={2024}
}

@article{Fathian24ral-Clipperplus,
  title={{CLIPPER+: A Fast Maximal Clique Algorithm for Robust Global Registration}},
  author={Fathian, Kaveh and Summers, Tyler},
  journal={IEEE Robotics and Automation Letters},
  year={2024},
  publisher={IEEE}
}

@inproceedings{Zhang24cvpr-FastMAC,
  title={{FastMAC: Stochastic Spectral Sampling of Correspondence Graph}},
  author={Zhang, Yifei and Zhao, Hao and Li, Hongyang and Chen, Siheng},
  booktitle={Proceedings of the IEEE/CVF Conference on Computer Vision and Pattern Recognition},
  pages={17857--17867},
  year={2024}
}

@article{Shi24ral-RANSAC,
  title={{RANSAC back to SOTA: A two-stage consensus filtering for real-time 3D registration}},
  author={Shi, Pengcheng and Yan, Shaocheng and Xiao, Yilin and Liu, Xinyi and Zhang, Yongjun and Li, Jiayuan},
  journal={IEEE Robotics and Automation Letters},
  year={2024}
}

@inproceedings{Mu24cvpr-ColorPCR,
  title={{ColorPCR: C}olor Point Cloud Registration with Multi-Stage Geometric-Color Fusion},
  author={Mu, Juncheng and Bie, Lin and Du, Shaoyi and Gao, Yue},
  booktitle={Proceedings of the IEEE/CVF Conference on Computer Vision and Pattern Recognition},
  pages={21061--21070},
  year={2024}
}

@inproceedings{Chen24cpvr-DCATr,
  title={Dynamic Cues-Assisted Transformer for Robust Point Cloud Registration},
  author={Chen, Hong and Yan, Pei and Xiang, Sihe and Tan, Yihua},
  booktitle={Proceedings of the IEEE/CVF Conference on Computer Vision and Pattern Recognition},
  pages={21698--21707},
  year={2024}
}

@InProceedings{Huang24cvpr-Scalable,
    author    = {Huang, Tianyu and Peng, Liangzu and Vidal, Rene and Liu, Yun-Hui},
    title     = {{Scalable 3D Registration via Truncated Entry-wise Absolute Residuals}},
    booktitle = {Proceedings of the IEEE/CVF Conference on Computer Vision and Pattern Recognition (CVPR)},
    month     = {June},
    year      = {2024},
    pages     = {27477-27487}
}

@article{Liu24tgrs-DeepSemanticMatching,
  title={Deep semantic graph matching for large-scale outdoor point cloud registration},
  author={Liu, Shaocong and Wang, Tao and Zhang, Yan and Zhou, Ruqin and Li, Li and Dai, Chenguang and Zhang, Yongsheng and Wang, Longguang and Wang, Hanyun},
  journal={IEEE Transactions on Geoscience and Remote Sensing},
  year={2024}
}

@inproceedings{Chen24icra-Tree-based-Transformer,
  title={Fast and Robust Point Cloud Registration with Tree-based Transformer},
  author={Chen, Guangyan and Wang, Meiling and Yang, Yi and Yuan, Li and Yue, Yufeng},
  booktitle={2024 IEEE International Conference on Robotics and Automation (ICRA)},
  pages={773--780},
  year={2024}
}

@article{Liu24tpami-NCMNet,
  title={NCMNet: Neighbor Consistency Mining Network for Two-View Correspondence Pruning},
  author={Liu, Xin and Qin, Rong and Yan, Junchi and Yang, Jufeng},
  journal={IEEE Transactions on Pattern Analysis and Machine Intelligence},
  year={2024}
}

@article{Ao20pr-Repeatable,
  title={{A repeatable and robust local reference frame for 3D surface matching}},
  author={Ao, Sheng and Guo, Yulan and Tian, Jindong and Tian, Yong and Li, Dong},
  journal={Pattern Recognition},
  volume={100},
  pages={107186},
  year={2020}
}

@article{ao22tpami-YOTO,
  title={You only train once: Learning general and distinctive 3D local descriptors},
  author={Ao, Sheng and Guo, Yulan and Hu, Qingyong and Yang, Bo and Markham, Andrew and Chen, Zengping},
  journal={IEEE Transactions on Pattern Analysis and Machine Intelligence},
  volume={45},
  number={3},
  pages={3949--3967},
  year={2022}
}

@article{Ao20ietcv-SGHs,
  title={{SGHs for 3D local surface description}},
  author={Ao, Sheng and Guo, Yulan and Gu, Shangtai and Tian, Jindong and Li, Dong},
  journal={IET Computer Vision},
  volume={14},
  number={4},
  pages={154--161},
  year={2020}
}

@article{Chen21nips-OnlyTrainOnce,
  title={{Only train once: A one-shot neural network training and pruning framework}},
  author={Chen, Tianyi and Ji, Bo and Ding, Tianyu and Fang, Biyi and Wang, Guanyi and Zhu, Zhihui and Liang, Luming and Shi, Yixin and Yi, Sheng and Tu, Xiao},
  journal={Advances in Neural Information Processing Systems},
  volume={34},
  pages={19637--19651},
  year={2021}
}

@inproceedings{Dosovitskiy19iclr-YOTO,
  title={{You only train once: Loss-conditional training of deep networks}},
  author={Dosovitskiy, Alexey and Djolonga, Josip},
  booktitle={International conference on learning representations},
  year={2019}
}

@inproceedings{Qingqing22iros-TIERS,
  title={{Multi-modal {LiDAR} dataset for benchmarking general-purpose localization and mapping algorithms}},
  author={Qingqing, Li and Xianjia, Yu and Queralta, Jorge Pena and Westerlund, Tomi},
  booktitle=iros,
  pages={3837--3844},
  year={2022}
}

@article{Jung23ijrr-HeLiPR,
  title={{HeLiPR: Heterogeneous LiDAR dataset for inter-LiDAR place recognition under spatial and temporal variations}},
  author={Jung, Minwoo and Yang, Wooseong and Lee, Dongjae and Gil, Hyeonjae and Kim, Giseop and Kim, Ayoung},
  journal={The International Journal of Robotics Research},
  year={2023}
}

@inproceedings{Lim24iros-HeLiMOS,
  title={Helimos: A dataset for moving object segmentation in 3d point clouds from heterogeneous lidar sensors},
  author={Lim, Hyungtae and Jang, Seoyeon and Mersch, Benedikt and Behley, Jens and Myung, Hyun and Stachniss, Cyrill},
  booktitle={IEEE/RSJ International Conference on Intelligent Robots and Systems (IROS)},
  pages={14087--14094},
  year={2024}
}

@article{Qin23tpami-GeoTransformer,
  title={{GeoTransformer: Fast and robust point cloud registration with geometric transformer}},
  author={Qin, Zheng and Yu, Hao and Wang, Changjian and Guo, Yulan and Peng, Yuxing and Ilic, Slobodan and Hu, Dewen and Xu, Kai},
  journal={IEEE Transactions on Pattern Analysis and Machine Intelligence},
  volume={45},
  number={8},
  pages={9806--9821},
  year={2023}
}

@inproceedings{Thomas19iccv-kpconv,
  title={Kpconv: Flexible and deformable convolution for point clouds},
  author={Thomas, Hugues and Qi, Charles R and Deschaud, Jean-Emmanuel and Marcotegui, Beatriz and Goulette, Fran{\c{c}}ois and Guibas, Leonidas J},
  booktitle={Proceedings of the IEEE/CVF international conference on computer vision},
  pages={6411--6420},
  year={2019}
}

@inproceedings{Zhu21cvpr-Cylindrical3D,
  title={Cylindrical and asymmetrical 3d convolution networks for lidar segmentation},
  author={Zhu, Xinge and Zhou, Hui and Wang, Tai and Hong, Fangzhou and Ma, Yuexin and Li, Wei and Li, Hongsheng and Lin, Dahua},
  booktitle={Proceedings of the IEEE/CVF conference on computer vision and pattern recognition},
  pages={9939--9948},
  year={2021}
}

@inproceedings{Wu24cvpr-PointTransformerV3,
  title={{Point Transformer V3: Simpler Faster Stronger}},
  author={Wu, Xiaoyang and Jiang, Li and Wang, Peng-Shuai and Liu, Zhijian and Liu, Xihui and Qiao, Yu and Ouyang, Wanli and He, Tong and Zhao, Hengshuang},
  booktitle={Proceedings of the IEEE/CVF Conference on Computer Vision and Pattern Recognition},
  pages={4840--4851},
  year={2024}
}

@article{Wu22nips-PointTransformerV2,
  title={Point transformer v2: Grouped vector attention and partition-based pooling},
  author={Wu, Xiaoyang and Lao, Yixing and Jiang, Li and Liu, Xihui and Zhao, Hengshuang},
  journal={Advances in Neural Information Processing Systems},
  volume={35},
  pages={33330--33342},
  year={2022}
}

@inproceedings{choy19cvpr-4DSTConv,
  title={{4D Spatio-Temporal Convnets: Minkowski convolutional neural networks}},
  author={Choy, Christopher and Gwak, JunYoung and Savarese, Silvio},
  booktitle=cvpr,
  pages={3075--3084},
  year={2019}
}

@inproceedings{Yeshwanth23iccv-Scannet++,
  title={Scannet++: A high-fidelity dataset of 3d indoor scenes},
  author={Yeshwanth, Chandan and Liu, Yueh-Cheng and Nie{\ss}ner, Matthias and Dai, Angela},
  booktitle={Proceedings of the IEEE/CVF International Conference on Computer Vision},
  pages={12--22},
  year={2023}
}

@inproceedings{Ramezani20iros-NewerCollege,
  title={The newer college dataset: Handheld lidar, inertial and vision with ground truth},
  author={Ramezani, Milad and Wang, Yiduo and Camurri, Marco and Wisth, David and Mattamala, Matias and Fallon, Maurice},
  booktitle={2020 IEEE/RSJ International Conference on Intelligent Robots and Systems (IROS)},
  pages={4353--4360},
  year={2020},
  organization={IEEE}
}

@inproceedings{Sun20cvpr-WaymoDataset,
	title={Scalability in perception for autonomous driving: Waymo open dataset},
	author={Sun, Pei and Kretzschmar, Henrik and Dotiwalla, Xerxes and Chouard, Aurelien and Patnaik, Vijaysai and Tsui, Paul and Guo, James and Zhou, Yin and Chai, Yuning and Caine, Benjamin and others},
	booktitle={Proceedings of the IEEE/CVF Conference on Computer Vision and Pattern Recognition},
	pages={2446--2454},
	year={2020}
}

@article{Kingma2014arxiv-Adam,
  title={{Adam: A method for stochastic optimization}},
  author={Kingma, Diederik P},
  journal={arXiv preprint arXiv:1412.6980},
  year={2014}
}

@article{Lim25icra-KISSMatcher,
  title={{KISS-Matcher: Fast and Robust Point Cloud Registration Revisited}},
  author={Lim, Hyungtae and Kim, Daebeom and Shin, Gunhee and Shi, Jingnan and Vizzo, Ignacio and Myung, Hyun and Park, Jaesik and Carlone, Luca},
  journal=icra,
  year={2025},
  note={Accepted. To appear.}
}

@inproceedings{Harris88avc-HarrisCorner,
  title={A combined corner and edge detector},
  author={Harris, Chris and Stephens, Mike and others},
  booktitle={Alvey vision conference},
  pages={10--5244},
  year={1988}
}

@article{Alexiou24jivp-PointPCA,
  title={{PointPCA: Point cloud objective quality assessment using PCA-based descriptors}},
  author={Alexiou, Evangelos and Zhou, Xuemei and Viola, Irene and Cesar, Pablo},
  journal={EURASIP Journal on Image and Video Processing},
  volume={2024},
  number={1},
  pages={20},
  year={2024}
}

@misc{Barron19-generaladaptiverobustloss,
      title={A General and Adaptive Robust Loss Function}, 
      author={Jonathan T. Barron},
      year={2019},
      eprint={1701.03077},
      archivePrefix={arXiv},
      primaryClass={cs.CV},
      url={https://arxiv.org/abs/1701.03077}, 
}

@article{Zhang24tpami-AcceleratingGloballyCM,
  title={Accelerating globally optimal consensus maximization in geometric vision},
  author={Zhang, Xinyue and Peng, Liangzu and Xu, Wanting and Kneip, Laurent},
  journal={IEEE Transactions on Pattern Analysis and Machine Intelligence},
  volume={46},
  number={6},
  pages={4280--4297},
  year={2024},
  publisher={IEEE}
}

@article{Hansen21remotesensing-Classification,
  title={{Classification of boulders in coastal environments using random forest machine learning on topo-bathymetric LiDAR data}},
  author={Hansen, Signe Schilling and Ernstsen, Verner Brandbyge and Andersen, Mikkel Skovgaard and Al-Hamdani, Zyad and Baran, Ramona and Niederwieser, Manfred and Steinbacher, Frank and Kroon, Aart},
  journal={Remote Sensing},
  volume={13},
  number={20},
  pages={4101},
  year={2021},
  publisher={MDPI}
}

@article{Mebius07arxiv-Derivation,
  title={{Derivation of the Euler-Rodrigues formula for three-dimensional rotations from the general formula for four-dimensional rotations}},
  author={Mebius, Johan Ernest},
  journal={arXiv preprint math/0701759},
  year={2007}
}

@article{Zhang97ivc-Parameter,
  title={{Parameter estimation techniques: A tutorial with application to conic fitting}},
  author={Zhang, Zhengyou},
  journal={Image and vision Computing},
  volume={15},
  number={1},
  pages={59--76},
  year={1997},
  publisher={Elsevier}
}

@inproceedings{Zhao24cvpr-CorrespondenceFree,
  title={{Correspondence-free non-rigid point set registration using unsupervised clustering analysis}},
  author={Zhao, Mingyang and Jiang, Jingen and Ma, Lei and Xin, Shiqing and Meng, Gaofeng and Yan, Dong-Ming},
  booktitle={Proceedings of the IEEE/CVF Conference on Computer Vision and Pattern Recognition},
  pages={21199--21208},
  year={2024}
}

@inproceedings{Yuan24cvpr-InlierConfidence,
  title={Inlier confidence calibration for point cloud registration},
  author={Yuan, Yongzhe and Wu, Yue and Fan, Xiaolong and Gong, Maoguo and Miao, Qiguang and Ma, Wenping},
  booktitle={Proceedings of the IEEE/CVF Conference on Computer Vision and Pattern Recognition},
  pages={5312--5321},
  year={2024}
}

@inproceedings{Liu24cvpr-Extend,
  title={Extend your own correspondences: Unsupervised distant point cloud registration by progressive distance extension},
  author={Liu, Quan and Zhu, Hongzi and Wang, Zhenxi and Zhou, Yunsong and Chang, Shan and Guo, Minyi},
  booktitle={Proceedings of the IEEE/CVF Conference on Computer Vision and Pattern Recognition},
  pages={20816--20826},
  year={2024}
}

@InProceedings{Yu24cvpr-InstanceAware,
  author    = {Yu, Zhiyuan and Qin, Zheng and Zheng, Lintao and Xu, Kai},
  title     = {Learning Instance-Aware Correspondences for Robust Multi-Instance Point Cloud Registration in Cluttered Scenes},
  booktitle = {Proceedings of the IEEE/CVF Conference on Computer Vision and Pattern Recognition (CVPR)},
  month     = {June},
  year      = {2024},
  pages     = {19605-19614}
}

@inproceedings{Ryu23cvpr-InstantDomainAugmentation,
  title={Instant domain augmentation for lidar semantic segmentation},
  author={Ryu, Kwonyoung and Hwang, Soonmin and Park, Jaesik},
  booktitle={Proceedings of the IEEE/CVF Conference on Computer Vision and Pattern Recognition},
  pages={9350--9360},
  year={2023}
}

@string{ral = {{IEEE} Robotics and Automation Letters ({RA-L})}}

@article{Yang20tro-teaser,
	title={{TEASER: Fast and Certifiable Point Cloud Registration}},
	author={H. Yang and J. Shi and L. Carlone},
	journal=tro,
	volume = 37,
	number = 2,
	pages = {314--333},	
	Year = {2020}
}

@article{Yang20ral-GNC,
	Author = {H. Yang and P. Antonante and V. Tzoumas and L. Carlone},
	FullAuthor = {Heng Yang, Pasquale Antonante, Vasileios Tzoumas, Luca Carlone},
	Title = {Graduated Non-Convexity for Robust Spatial Perception: From Non-Minimal Solvers to Global Outlier Rejection},
	Volume  = 5,
	Number = 2,
	Pages = {1127--1134},
	journal=ral,
	year={2020}
}

@InProceedings{Tian23iros-KimeraMultiExperiments,
  author = {Y. Tian and Y. Chang and L. Quang and A. Schang and C. Nieto-Granda and J.P. How and L. Carlone},
  title = {Resilient and Distributed Multi-Robot Visual {SLAM}: Datasets, Experiments, and Lessons Learned},
  booktitle = iros,
  funding = {DCIST, MathWorks},
  Year = {2023}
}

@inproceedings{Yang19rss-teaser,
	Author = {H. Yang and L. Carlone},
	Title = {A Polynomial-time Solution for Robust Registration with Extreme Outlier Rates},
	booktitle= rss,
	Year = 2019}

@inproceedings{Yin23icra-Segregator,
  title={{Segregator: Global point cloud registration with semantic and geometric cues}},
  author={Yin, Pengyu and Yuan, Shenghai and Cao, Haozhi and Ji, Xingyu and Zhang, Shuyang and Xie, Lihua},
  booktitle=icra,
  pages={2848--2854},
  year={2023}
}
}

\clearpage
\setcounter{page}{1}
\setcounter{section}{0}
\setcounter{figure}{0}
\setcounter{table}{0}

\renewcommand{\thesection}{\Alph{section}}
\renewcommand{\theequation}{A\arabic{equation}}
\renewcommand{\thetheorem}{A\arabic{theorem}}
\renewcommand{\thefigure}{A\arabic{figure}}
\renewcommand{\thetable}{A\arabic{table}}

\section*{Appendix}

\section{Parameters setup and pseudo code}\label{sec:parameters}

Here, we summarize the parameter values in \Cref{table:bufferx_params}.
While some values were set empirically based on prior work, others were tuned to maintain generalizability without dataset-specific fine-tuning.
 In addition, we also provide pseudo code in \Cref{alg:buffer_x} to outline the core steps of our approach, making it easier to understand the implementation details and facilitate reproducibility.

\section{Detailed experimental setups and datasets}\label{app:dataset_details}
Here, we elaborate on the experimental setups that are briefly outlined in \Cref{tab:overview_of_datasets}.
For widely used datasets, we followed the existing training and testing protocols.
The details of the datasets employed in our generalizability benchmark are described as follows.

\begin{itemize}
\item{\ThreeDMatch} follows conventional protocol proposed by Zeng~\textit{et al.}\cite{Zeng17cvpr-3dmatch}.

\item{\ThreeDLoMatch} follows conventional protocol proposed by Huang~\textit{et al.}\cite{Huang21cvpr-PREDATORRegistration}.
This dataset is derived from {\ThreeDMatch} by selectively extracting pairs with low overlap (\ie~10–30\%), allowing for the evaluation of robustness to low-overlap scenarios.
 
\item{\ScanNetppF} 
  is from \texttt{ScanNet++}~\cite{Yeshwanth23iccv-Scannet++}. 
For each sequence, a pre-merged PLY file representing the entire space is provided, along with the poses of the stationary FARO LiDAR sensor used to measure it.
That is, the dataset does not provide individual scans.
For that reason, using the given scanner poses, we generate per-scanner point clouds by sampling along raycasting paths from each position within the full map, simulating virtual scans.
The sampling process respects the scanner's horizontal angular resolution as well as its vertical resolution between consecutive rays~\cite{Ryu23cvpr-InstantDomainAugmentation}, ensuring a realistic approximation of the original scans.
For each ray, the nearest intersecting point in the merged point cloud is selected to simulate the original scan.

\item{\ScanNetppi} is also from \texttt{ScanNet++}~\cite{Yeshwanth23iccv-Scannet++}.
  There are depth images captured using a LiDAR sensor attached to an iPhone 13 Pro.
  These depth images were converted into point clouds using the toolbox provided by 3DMatch~\cite{Zeng17cvpr-3dmatch}.
  To generate dense point cloud fragments, 50 consecutive frames were accumulated.
  Finally, pairs with an overlap ratio of at least 0.4 (\ie 40\% overlap between two fragments) were selected as the final test pairs.

\item{\TIERS} consists of \texttt{Indoor06}, \texttt{Indoor08}, \texttt{Indoor09}, \texttt{Indoor10}, and \texttt{Indoor11} sequences in the TIERS dataset~\cite{Qingqing22iros-TIERS}.
To reduce redundant test pairs, we exclude \texttt{Indoor07} because it was acquired in the same room as \texttt{Indoor06}. 
\begingroup
\vspace{-0.5em}
\begin{table}[h!]
	\setlength{\tabcolsep}{3pt}
	{\scriptsize
		\begin{tabular}{ccc}
			\toprule \midrule
			Param. & Description & Value  \\ \midrule
      $\kappa_\text{spheric}$ & Coefficient for voxel size when sphericity is high & 0.10 \\
      $\kappa_\text{disc} $   & Coefficient for voxel size when sphericity is low  & 0.15 \\
      $\tau_v $               & Threshold in Eq.~\Cref{eq:voxel_size}  & 0.05 \\
      $\tau_l $         & Threshold for the local ($l$) search radius  & 0.005 \\
      $\tau_m $         & Threshold for the middle ($m$) search radius  & 0.02 \\
      $\tau_g $         & Threshold for the global ($g$) search radius  & 0.05 \\
      $\delta_v $         & Sampling ratio for sphericity-based voxelization & 10\% \\
      $N_r $         & Number of sampling points for radius estimation & 2,000 \\
      $r_\text{max}$          & Maximum radius  & 5.0\,m \\ \midrule
      $N_\text{FPS} $         & Number of sampled points by FPS & 1,500 \\
      $N_\text{patch} $         & Maximum number of points in each patch  & 512 \\
      $\delta$  & Truncation threshold for Huber loss & 1.0 \\ 
      $H$  & Height of the cylindrical map in Mini-SpinNet & 7 \\ 
      $W$  & Sector size of the cylindrical map in Mini-SpinNet & 20 \\ 
      $D$  & Feature dimension of the cylindrical map in Mini-SpinNet & 32 \\ \midrule \bottomrule
	\end{tabular}
	}
  \vspace{-0.8em}
  \captionsetup{font=small}
	\centering
	\caption{Parameters of each module in our BUFFER-X. Note that with this parameter setup, our approach operates in our generalizability benchmark in an out-of-the-box manner without any human intervention.}
	\label{table:bufferx_params}
\end{table}
\endgroup
 \vspace{-2.0em}
\newcommand{\annot}[1]{\textcolor{gray!80}{\% #1}}

\begin{algorithm}[h!]
{\footnotesize
\SetAlgoLined
\textbf{Input:} \ Source cloud $\srccloud$ and target cloud $\tgtcloud$; User-defined parameters $\tau_v$, $\delta_v$, $\delta_r$, $[\tau_l, \tau_m, \tau_g]$, and $N_\text{FPS}$ \\
\textbf{Output:} 3D inliers $\mathcal{I}$ \\
$\mathcal{P}_r \leftarrow \texttt{select\_larger\_cloud}(\srccloud, \tgtcloud)$  \\
$\mathcal{P}_\text{sampled} = \texttt{sample}(\mathcal{P}_r, \delta_v)$ \annot{Sample $\delta_v$\% of cloud points} \\
\annot{Step 1. Geometric bootstrapping} \\
$v = \texttt{calc\_voxel\_size}(\mathcal{P}_\text{sampled}, \tau_v)$ \annot{See Eq.~\Cref{eq:voxel_size}} \\
$\mathcal{P} \leftarrow f_v(\mathcal{P}), \mathcal{Q} \leftarrow f_v(\mathcal{Q})$
\annot{Downsample the point clouds} \\
$\mathcal{P}_r \leftarrow \texttt{select\_larger\_cloud}(\srccloud, \tgtcloud)$  \\
$\mathcal{R} = \texttt{estimate\_radii}(\texttt{sample}(\mathcal{P}_r, N_r), [\tau_l, \tau_m, \tau_g]),$ \\
\qquad where $\mathcal{R} = [r_l, r_m, r_g] $ \annot{See Eq.~\Cref{eq:density_radius}} \\
\annot{Step 2. Multi-scale patch embedder} \\
$\mathcal{M}^\mathcal{P} = \varnothing, \mathcal{M}^\mathcal{Q} = \varnothing$ \annot{Containers of embedding output} \\
\For{$r_\xi$ \normalfont{in} $\mathcal{R}$}{
    $\mathcal{P}_\xi = \texttt{farthest\_point\_sampling}(\mathcal{P}, N_\text{FPS})$  \\
    $\mathcal{Q}_\xi = \texttt{farthest\_point\_sampling}(\mathcal{Q}, N_\text{FPS})$ \\
    $\mathcal{F}^\mathcal{P}_\xi, \mathcal{C}^\mathcal{P}_\xi = \texttt{MiniSpinNet}(\mathcal{P}_\xi, \mathcal{P}, r_\xi)$ \\
    $\mathcal{F}^\mathcal{Q}_\xi, \mathcal{C}^\mathcal{Q}_\xi = \texttt{MiniSpinNet}(\mathcal{Q}_\xi, \mathcal{Q}, r_\xi)$ \\
    $\mathcal{M}^\mathcal{P}.\texttt{append}((\mathcal{P}_\xi, \mathcal{F}^\mathcal{P}_\xi, \mathcal{C}^\mathcal{P}_\xi))$ \\
    $\mathcal{M}^\mathcal{Q}.\texttt{append}((\mathcal{Q}_\xi, \mathcal{F}^\mathcal{Q}_\xi, \mathcal{C}^\mathcal{Q}_\xi))$ \\
}
\annot{Step 3. Hierarchical inlier search} \\
$\mathcal{D} = \varnothing, \mathcal{T} = \varnothing$ \\ 
\For{$i$ \normalfont{in} \texttt{range}(\texttt{size}($\mathcal{M}^\mathcal{P}$))}{
    $(\mathcal{P}_\xi, \mathcal{F}^\mathcal{P}_\xi, \mathcal{C}^\mathcal{P}_\xi) = \mathcal{M}^\mathcal{P}[i]$ \\
    $(\mathcal{Q}_\xi, \mathcal{F}^\mathcal{Q}_\xi, \mathcal{C}^\mathcal{Q}_\xi) = \mathcal{M}^\mathcal{Q}[i]$ \\
    \annot{Step 3-1. Nearest neighbor-based intra-scale matching}\\ 
    $\mathcal{A}_\xi = \texttt{mutual\_matching}(\mathcal{F}^\mathcal{P}_\xi, \mathcal{F}^\mathcal{Q}_\xi)$ \\
    $ (\widehat{\mathcal{P}}_\xi, \widehat{\mathcal{Q}}_\xi, \widehat{\mathcal{C}}^{\mathcal{P}}_\xi, \widehat{\mathcal{C}}^{\mathcal{Q}}_\xi) = \texttt{filter}(\mathcal{M}^\mathcal{P}[i], \mathcal{M}^\mathcal{Q}[i], \mathcal{A}_\xi)$ \\
    \annot{Step 3-2. Pairwise transformation estimation}\\
    $\mathcal{T}_\xi = \texttt{calc\_pairwise\_R\_and\_t}(\widehat{\mathcal{C}}^{\mathcal{P}}_\xi, \widehat{\mathcal{C}}^{\mathcal{Q}}_\xi)$ \\
    $\mathcal{D}.\texttt{append}(\widehat{\mathcal{P}}_\xi, \widehat{\mathcal{Q}}_\xi))$, $\mathcal{T}.\texttt{append}(\mathcal{T}_\xi)$\\
}
\annot{Step 3-3. Cross-scale consensus maximization}\\
$\mathcal{I} = \texttt{consensus\_maximization}(\mathcal{D}, \mathcal{T})$ \annot{See Eq.~\Cref{eq:consensus_max}} \\
}
    \caption{BUFFER-X pipeline\label{alg:buffer_x}}
\end{algorithm}

 \vspace{-1em}

In particular, we used data obtained from the Velodyne VLP-16, Ouster OS1-64, and Ouster OS0-128. 
While more sensors are available, we observed that point clouds from the Livox Horizon and Livox Avia contain too few points. 
Specifically, when a human surveyor moves close to a wall in an indoor environment, the narrow field of view of these feed-forwarding LiDAR sensors causes only partial wall surfaces to be captured, unlike omnidirectional LiDAR sensors. 
Additionally, during rotation, there exist segments where the overlap between consecutive scans becomes completely zero, making them unsuitable for the registration problem.

\newcommand{\solidstate}{$^\text{\ding{93}}$}
\begin{table*}[t!]
  \centering
  \setlength{\tabcolsep}{2pt}
  {\scriptsize
  \begin{tabular}{c|cccc}
    \toprule \midrule
    Dataset/Sequence Name & \ThreeDMatch, \ThreeDLoMatch  & \ScanNetppi &  \ScanNetppF & \TIERS \\ \midrule
    Environment & Room & Room, Interior & Room, Interior & Room, Campus Interior  \\ \midrule 
    Acquisition & Handheld & Handheld & Tripod & Cart \\ \midrule 
     Acquisition Site & N/A & N/A & N/A & Univ. Turku, Turku, Finland \\ \midrule 
    Measurement Type & Structured Light & Laser & Laser & Laser  \\ \midrule
    Employed Sensor(s) & 
    \begin{tabular}{@{}c@{}}Microsoft Kinect,\\ Structure Sensor, \\ Asus Xtion Pro Live, \\ and Intel RealSense \end{tabular}
     & Iphone RGB-D Sensor\solidstate & Faro Focus Premium &  \begin{tabular}{@{}c@{}}Velodyne VLP-16,\\ Ouster OS1-64, \\ Ouster OS0-128 \end{tabular} \\ \midrule
    Approx. Range [m] & 3.5 & 3.5 & 7.0 & 110 \\ \midrule 
    \# of Points/Frame & 336,274 & 444,876 &  1,381,013 & 34,777  \\ \midrule 
    \# of Test Pairs & 1,623 \text{/} 1,781 & 2,074 & 2,016 & 870 \\ \midrule
    \ Success Criteria &  
    \begin{tabular}{@{}c@{}}Follow Zeng~\textit{et al.}~\cite{Zeng17cvpr-3dmatch}'s \\ criteria \\ (Point-wise RMSE $2$\,m) \end{tabular}
      & RTE $0.3$\,m, RRE $15^\circ$ & RTE $0.3$\,m, RRE $15^\circ$ & RTE $2.0$\,m, RRE $5^\circ$ \\ 
  \end{tabular}  
  \begin{tabular}{c|cccccc}
    \toprule \midrule
    Dataset/Sequence Name & \WOD & \KITTI & \ETH &  \KAIST & \MIT & \Oxford \\ \midrule
    Environment & Urban & Urban & Forest & Campus & Campus & Campus \\ \midrule 
    Acquisition & Vehicle & Vehicle & Tripod & Vehicle & Wheeled & Quadruped / Handheld \\ \midrule 
    Acquisition Site & Phoenix, USA & Karlsruhe, Germany & N/A & 
    \begin{tabular}{@{}c@{}}KAIST campus, Daejeon,\\ South Korea \end{tabular}  &  
    \begin{tabular}{@{}c@{}}MIT campus, \\ Cambridge, USA \end{tabular}
    & 
    \begin{tabular}{@{}c@{}}Oxford campus, \\ London, UK  \end{tabular} \\ \midrule 
    Measurement Type &Laser & Laser & Laser & Laser & Laser & Laser \\ \midrule
    Employed Sensor(s) & 
    Laser Bear Honeycomb\solidstate & Velodyne HDL-64E & Rotating Hokuyo UTM-30LX  & \begin{tabular}{@{}c@{}}Livox Avia\solidstate,\\ Aeva Aeries~\rom{2}\solidstate, \\ Ouster2-128 \end{tabular} & Velodyne VLP-16 & Ouster OS-1 64  \\ \midrule
    Approx. Scale [m] & 120 & 80 & 85 & 240 & 100 & 100 \\ \midrule 
    \# Points/Frame & 143,960 & 123,518 & 96,884 & 68,790 & 24,792 & 55,914 \\ \midrule 
    \# of Test Pairs & 130 & 555 & 713 & 1,991 & 230 & 301  \\ \midrule
    \ Success Criteria & RTE $2.0$\,m, RRE $5^\circ$ & RTE $2.0$\,m, RRE $5^\circ$ &  RTE $0.3$\,m, RRE $2^\circ$ & RTE $2.0$\,m, RRE $5^\circ$ & RTE $2.0$\,m, RRE $5^\circ$ & RTE $2.0$\,m, RRE $5^\circ$ \\ 
   \midrule \bottomrule
  \end{tabular}  
  }
 	\captionsetup{font=small}
  \caption{Overview of the datasets used in our experiments, including their environments, acquisition sites, measurement types, employed sensors, approximate ranges or scales, and the number of test pairs.
  The datasets cover a variety of indoor and outdoor environments, spanning different geographic and cultural regions.
  The superscript \ding{93} highlights solid-state LiDAR sensors to especially emphasize that our evaluation is not limited to conventional omnidirectional spinning LiDAR sensors, but contains different scanning patterns.
  \vspace{-2mm}
  }
  \label{tab:overview_of_datasets}
\end{table*}
\item{\WOD} follows protocol proposed by Liu~\textit{et al.}\cite{Liu24cvpr-Extend}. This dataset is from the Waymo Open Dataset\cite{Sun20cvpr-WaymoDataset} Perception dataset by extracting LiDAR sequences and corresponding pose files, which are then converted into the KITTI format to ensure compatibility with standard benchmarking pipelines. We set $\tau_\text{dist} = 10\,\text{m}$

\item{\KITTI} follows conventional protocol proposed by Yew~\textit{et al.}\cite{Yew18eccv-3dfeatnet}. In test scenes, \texttt{08}, \texttt{09}, and \texttt{10} sequences are employed. Originally, $\tau_\text{dist} = 10\,\text{m}$.

\item{\ETH} is from the \texttt{gazebo\_summer}, \texttt{gazebo\_winter}, \texttt{wood\_autmn}, and \texttt{wood\_summer} sequences of the dataset proposed by Pomerleau~\textit{et al.}~\cite{Pomerleau12ijrr-ETH}.
The original dataset contains a wider various of scenes; however, following the existing protocol proposed by Ao~\textit{et al.}~\cite{Ao23CVPR-BUFFER}, we used four sequences.

\item{\KAIST} is from the \texttt{KAIST05} sequence of the HeLiPR dataset~\cite{Jung23ijrr-HeLiPR}. 
  Originally, HeLiPR contains multiple sequences, but each sequence in the HeLiPR is much longer than those in {\MIT} and {\Oxford}, resulting in many more test pairs compared to other campus scenes~(\ie 1991 vs. 230 or 301 in \Cref{tab:overview_of_datasets}).
  For this reason,  we balanced the datasets by using only one sequence. We set $\tau_\text{dist} = 10\,\text{m}$.

Instead of emphasizing that HeLiPR is a heterogeneous LiDAR sensor dataset, we refer to a subset of it as {\KAIST} in our paper to highlight that our dataset was curated with consideration for geographic and cultural environments.
Similarly, we use {\MIT} (from the Kimera-Multi dataset~\cite{Tian23iros-KimeraMultiExperiments}) and {\Oxford}~(from the NewerCollege dataset~\cite{Ramezani20iros-NewerCollege}) to emphasize the institutions for the same reason.
Further details are provided in Appendix~\ref{app:dataset_rationale}.
\item{\MIT} is from \texttt{10\_14\_acl\_jackal} sequence from the Kimera-Multi dataset~\cite{Tian23iros-KimeraMultiExperiments}, which is a multi-robot multi-session SLAM dataset. 
We could have used more scenes, but we chose to use only one sequence to a)~match \Oxford's frame count as closely as possible~(\ie 230 vs. 301 in \Cref{tab:overview_of_datasets}) and b)~reduce the redundant test pairs, as multi-robot SLAM datasets often observe the same space multiple times.

Note that the data was taken from a wheeled robot and acquired using a Velodyne VLP 16 sensor, so we observed that registration almost fails due to point cloud sparsity when $\tau_\text{dist} = 10\,\text{m}$ like \KITTI, \WOD, and \KAIST. 
Therefore, we decided to set $\tau_\text{dist} = 5\,\text{m}$.

\item{\Oxford} is from the \texttt{01}, \texttt{05}, and \texttt{07} sequences of the NewerCollege dataset~\cite{Ramezani20iros-NewerCollege}. 
  An interesting aspect is that \texttt{01} and \texttt{07} were acquired by handheld setup, while the \texttt{05} sequence was acquired using a quadruped robot.
  As shown in \Cref{fig:campus_scale}, the campus scale was relatively small, so if we use only a single sequence, it only generates few test pairs.
  For that reason, to ensure at least a similar number of test pairs as \MIT, unlike {\KAIST} and {\MIT}, we used three sequences and set  $\tau_\text{dist}=5\,\text{m}$.
\end{itemize} 

\begin{figure}[t!]
 	\centering
 	\begin{subfigure}[b]{0.45\textwidth}
 		\includegraphics[width=1.0\textwidth]{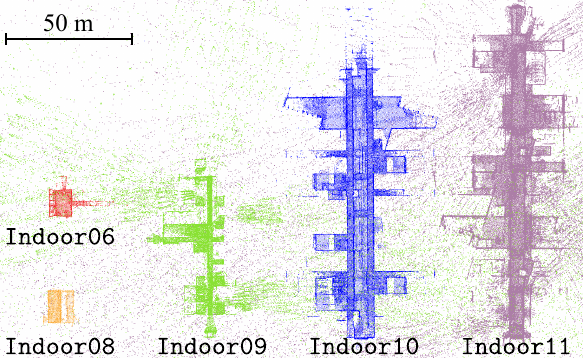}
        \caption{}
 	\end{subfigure}
        \begin{subfigure}[b]{0.45\textwidth}
 		\includegraphics[width=1.0\textwidth]{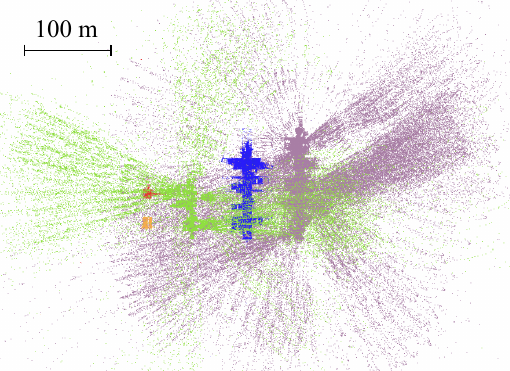}
        \caption{}
 	\end{subfigure}
 	\captionsetup{font=small}
 	\caption{Visualized map clouds of the {\TIERS} dataset in our experiments. (a)~Different scales of the sequences used in our experiments. While each scan is utilized for our evaluation, we build and then visualize map clouds using LiDAR point cloud scans and corresponding poses to illustrate the different scales of the surroundings. 
    (b)~A zoomed-out visualization of (a), highlighting the noisy characteristics inherent to indoor LiDAR scanning. Unlike RGB-D sensors, LiDAR sensors emit laser rays and calculate distance by measuring time-of-flight. However, in indoor environments, materials with high reflectivity, such as marble and glass, are highly likely to cause specular and multiple reflections. As a result, these reflections increase the time-of-flight, leading to noisy and incorrect range measurements.} 	
    \label{fig:tiers_scene_viz}
    \vspace{-4mm}
\end{figure}

\section{Rationale behind dataset selection}\label{app:dataset_rationale}

Here, we explain our detailed rationale for why we chose the aforementioned data as our comprehensive dataset. 

\vspace{2mm}
\noindent\textbf{Variation in environmental scales.} First, we want to construct sufficient domain generalization between indoor and outdoor scenes.
For existing approaches, only {\ThreeDMatch} and {\KITTI} were used for indoor-to-outdoor (and vice versa) evaluations.
Unfortunately, these experimental setups could not assess how well the methods would perform indoors when using a non-RGB-D camera.
Specifically, the maximum range was set to 5\,m for {\ThreeDMatch} and 80–100\,m for \KITTI.

However, even in indoor environments, LiDAR sensors can measure much farther, as presented in \Cref{fig:key_elements}(b) and \Cref{fig:tiers_scene_viz}.
Therefore, we aimed to challenge the bias that indoor and outdoor settings can be strictly distinguished by maximum range and decided to include \TIERS.
That is, in \TIERS, although all sequences were collected within the same indoor building, they were captured in diverse environments such as rooms, classrooms, and hallways.
As a result, the scale of the surroundings captured by the sensor varies significantly.
This was intended to evaluate whether methods could still perform well on indoor scenes beyond the room level, as opposed to the conventional settings in \ThreeDMatch.
Moreover, in the \TIERS, when data is acquired in a real corridor environment using LiDAR, the point cloud becomes noisy due to the diffuse reflection of laser rays; see \Cref{fig:tiers_scene_viz}(b).

\begin{figure}[t!]
 	\centering
 	\begin{subfigure}[b]{0.45\textwidth}
 		\includegraphics[width=1.0\textwidth]{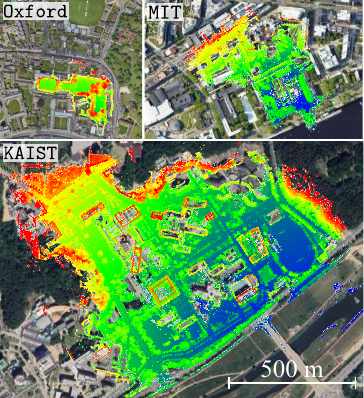}
 	\end{subfigure}        
 	\captionsetup{font=small}
 	\caption{Scale comparison of three sequences: {\Oxford} from the NewerCollege~\cite{Ramezani20iros-NewerCollege}, {\MIT} from the Kimera-Multi~\cite{Tian23iros-KimeraMultiExperiments}, and {\KAIST} from the HeLiPR dataset~\cite{Jung23ijrr-HeLiPR} at the same scale~(\ie~500\,m).
  Note that although these sequences fall under the same campus category, their scales differ. For clarity, the map clouds are visualized with respect to their $z$ values.} 	
    \label{fig:campus_scale}
    \vspace{-4mm}
\end{figure}

Likewise, one might think that the scales of campus environments are similar; however, as shown in \Cref{fig:campus_scale}, even within the campus category, variations in campus size can lead to differences in the distribution of LiDAR sensor data.
Therefore, by incorporating datasets with varying environmental scales, we aim to ensure that our generalizability benchmark includes the full spectrum of scale variations, enabling a more comprehensive assessment of generalization across different settings.

\begin{figure}[h!]
 	\centering
 	\begin{subfigure}[b]{0.45\textwidth}
 		\includegraphics[width=1.0\textwidth]{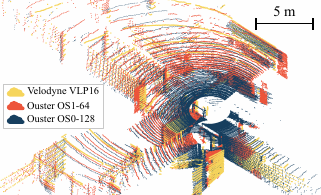}
        \caption{}
        \end{subfigure}
        \begin{subfigure}[b]{0.45\textwidth}
 		\includegraphics[width=1.0\textwidth]{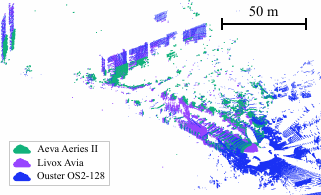}
        \caption{}
 	\end{subfigure}
 	\captionsetup{font=small}
 	\caption{Examples of visualized LiDAR scans from different LiDAR sensors in (a)~the \texttt{Indoor10} of {\TIERS} dataset~\cite{Qingqing22iros-TIERS} and (b)~\texttt{KAIST05} sequence of the HeLiPR dataset~\cite{Jung23ijrr-HeLiPR}. Note that even in the same environment, differences in the number of LiDAR rays and field of view result in point clouds with different patterns.} 	
    \label{fig:hetero_lidars}
    \vspace{-4mm}
\end{figure}

\vspace{2mm}
\noindent\textbf{Different scanning patterns with different sensor types.} In addition to using the {\TIERS} for the reasons mentioned above, we also aimed to evaluate whether the same space remains robust to different scanning patterns.
To this end, we employed {\KAIST} and \TIERS. As shown in \Cref{fig:hetero_lidars}, even when the same space is captured, variations in the number of laser rays and sensor patterns result in different representations.

\vspace{2mm}
\noindent\textbf{Acquisition setups.} \, We also considered that acquisition setups vary in multiple ways.
A common bias in previous evaluations in the state-of-the-art approaches is that indoor scanning is performed using handheld devices, whereas outdoor scanning is conducted using vehicles~(\ie the assumption that indoor scanning is performed using a handheld setup, while outdoor scanning is conducted using a vehicle).
Thus, we aimed to evaluate whether registration remains robust to different acquisition setups to challenge this assumption.

For this reason, we included \TIERS, which was acquired using a sensor cart, and {\Oxford}, which was acquired using both a handheld device and a quadruped robot.
Additionally, {\MIT} was captured using a mobile robot, which performs planar motion similar to a vehicle.
However, due to its smaller size, the robot's body experiences significantly more roll and pitch motion, potentially introducing greater motion noise compared to a vehicle.

\vspace{2mm}
\noindent\textbf{Diversity of geographic and cultural environments.} Lastly, we also aimed to assess whether the acquired data could be generalized across geographic and cultural differences.
Some studies~\cite{Cattaneo22tro-LCDNet} have claimed that training on {\KITTI} and testing on \texttt{KITTI-360} could demonstrate domain generalization.
However, since both datasets were collected in Germany using the same vehicle platform, they fall short of demonstrating actual cultural differences.

To address this issue, we leveraged datasets collected from campuses in Asia, Europe, and the USA to evaluate whether such geographic and cultural variations could be accounted for in our proposed benchmark.

\vspace{6mm}
\section{Trade-off between generalization and robustness against partial overlaps}\label{app:trade_off}

\begin{figure}[t!]
 	\centering
        \begin{subfigure}[b]{0.23\textwidth}
 		\includegraphics[width=1.0\textwidth]{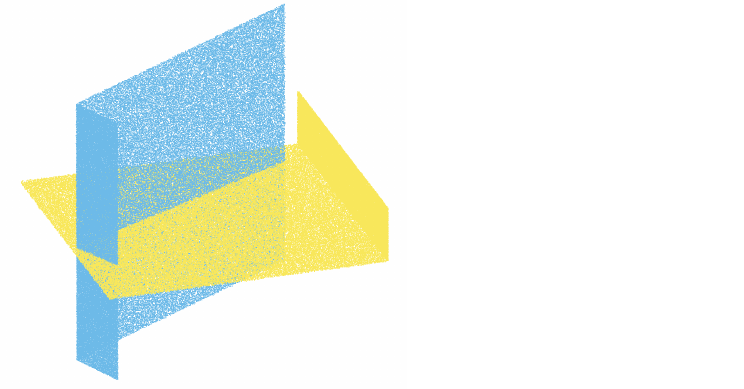}
        \caption{}
 	\end{subfigure}
        \begin{subfigure}[b]{0.23\textwidth}
 		\includegraphics[width=1.0\textwidth]{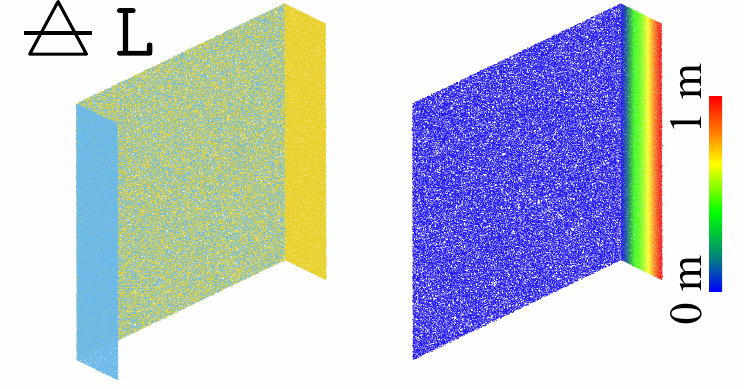}
        \caption{}
 	\end{subfigure}
        \begin{subfigure}[b]{0.23\textwidth}
 		\includegraphics[width=1.0\textwidth]{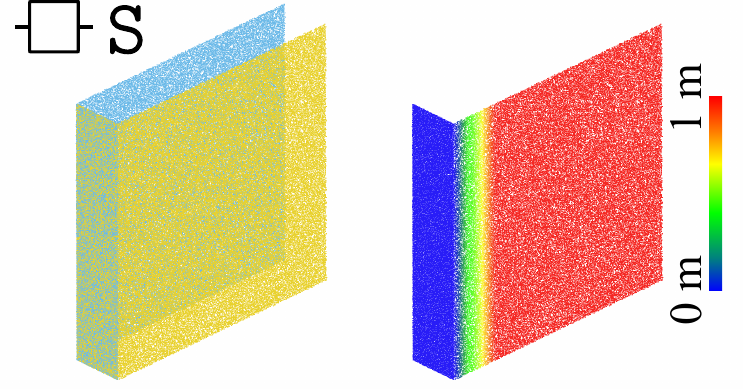}
        \caption{}
 	\end{subfigure}
 	\begin{subfigure}[b]{0.23\textwidth}
 		\includegraphics[width=1.0\textwidth]{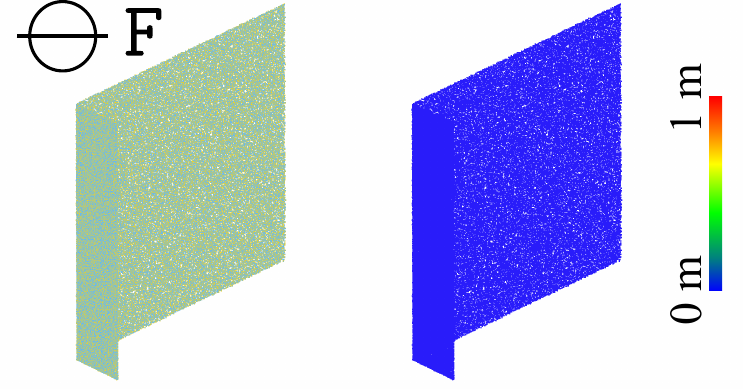}
        \caption{}
 	\end{subfigure}
        \begin{subfigure}[b]{0.23\textwidth}
 		\includegraphics[width=1.0\textwidth]{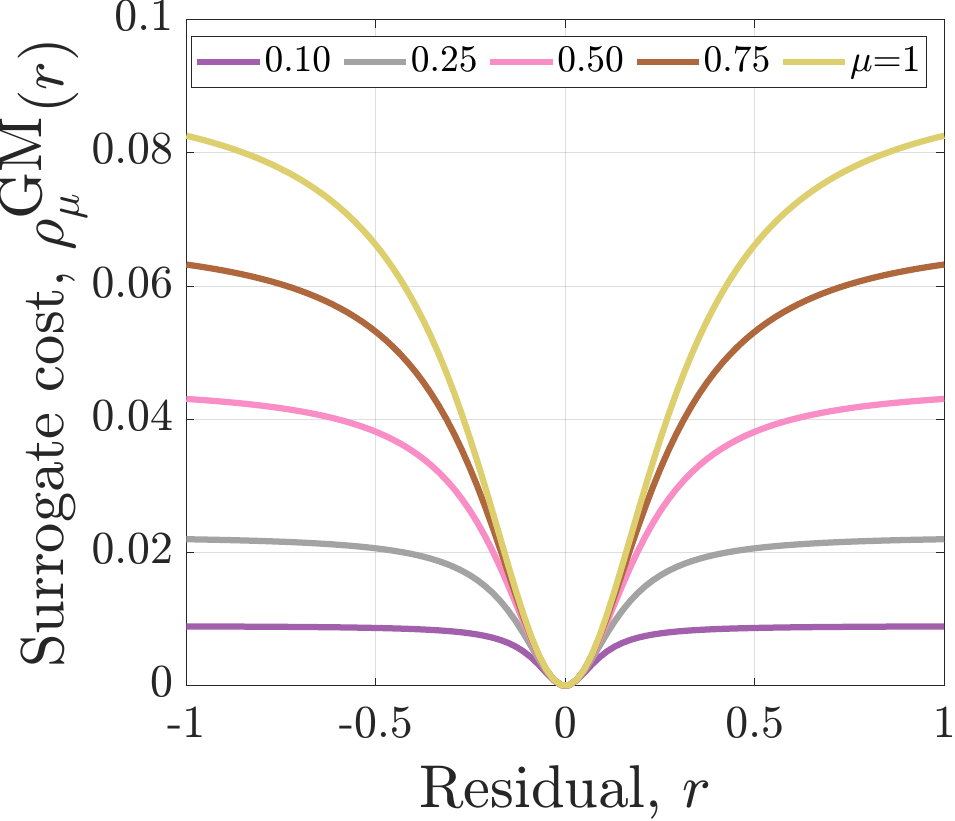}
        \caption{}
 	\end{subfigure}
        \begin{subfigure}[b]{0.23\textwidth}
 		\includegraphics[width=1.0\textwidth]{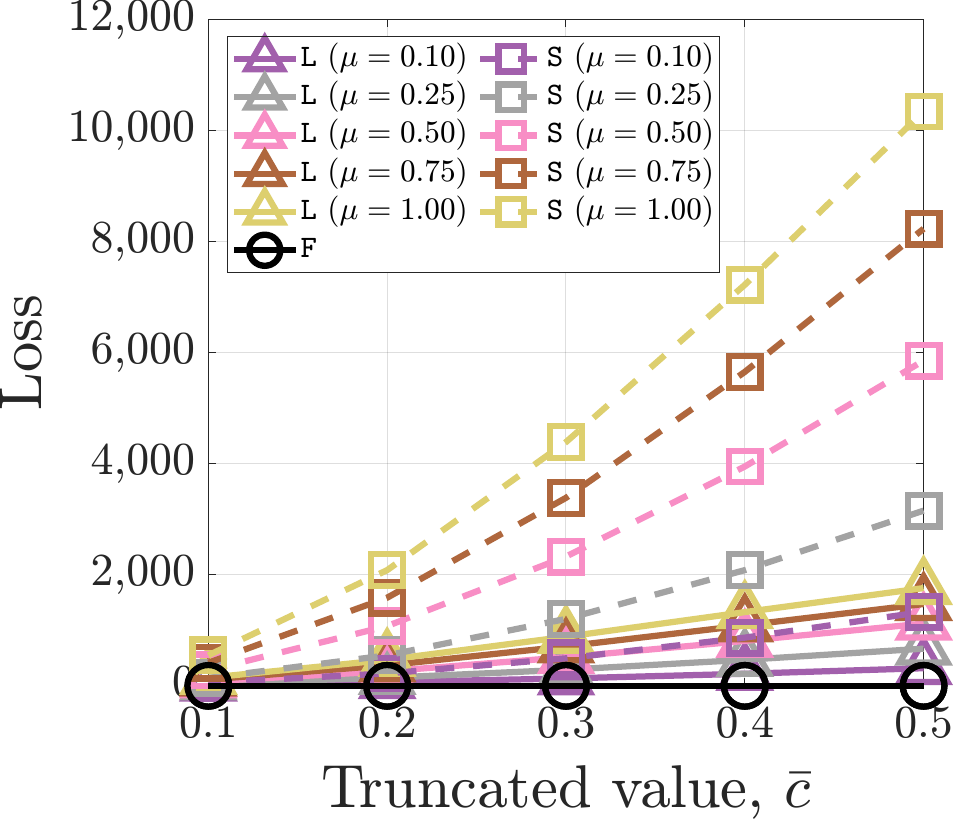}
        \caption{}
 	\end{subfigure} 
        \begin{subfigure}[b]{0.23\textwidth}
 		\includegraphics[width=1.0\textwidth]{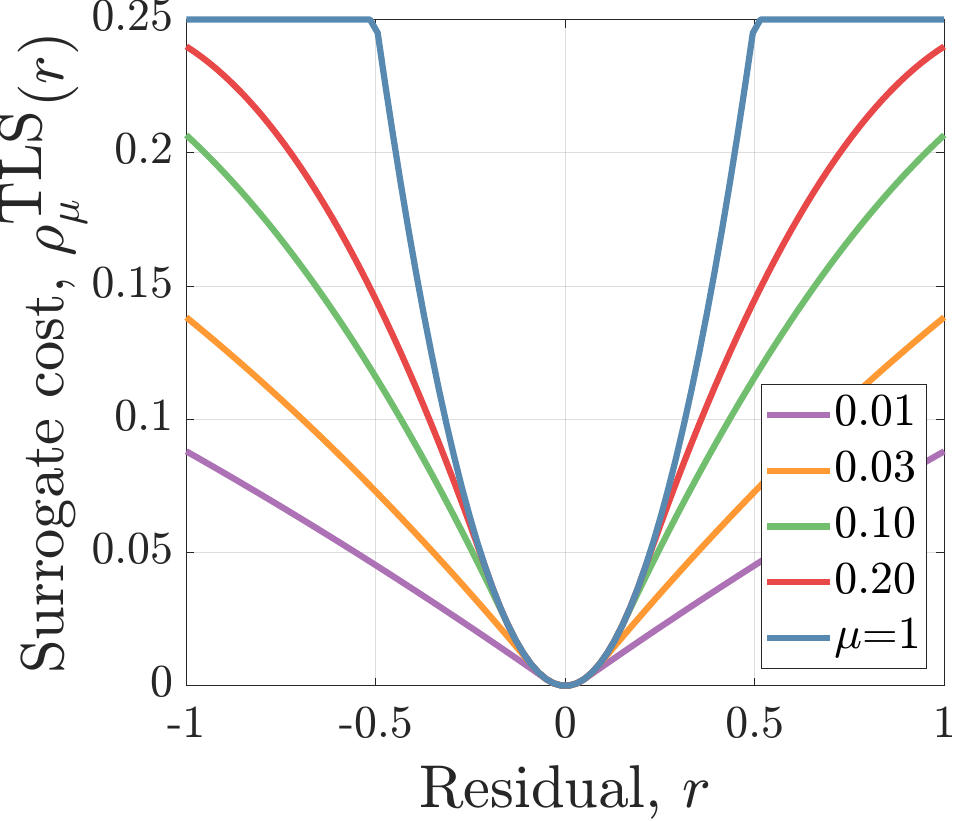}
        \caption{}
 	\end{subfigure}
        \begin{subfigure}[b]{0.23\textwidth}
 		\includegraphics[width=1.0\textwidth]{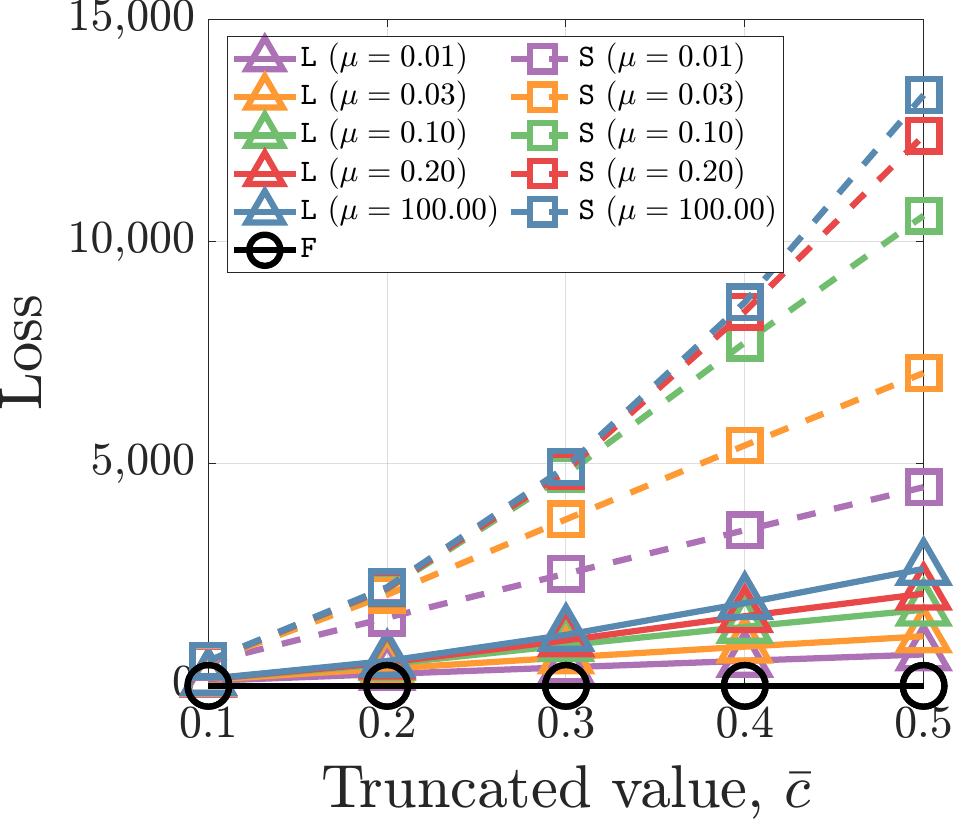}
        \caption{}
 	\end{subfigure}
 	\captionsetup{font=small}
 	\caption{Illustration of the \textit{global optimum ambiguity} problem in point cloud registration. (a) Example of two misaligned L-shaped point clouds and (b)–(d) possible registration cases. (b) An overlapped case along the longer segment~(expressed as \texttt{L}), (c) an overlapped case along the shorter segment~(expressed as \texttt{S}), and (d) a fully overlapped case~(expressed as \texttt{F}). (e) The behavior of the surrogate cost function of German-McClure~(GM) with varying control parameter $\mu$ and (f)~corresponding loss values for various $\mu$ with the user-defined threshold parameter $\bar{c}$. (g) The behavior of the surrogate cost function of truncated least squares~(TLS) with $\mu$ and (f)~corresponding loss values for various $\mu$ with the user-defined threshold parameter $\bar{c}$. Note that, regardless of the given $\mu$ and $\bar{c}$, the fully overlapped case~(\ie~(d)) always results in the lowest loss (since the loss value is zero for any $\mu$ and $\bar{c}$, for simplicity, only a single \texttt{F} is presented in (f) and (h)).}
    \label{fig:global_optim}
\end{figure}

Here, as a supplement to the explanation in \Cref{sec:limitation}, we explain with an example why our methodology performed relatively poorly on 3DLoMatch in detail~\cite{Huang21cvpr-PREDATORRegistration}.
Suppose we have two arbitrary L-shaped point clouds, as presented in \Cref{fig:global_optim}(a). Then, these two L-shaped point clouds can be registered in the following cases:
\begin{itemize}
\item Case A:~overlapped along the longer segment (\Cref{fig:global_optim}(b))
\item Case B:~overlapped in the short direction~(\Cref{fig:global_optim}(c))
\item Case C:~fully overlapped~(\Cref{fig:global_optim}(d)).
\end{itemize}

Obviously, the lowest value of the loss function is the fully overlapped case~(\ie Case C); however, in the partial overlap problem, the other local optima~(\ie~the partially overlapped case along the longer or shorter segment) can be the actual global optimum. 
This phenomenon implies that when partial overlap is severe, the relative pose with the smallest loss value is not necessarily the true global optimum.

Therefore, in terms of the optimization problem, we can say that the state with the actual global optimum is not necessarily equal to the state with the global optimum in the cost function.
Furthermore, this problem cannot be solved mathematically without prior knowledge of how the two point clouds should be aligned. 
We refer to this phenomenon as \textit{global optimum ambiguity}.

In addition, even with a learnable non-linear robust kernel, this problem cannot be perfectly resolved.
For instance, we examine two renowned non-convex cost functions: a)~German-McClure~(GM) function and b)~truncated least squares~(TLS) function, and use them as surrogate cost functions $\rho_\mu(r)$ that adjust their non-linearity by changing the control parameter~$\mu$~\cite{Yang20ral-GNC}.
Formally, by letting the $r$ be the residual and $\bar{c}$ be the user-defined parameter that determines the shape of a kernel, the GM function with $\mu$ can be expressed as follows:

\begin{equation}
\rho^\text{GM}_\mu(r)=\frac{\mu \bar{c}^2 r^2}{\mu \bar{c}^2+r^2},
\end{equation}

and the TLS function with $\mu$ can be expressed as follows:

\begin{equation}
\resizebox{0.5\textwidth}{!}{$
\rho^\text{TLS}_\mu(r)=\left\{\begin{array}{ll}
r^2 & \text {\!\!\! if } r^2 \in\left[0, \frac{\mu}{\mu+1} \bar{c}^2\right] \\
2 \bar{c}|r| \sqrt{\mu(\mu+1)}-\mu\left(\bar{c}^2+r^2\right) & \text {\!\!\! if } r^2 \in\left[\frac{\mu}{\mu+1} \bar{c}^2, \frac{\mu+1}{\mu} \bar{c}^2\right] . \\
\bar{c}^2 & \text {\!\!\! if } r^2 \in\left[\frac{\mu+1}{\mu} \bar{c}^2,+\infty\right)
\end{array}\right.
$}
\end{equation}

As shown in \Cref{fig:global_optim}, even if we vary the non-linearity of the kernel by adjusting the shape of the surrogate cost via $\mu$; see Figs.~\ref{fig:global_optim}(e) and (g), the fully overlapped case still yields the lowest loss; see Figs.~\ref{fig:global_optim}(f) and (h).
This supports our claim that the global optimum ambiguity problem cannot be easily resolved, no matter how much we formulate it as a non-linear function.

\newcommand{\scaleup}{\raisebox{-0.6ex}{\includegraphics[width=0.30cm]{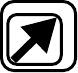}}}
\newcommand{\osu}{+ {\oracle} + {\scaleup}}
\begingroup
\begin{table*}[ht!]
        \setlength{\tabcolsep}{1pt}
        \centering
	{\scriptsize
		\begin{tabular}{l|l|cccccccccccc}
			\toprule \midrule
			& Env. & \multicolumn{5}{c}{Indoor} & \multicolumn{6}{c}{Outdoor} & \multirow{2}{*}{\begin{tabular}{@{}c@{}}Average \\ rank\end{tabular}} \\  \cmidrule(lr){3-7} \cmidrule(lr){8-13}
			& Dataset & \ThreeDMatch & \ThreeDLoMatch & \ScanNetppi & \ScanNetppF & \TIERS & \KITTI & \WOD & \KAIST &  \MIT & \ETH & \Oxford \\ \midrule
            \parbox[t]{5mm}{\multirow{3}{*}
            {\rotatebox[origin=c]{90}{\begin{tabular}{@{}c@{}}Conven- \\tional \end{tabular}}}}
       & FPFH~\cite{Rusu09icra-fast3Dkeypoints} + FGR~\cite{Zhou16eccv-FastGlobalRegistration} \ora & 62.53 & 15.42 & 77.68 & 92.31 & 80.60 & 98.74 & \firstc \textbf{100.00} & 89.80 & 74.78 & 91.87 & 99.00  & 8.55 \\
       & FPFH~\cite{Rusu09icra-fast3Dkeypoints} + Quatro~\cite{Lim22icra-Quatro} \ora & 8.22 & 1.74 & 9.88 & 97.27 & 86.57 & 99.10 & \firstc \textbf{100.00} & 91.46 & 79.57 & 51.05 & 91.03 & 10.82 \\
       & FPFH~\cite{Rusu09icra-fast3Dkeypoints} + TEASER++~\cite{Yang20tro-teaser}  \ora & 52.00 & 13.25 & 66.15 & 97.22 & 73.13 & 98.92 & \firstc \textbf{100.00} & 89.20 & 71.30 & 93.69 & \thirdc 99.34 & 8.82 \\ \midrule
			\parbox[t]{2mm}{\multirow{16}{*}{\rotatebox[origin=c]{90}{Deep}}}
       & FCGF~\cite{Choi19iccv-FCGF}
       & 8.04 & 0.17 & 19.96 & 23.07 & 77.82 & \cg 98.92 & 95.38 & 88.34& 82.17 & 6.59 & 75.08 & 14.82\\
       & \ora & 34.97 & 4.12 & 31.00 & 25.10 & 77.93 & \cg  98.92& 96.92 & 94.22 & 89.13 & 39.97 & 86.05 & 12.00 \\
       & \osu & 34.97 & 4.12 & 33.37 &25.10 & 77.93 & \cg 98.92 & 99.23 &94.22 & 90.43 & 39.97 & 90.03 & 11.27\\
       & Predator~\cite{Huang21cvpr-PREDATORRegistration} & N/A (Err) & N/A (Err) & N/A (Err) & N/A (Err) & 69.43 & \cg 99.82 & \firstc \textbf{100.00} & N/A (OOM) & 54.78 & 55.68 & 89.04 & 14.09 \\
       &  \ora & 16.47 & 0.00 & 9.40 & 3.72 &  69.77 & \cg 99.82 & \firstc \textbf{100.00}  & 71.3 & 76.52 & 56.67 & 89.04 & 12.00 \\
       &  \osu & 23.2 & 3.31 & 9.40 & 3.72 & 69.77 & \cg 99.82 & \firstc \textbf{100.00} & 94.02 & 86.08 & 71.95 & 95.02 & 9.91 \\
       & GeoTransformer~\cite{Qin23tpami-GeoTransformer}  & N/A (Err) & N/A (Err) & N/A (Err) & N/A (Err) & N/A (Err) & \cg 99.82 & \firstc \textbf{100.00} & 63.84 & 93.91 & 77.00 & 73.42 & 12.73\\
       &  \ora & 5.94 & 0.30 & 15.91 & 34.18 & 20.57 & \cg 99.82 & \firstc \textbf{100.00} & 63.84 & 93.91 & 77.56 & 73.42 & 10.90\\
       &  \osu & 62.17 & 14.38 & 76.52 & 90.63 & 87.36 & \cg 99.82 & \firstc \textbf{100.00} & 96.84 & 96.52 & 81.77 & 98.01 & 5.55\\
       & BUFFER~\cite{Ao23CVPR-BUFFER} & N/A (Err) & N/A (Err) & 17.60 & 88.84 & 93.34& \cg 99.64 & \firstc \textbf{100.00} & 99.50 & 95.22 & 98.18 & \thirdc 99.34 & 8.27 \\
     &  \ora  & \thirdc 91.19 & \secondc 64.51 & \secondc 93.15 & \thirdc 97.81 & \thirdc 93.57 & \cg 99.64 & \firstc \textbf{100.00} & \firstc \textbf{99.55} & \thirdc 97.39 & \thirdc 99.86 & \thirdc 99.34 & \thirdc 3.27\\
       &  \osu & \thirdc 91.19 & \secondc 64.51 & \secondc 93.15 & \thirdc 97.81 & \thirdc 93.57 & \cg 99.64 & \firstc \textbf{100.00} & \firstc \textbf{99.55} & \thirdc 97.39 & \thirdc 99.86 & \thirdc 99.34 & \thirdc 3.27 \\
	& PARENet~\cite{Yao24iccv-PARENet}  & N/A (Err) & N/A (Err) & N/A (Err) & N/A (Err) & N/A (Err) & \cg 99.82 & 97.69 & 57.51 & 75.22 & 68.30 & 66.11 & 15.82 \\
      & \ora & 0.77 & 0.10 & 3.04 & 12.00 & 19.20 & \cg 99.82 & 98.46 & 57.51 & 75.22 & 68.44 & 66.11 & 15.00\\
      & \osu & 22.09 & 4.98 & 29.99 & 42.91 & 52.99 & \cg 99.82 & \firstc \textbf{100.00} & 89.50 & 87.39 & 72.65 & 94.02 & 9.18 \\ 
      \cmidrule(lr){2-14}
        & Ours with only $r_m$  & \secondc 91.96 & 63.59 & 92.38 & \secondc 99.45 & \secondc 94.37 & \cg 99.82 & \firstc \textbf{100.00} &\firstc \textbf{99.55}&\firstc \textbf{99.13}&\firstc \textbf{100.00} & \firstc \textbf{99.67} & \secondc 1.82 \\ 
    
        & Ours & \firstc \textbf{93.79} & \firstc \textbf{65.89} & \firstc \textbf{95.13} & \firstc \textbf{99.65} & \firstc \textbf{94.83} & \cg 99.82 & \firstc \textbf{100.00} & \firstc \textbf{99.55} & \firstc \textbf{99.13} & \firstc \textbf{100.00} & \firstc \textbf{99.67} & \firstc \textbf{1.00} \\ \midrule \bottomrule
		\end{tabular}
	}
  \captionsetup{font=small}
    \vspace{-2mm}
    \caption{
    Quantitative comparison of generalization performance in terms of success rate (unit: \%). Deep learning-based models were trained only on {\KITTI}~\cite{Geiger13ijrr-KITTI} and RANSAC was used with a maximum iteration of 50K.
    Icons represent oracle tuning~(\oracle) for voxel size and radius, and scale alignment~(\scaleup) to match dataset scales (e.g., adjusting {\ThreeDMatch}’s scale to {\KITTI} by multiplying $\frac{0.3}{0.025}$).
    N/A~(Err) indicates failure due to an insufficient number of points remaining after voxelization with the voxel size typically used for outdoor settings, making keypoint extraction or descriptor generation infeasible. N/A~(OOM) indicates an out-of-memory error caused by excessive memory usage.
    }
	\label{table:success_rates_kitti}
  \vspace{-6mm}
\end{table*}
\endgroup

\section{Quantitative results using \KITTI}
\label{sec:add_quantitative}

One may wonder how the model performs when trained on \KITTI, so we also present the results of training on {\KITTI} in \Cref{table:success_rates_kitti}.
Interestingly, the success rates in {\ThreeDMatch} and {\ThreeDLoMatch} become slightly lower, while those in {\TIERS}, {\KAIST}, {\MIT}, {\ETH} improve when our BUFFER-X was trained on {\KITTI} instead of \ThreeDMatch. 
We speculate that because the model was trained on LiDAR data, it is better optimized for the distribution of LiDAR point clouds. Additionally, training on sparse point clouds results in exposure to relatively less diverse patch patterns, which may explain the performance drop when testing on {\ThreeDMatch}.

Specifically, because the cloud points in {\ThreeDMatch} are much denser than those in \KITTI, randomly sampling $N_\text{patch}$ points from these clouds for each patch results in more varied local coordinate patterns.
Thus, training in {\ThreeDMatch} enables the model to learn more diverse local neighborhood patterns during training by randomly adjusting the size of $N_\text{patch}$. 
In contrast, in the case of \KITTI, the points are relatively sparse because they are acquired using a LiDAR sensor. As a result, even when randomly sampling $N_\text{patch}$ points within the local neighborhoods, the diversity of local patterns remains relatively limited.

Moreover, our method demonstrated a higher rank compared to the state-of-the-art approaches. 
In particular, we observed that applying the voxel size used for outdoor training to indoor environments resulted in too few points remaining, leading to unexpected errors (referred to as ``N/A (Err)'' in \Cref{table:success_rates_kitti}).
For example, when downsampling a 
$3\times 3 \times 6$\,m$^3$ space with the 0.3\,m voxel size, which is a typical size used for outdoor settings, only 2,000 points remain. 
In addition, we found that networks requiring large memory, such as Predator, encountered out-of-memory errors when processing denser point clouds~(referred to as ``N/A (OOM)'' in \Cref{table:success_rates_kitti}).
This means that an out-of-memory issue occurs when handling a 128-channel LiDAR point cloud without manual tuning of the parameters used for a 64-channel LiDAR point cloud.

Therefore, while training on {\KITTI} led to some variations in overall performance, our approach remains more robust than other state-of-the-art methods.

\section{Qualitative results in diverse scenes}

We present the qualitative results in \Cref{fig:viz_indoor} and \Cref{fig:viz_outdoor}. 
Remarkably, even when trained on \ThreeDMatch, which consists solely of RGB-D point clouds with an approximate maximum range of 3.5\,m, the model performs robustly on sparse LiDAR point clouds in both indoor and outdoor environments.
In particular, our approach successfully performed registration regardless of the sensor type or acquisition setup, whether the sparse 3D point clouds were acquired by solid-state LiDAR sensors~(the first row among the {\KAIST} rows in \Cref{fig:viz_outdoor}) or captured from a robot platform~({\MIT} rows in \Cref{fig:viz_outdoor}).

\newcommand{\qualitativewidth}{0.265}
\begin{figure*}[t!]
    \centering
    \setlength{\arrayrulewidth}{0.3pt}  
    \arrayrulecolor[gray]{0.7} 
    \begin{tabular}{c c c c}
        \multirow{2}{*}{\makebox[0pt][r]{\rotatebox{90}{\ScanNetppi}\hspace{-0.1cm}}} &
        \begin{subfigure}[b]{\qualitativewidth\textwidth}
            \includegraphics[width=1.0\textwidth]{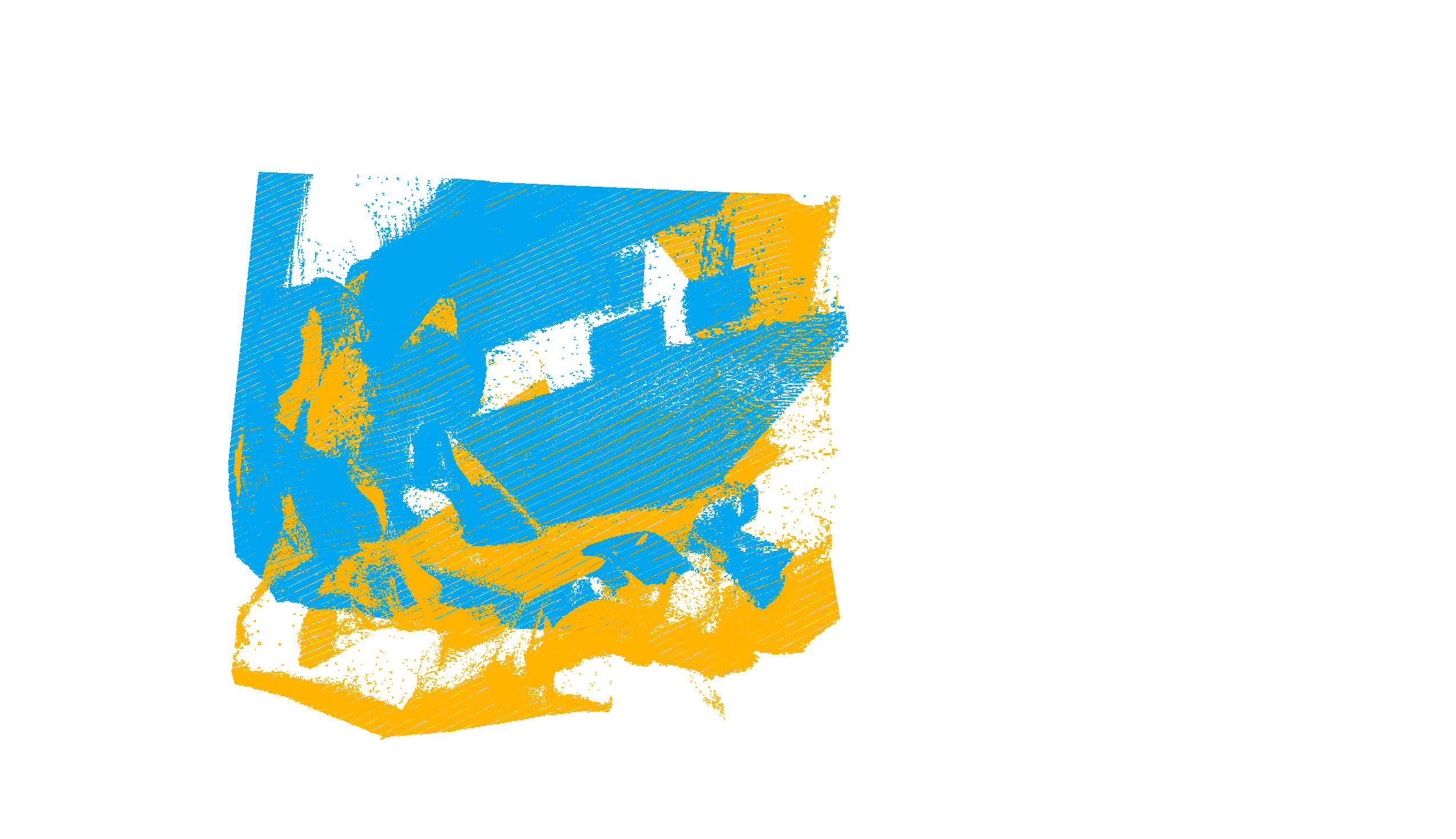} 
        \end{subfigure} &
        \begin{subfigure}[b]{\qualitativewidth\textwidth}
            \includegraphics[width=1.0\textwidth]{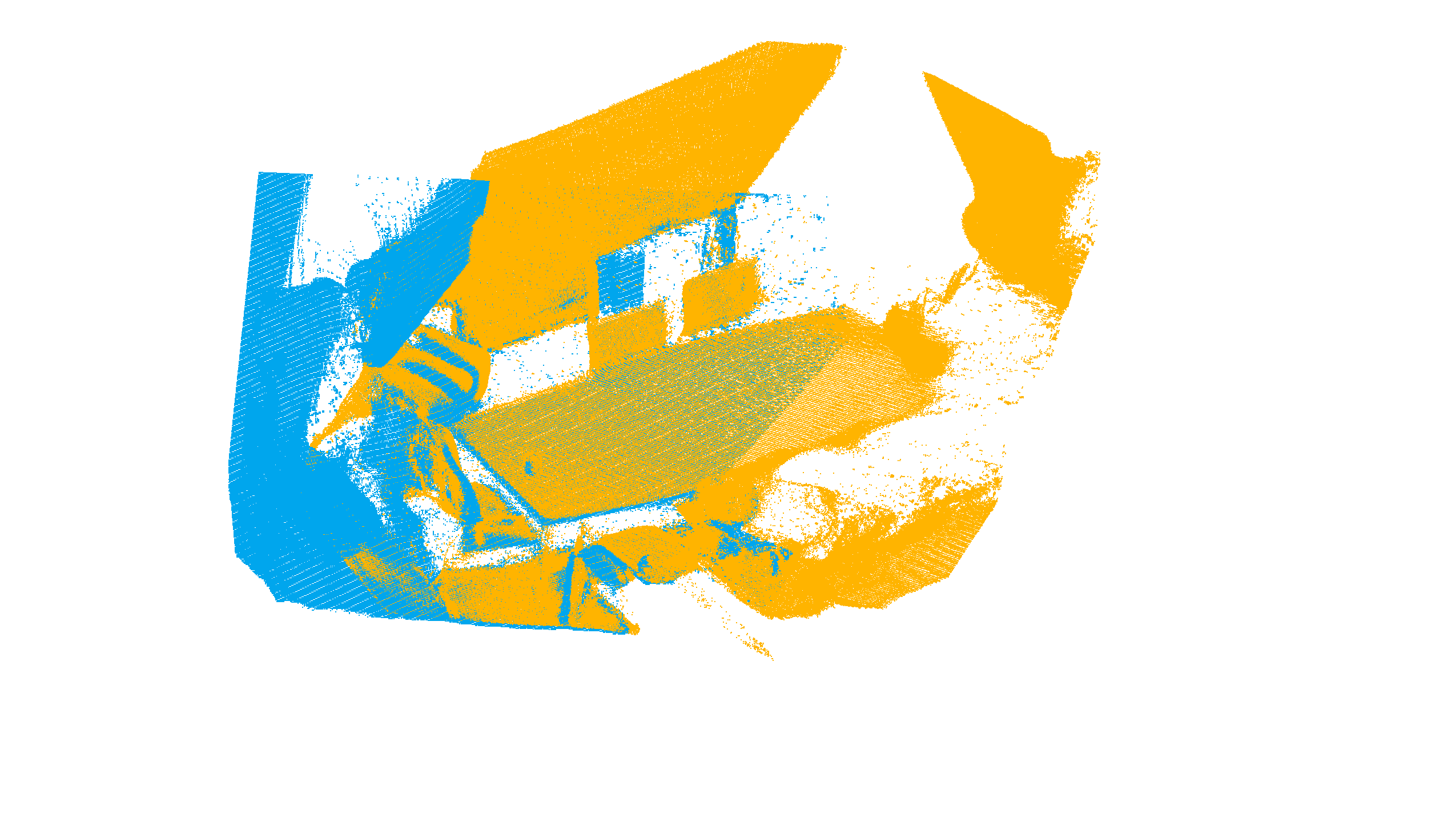}
        \end{subfigure} &
        \begin{subfigure}[b]{\qualitativewidth\textwidth}
            \includegraphics[width=1.0\textwidth]{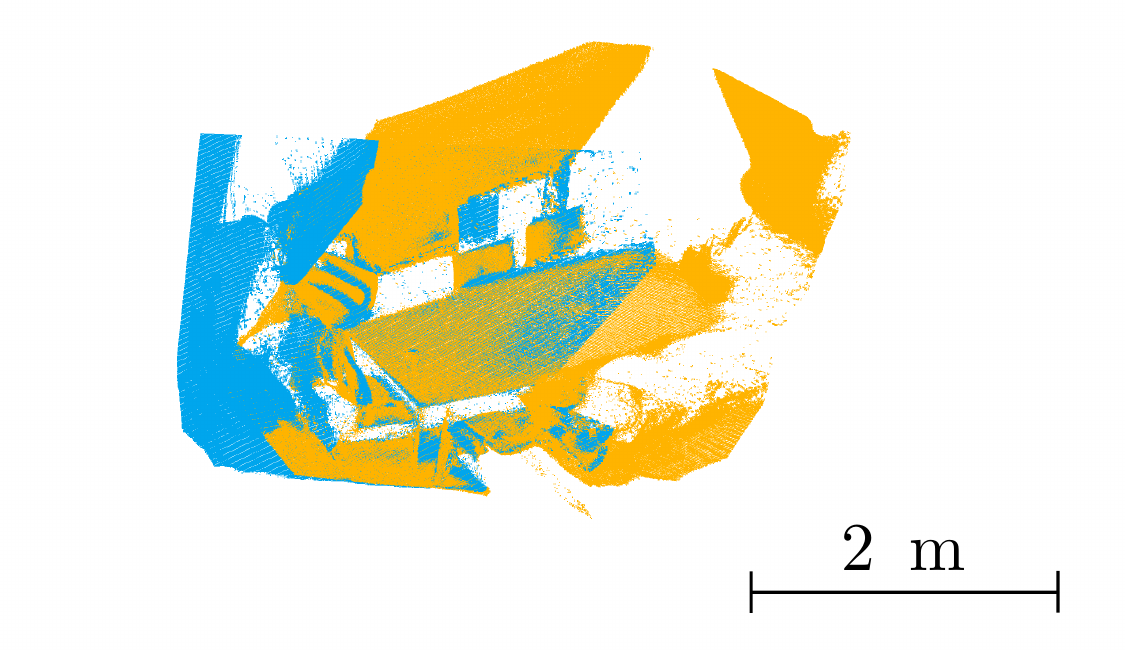}
        \end{subfigure} \\
        &
        \begin{subfigure}[b]{\qualitativewidth\textwidth}
            \includegraphics[width=1.0\textwidth]{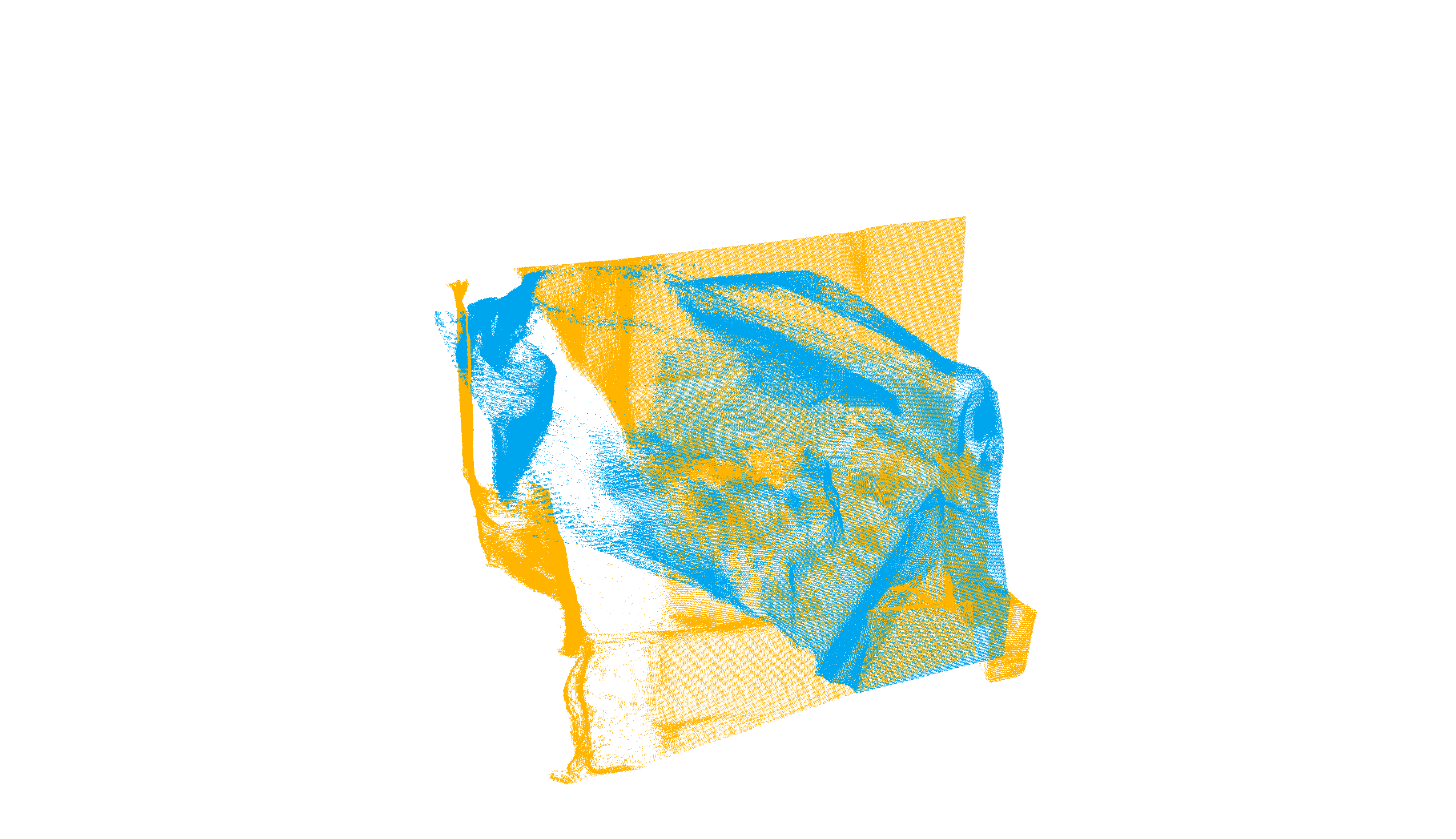} 
        \end{subfigure} &
        \begin{subfigure}[b]{\qualitativewidth\textwidth}
            \includegraphics[width=1.0\textwidth]{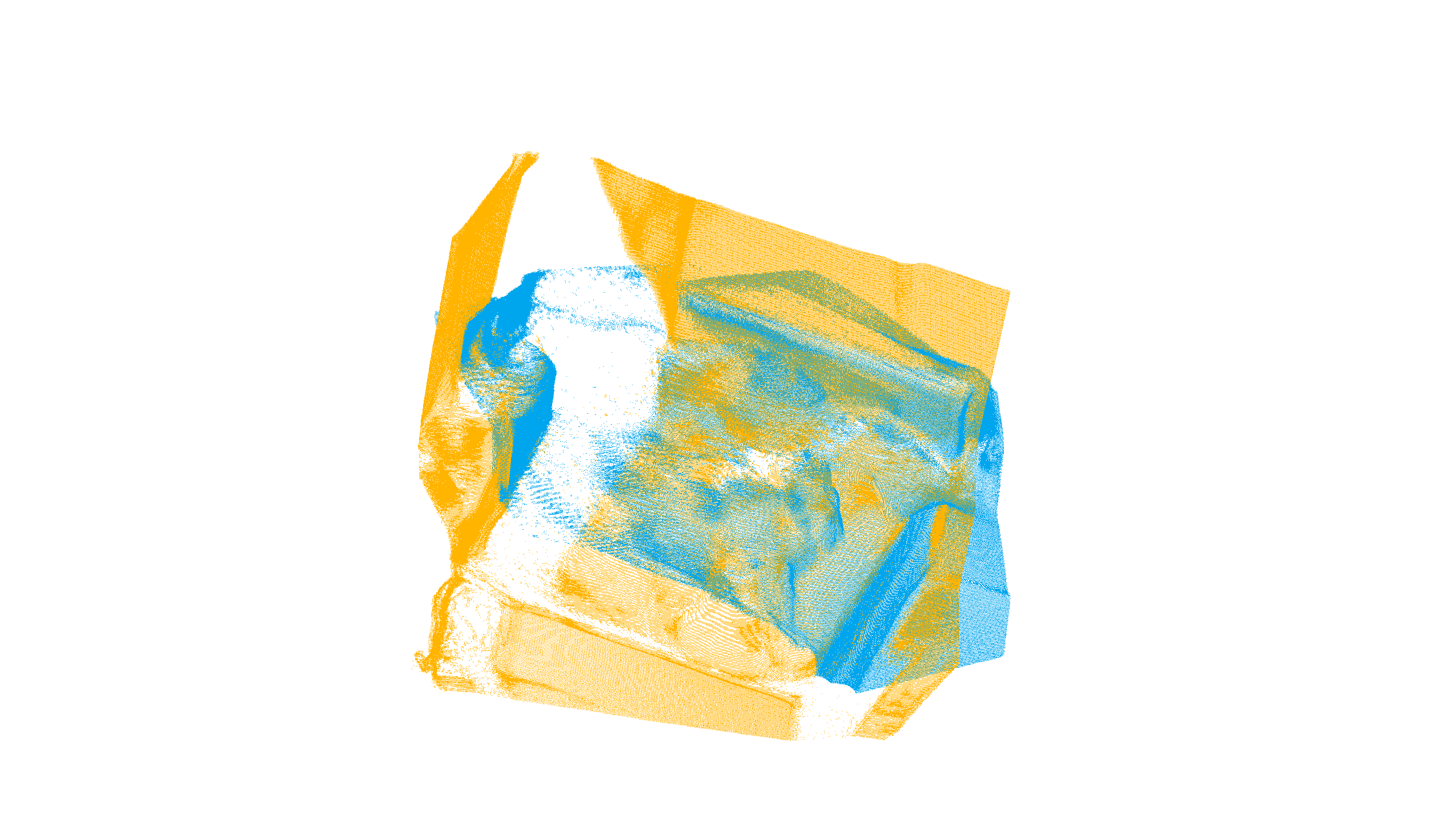}
        \end{subfigure} &
        \begin{subfigure}[b]{\qualitativewidth\textwidth}
            \includegraphics[width=1.0\textwidth]{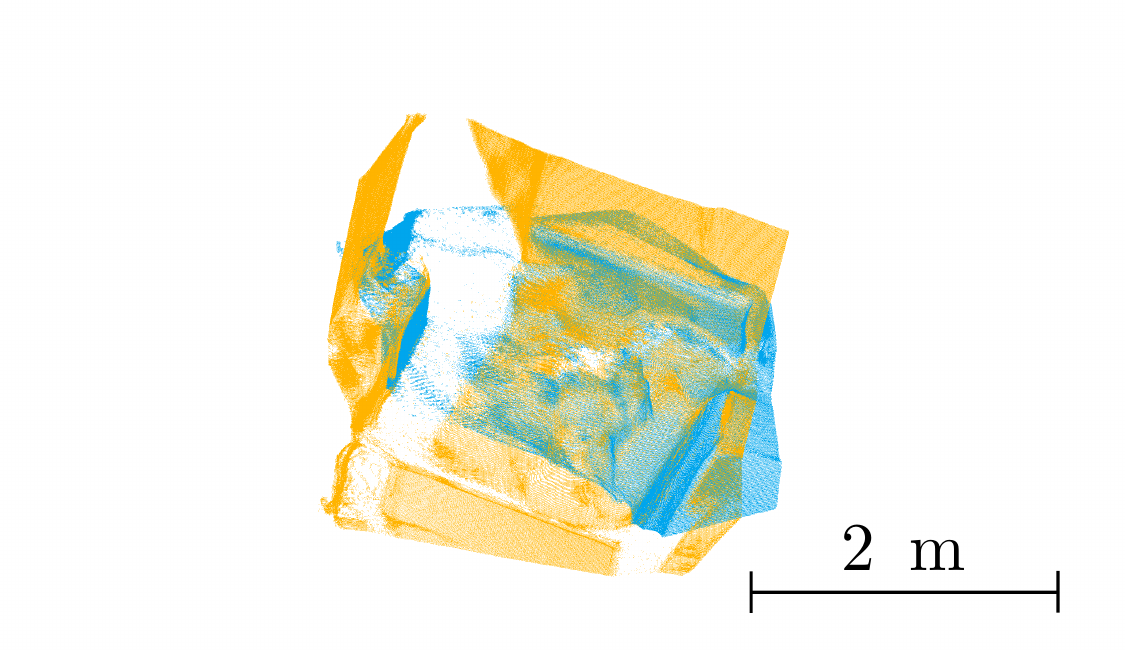}
        \end{subfigure} \\
    \\ \midrule \\
    \multirow{2}{*}{\makebox[0pt][r]{\rotatebox{90}{\ScanNetppF}\hspace{-0.1cm}}} &
    \begin{subfigure}[b]{\qualitativewidth\textwidth}
        \includegraphics[width=1.0\textwidth]{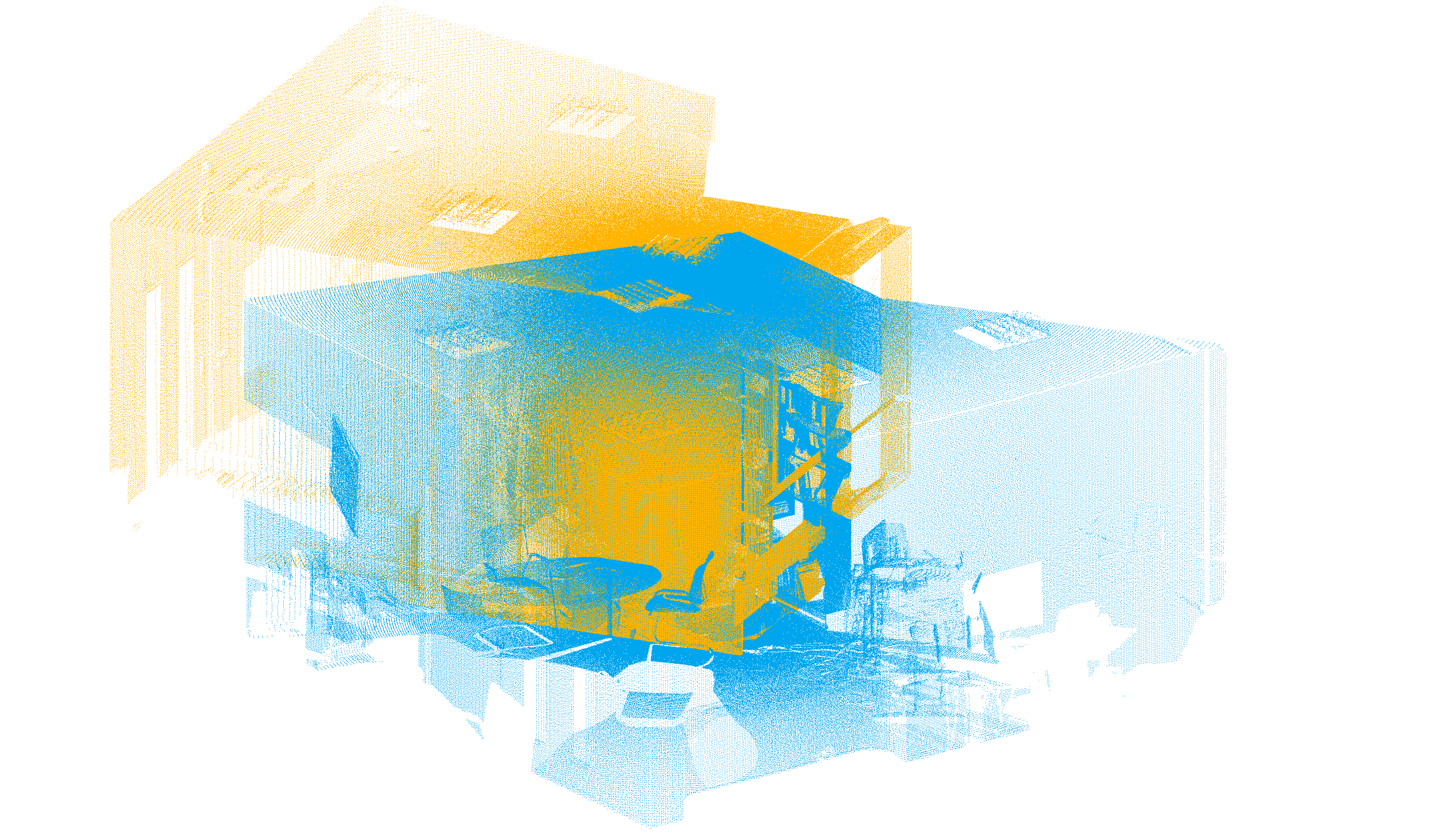} 
    \end{subfigure} &
    \begin{subfigure}[b]{\qualitativewidth\textwidth}
        \includegraphics[width=1.0\textwidth]{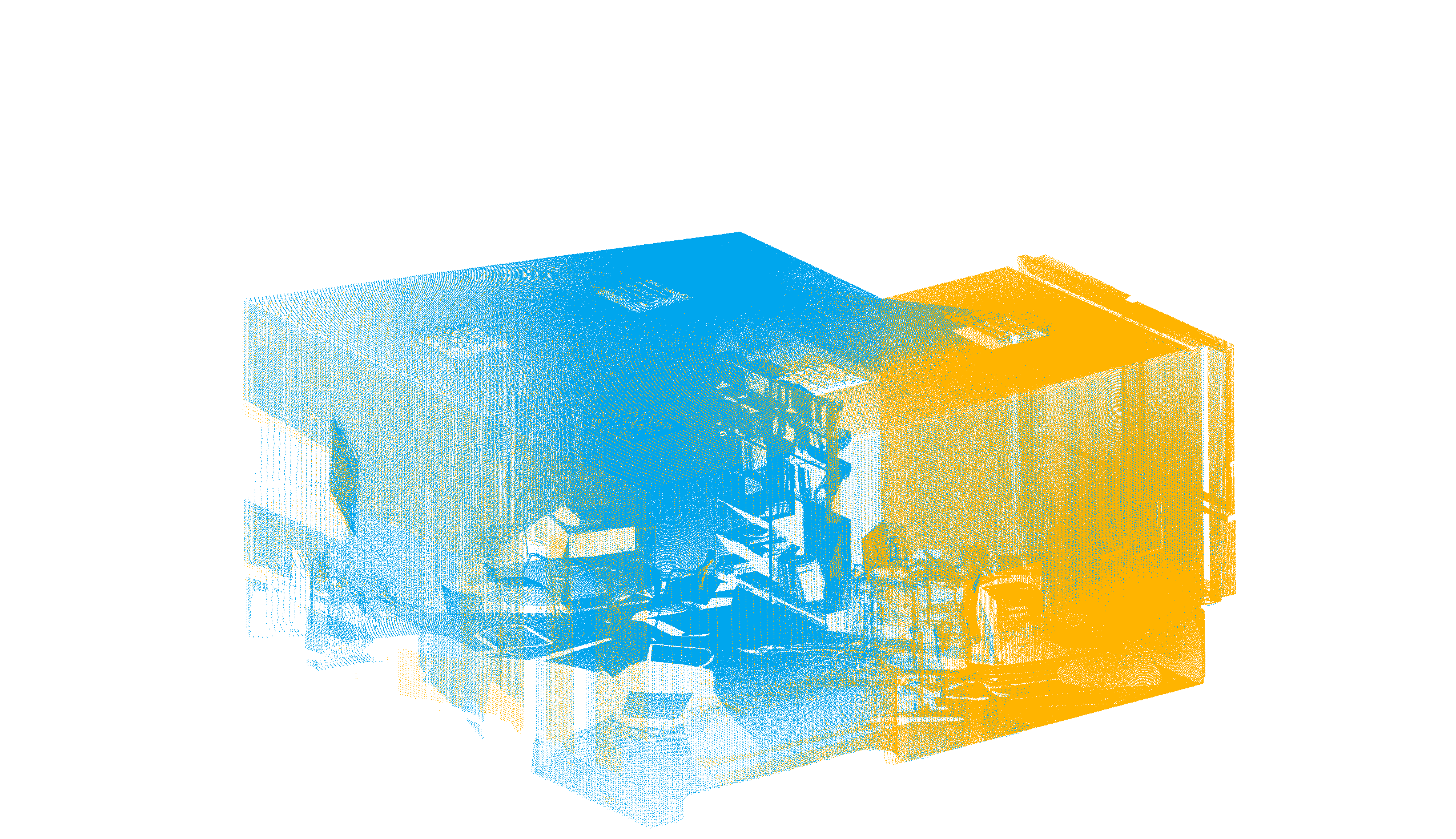}
    \end{subfigure} &
    \begin{subfigure}[b]{\qualitativewidth\textwidth}
        \includegraphics[width=1.0\textwidth]{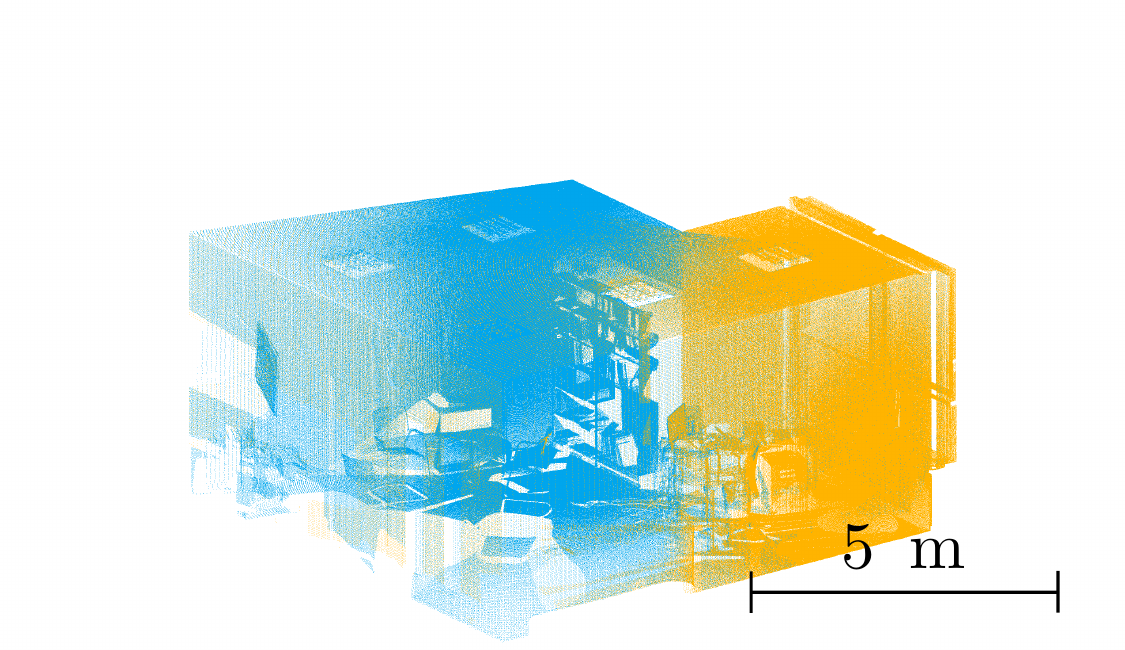}
    \end{subfigure} \\
    &
    \begin{subfigure}[b]{\qualitativewidth\textwidth}
        \includegraphics[width=1.0\textwidth]{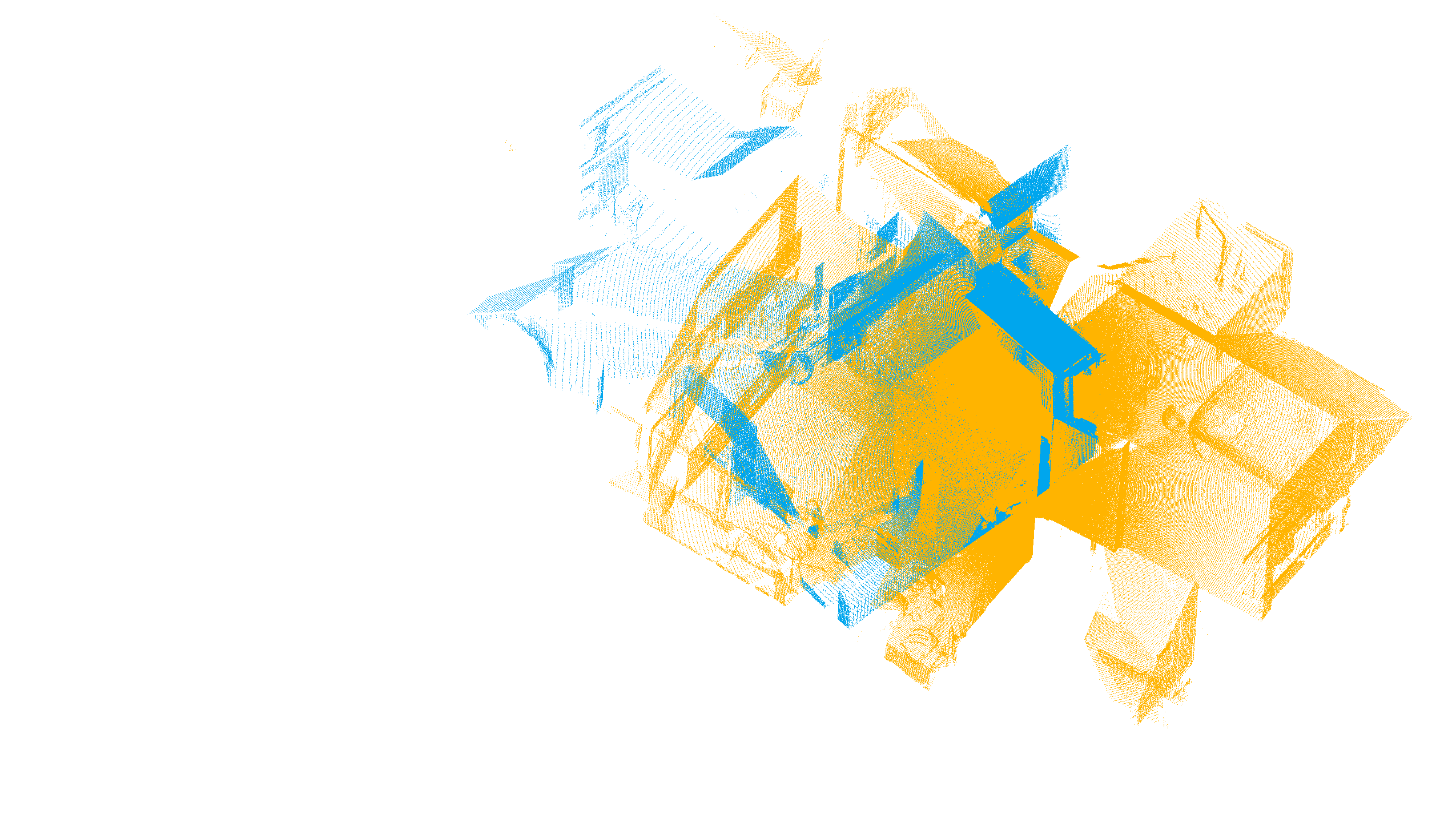}
    \end{subfigure} &
    \begin{subfigure}[b]{\qualitativewidth\textwidth}
        \includegraphics[width=1.0\textwidth]{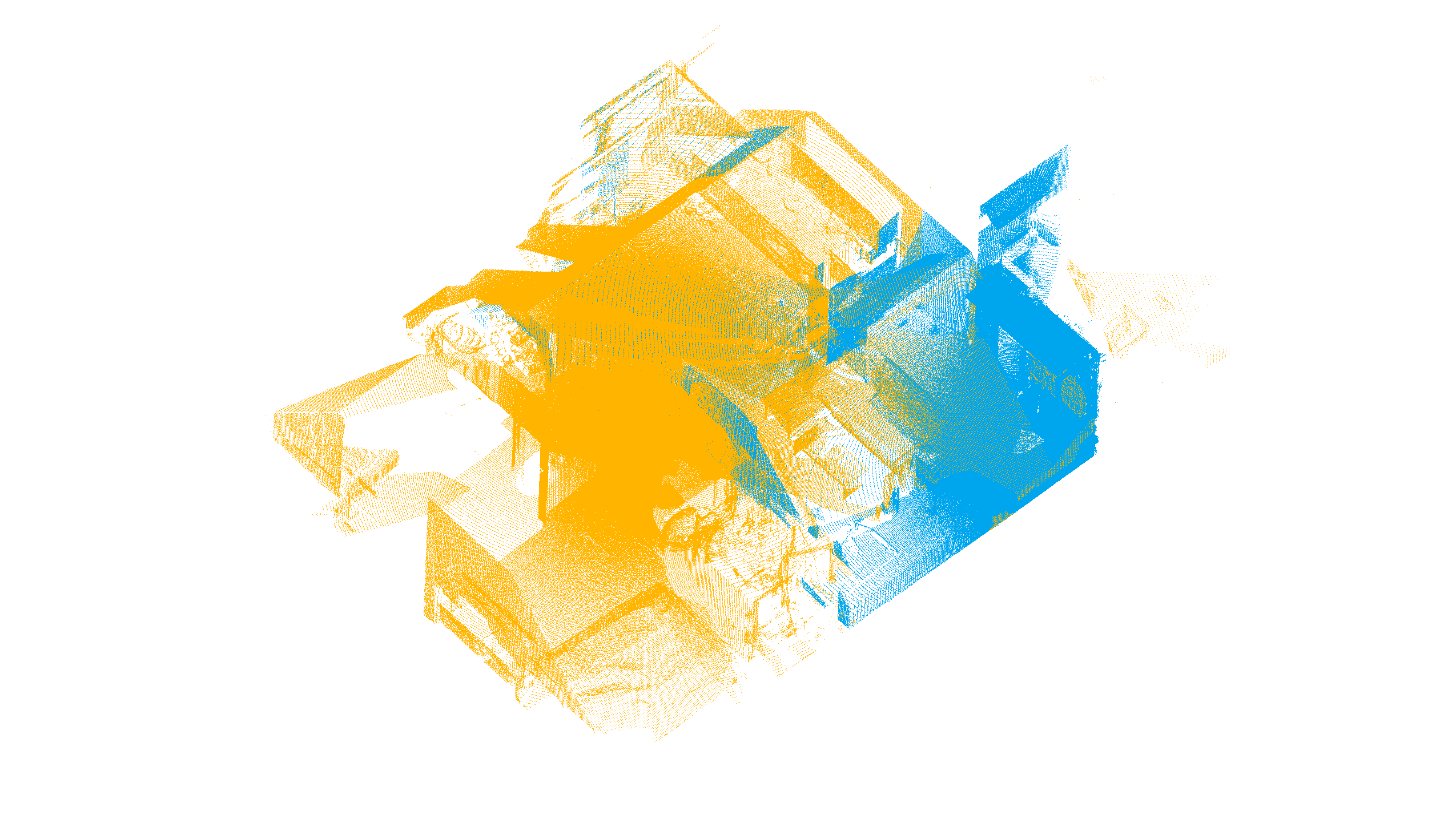}
    \end{subfigure} &
    \begin{subfigure}[b]{\qualitativewidth\textwidth}
        \includegraphics[width=1.0\textwidth]{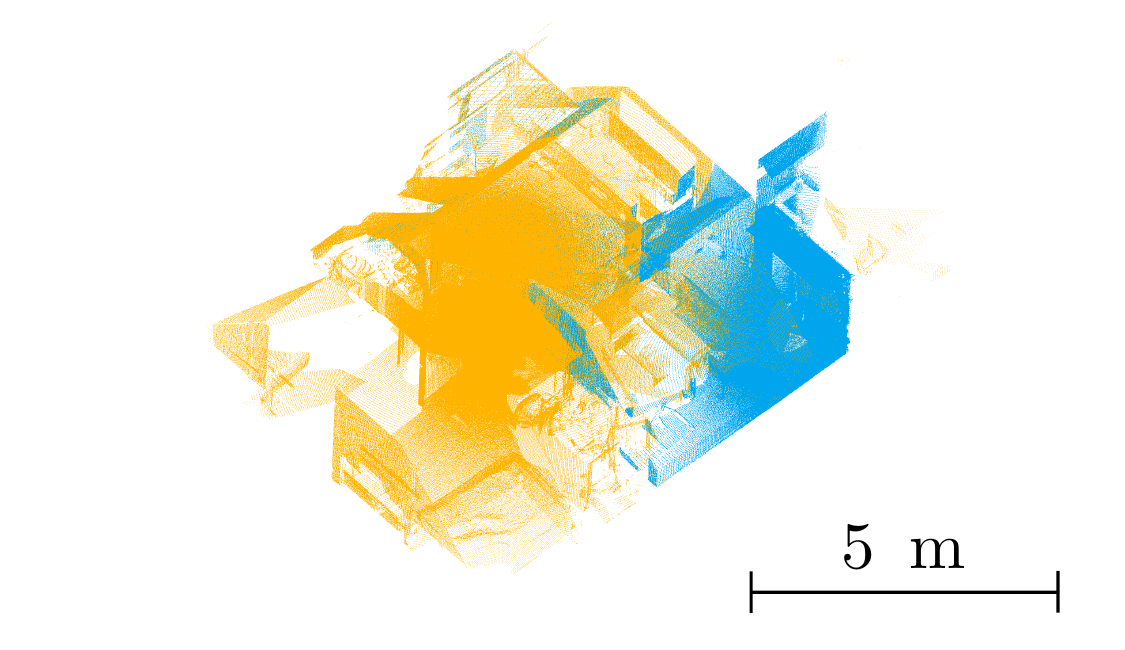}
    \end{subfigure} \\
    \\ \midrule \\
    \multirow{2}{*}{\makebox[0pt][r]{\rotatebox{90}{\TIERS}\hspace{-0.1cm}}} &
    \begin{subfigure}[b]{\qualitativewidth\textwidth}
        \includegraphics[width=1.0\textwidth]{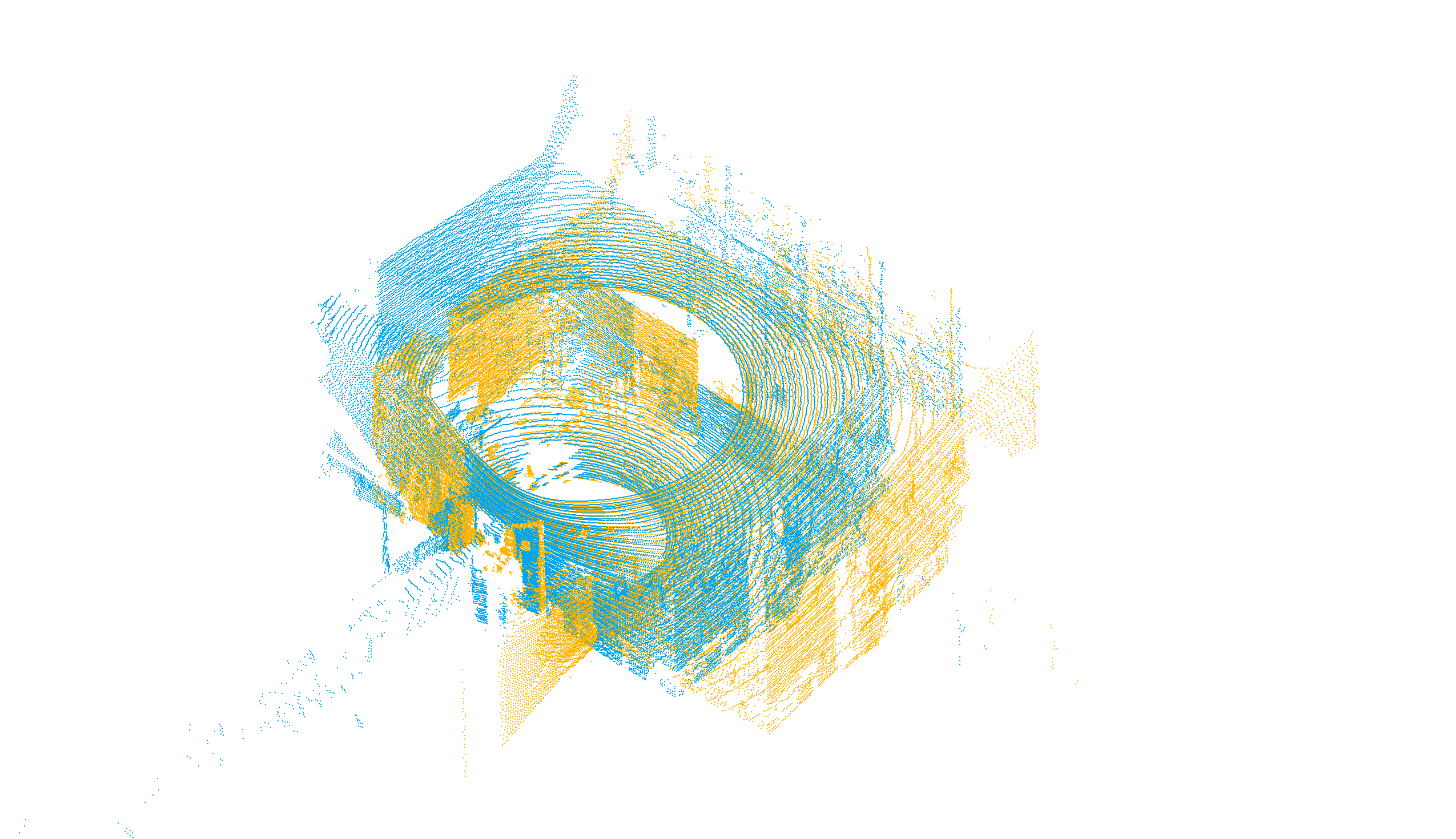}
    \end{subfigure} &
    \begin{subfigure}[b]{\qualitativewidth\textwidth}
        \includegraphics[width=1.0\textwidth]{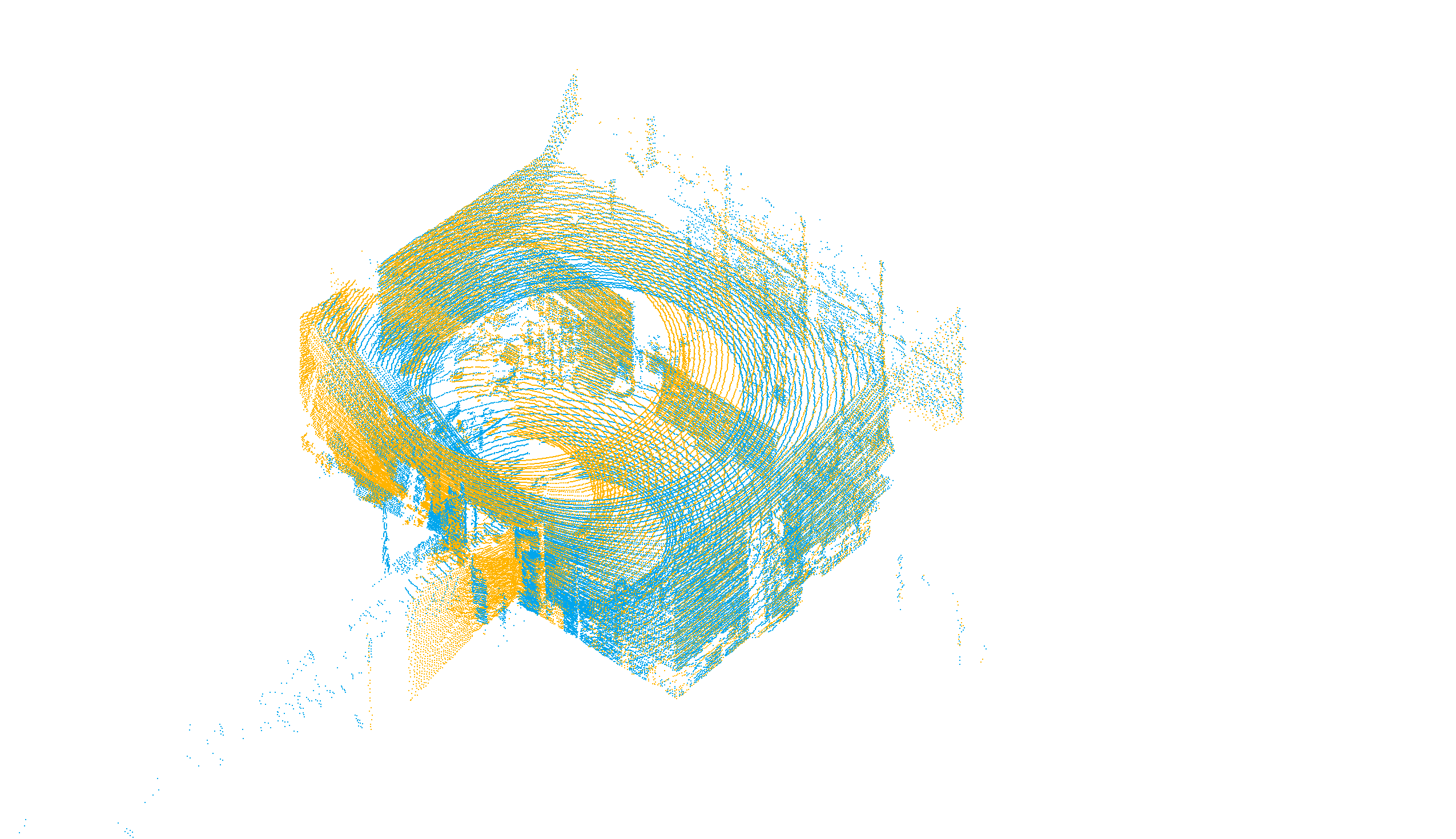}
    \end{subfigure} &
    \begin{subfigure}[b]{\qualitativewidth\textwidth}
        \includegraphics[width=1.0\textwidth]{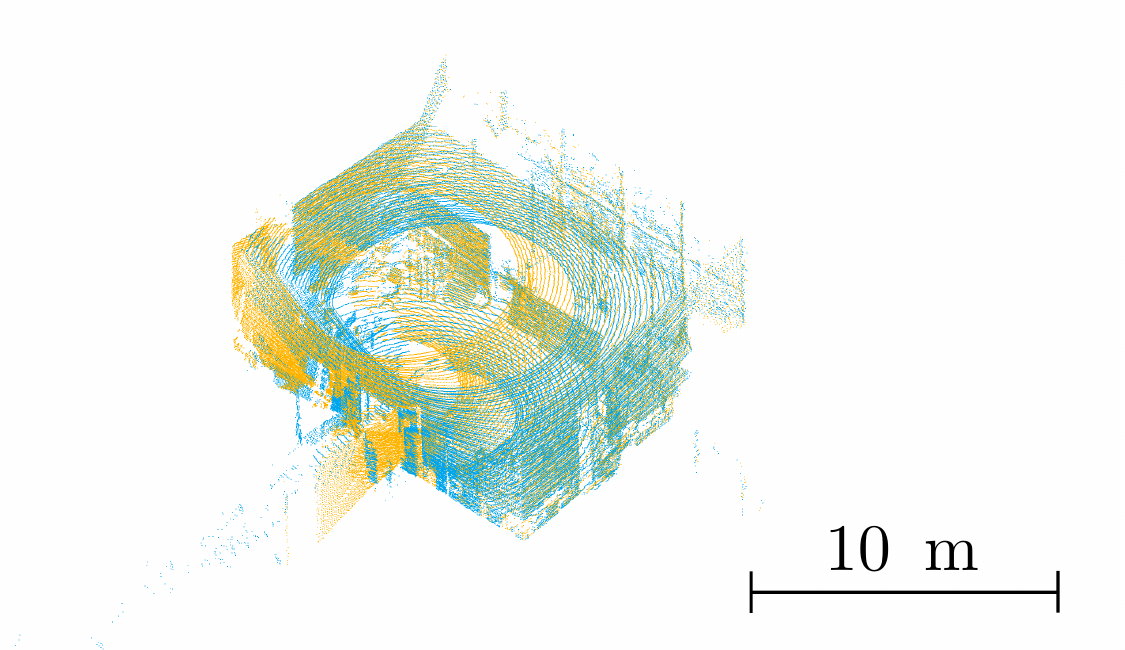}
    \end{subfigure} \\
    &
    \begin{subfigure}[b]{\qualitativewidth\textwidth}
        \includegraphics[width=1.0\textwidth]{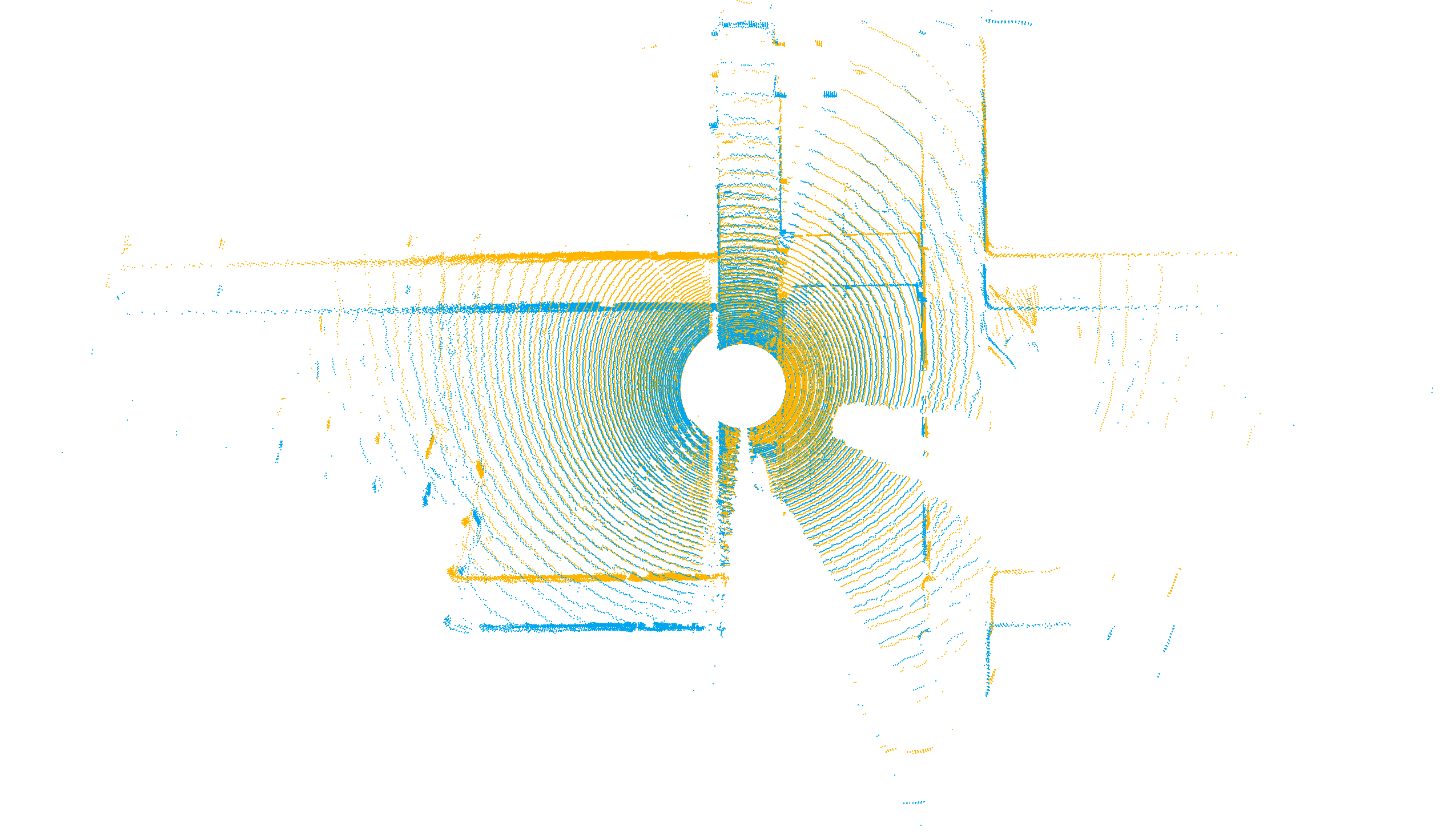}
        \caption*{(a) Source and target}
    \end{subfigure} &
    \begin{subfigure}[b]{\qualitativewidth\textwidth}
        \includegraphics[width=1.0\textwidth]{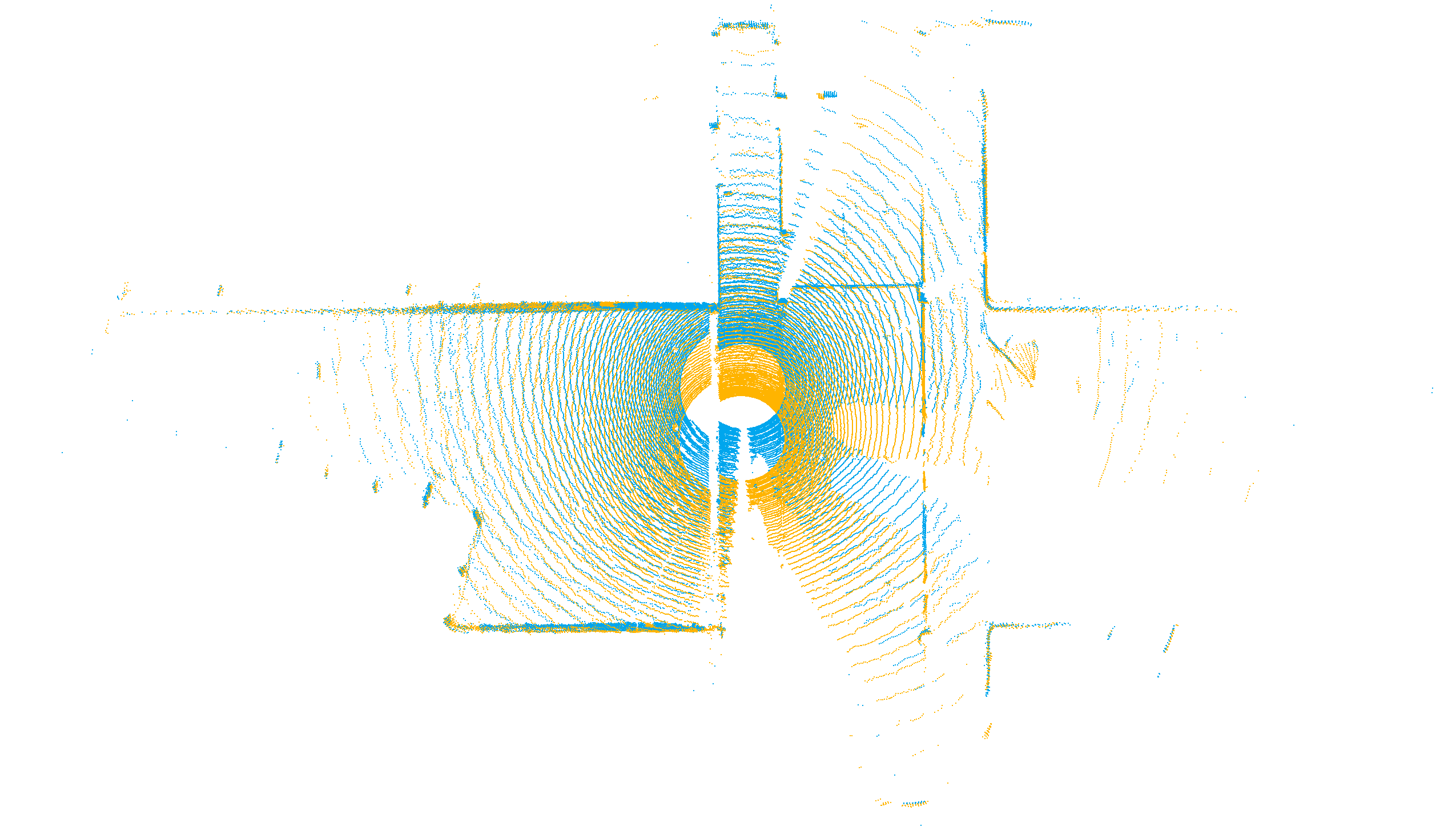}
        \caption*{(b) BUFFER-X~(Ours)}
    \end{subfigure} &
    \begin{subfigure}[b]{\qualitativewidth\textwidth}
        \includegraphics[width=1.0\textwidth]{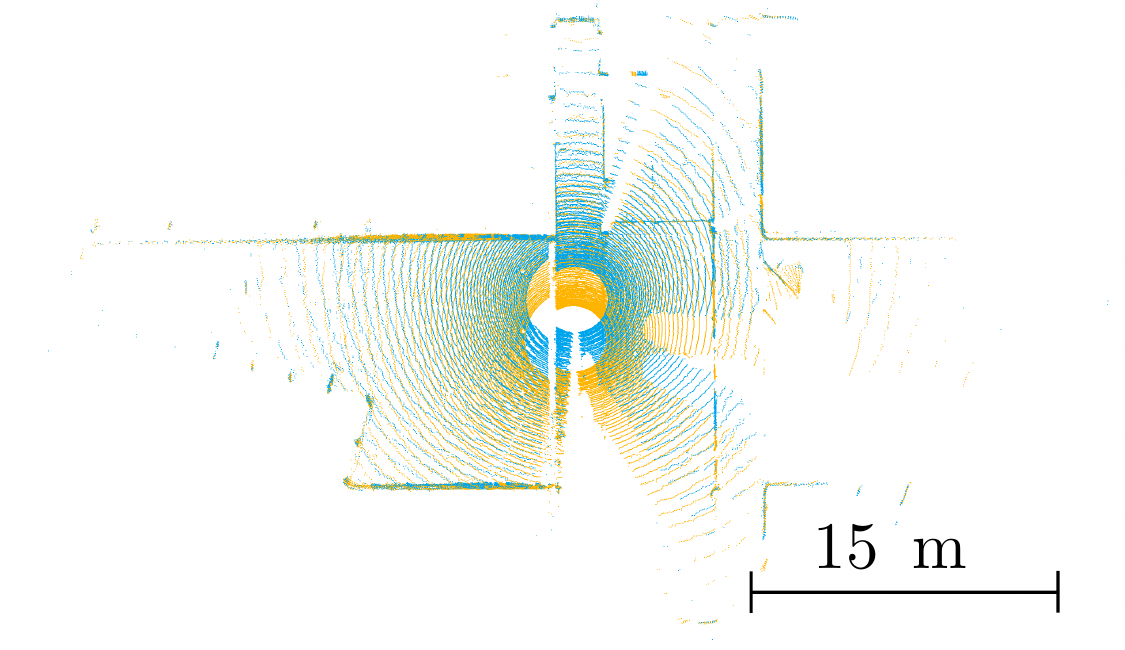}
        \caption*{(c) Ground truth}
    \end{subfigure} \\
    \end{tabular}
    \setlength{\arrayrulewidth}{0.4pt}  
    \arrayrulecolor{black} 
    \caption{Qualitative results on indoor point cloud registration (T-B): \ScanNetppi, \ScanNetppF, and {\TIERS} sequences. (a)~Input source (yellow) and target (cyan) point clouds before registration. (b)~Registration results obtained using our BUFFER-X, trained only on {\ThreeDMatch}. (c)~Ground truth alignment. Visualization demonstrates that BUFFER-X achieves accurate alignment, closely matching the ground truth.}
    \label{fig:viz_indoor}
\end{figure*}
\renewcommand{\qualitativewidth}{0.272}
\begin{figure*}[t!]
    \centering
    \setlength{\arrayrulewidth}{0.3pt}  
    \arrayrulecolor[gray]{0.7}  
    \begin{tabular}{c c c c}
    \multirow{2}{*}{\makebox[0pt][r]{\rotatebox{90}{\Oxford}\hspace{-0.1cm}}} &
    \begin{subfigure}[b]{\qualitativewidth\textwidth}
        \includegraphics[width=1.0\textwidth]{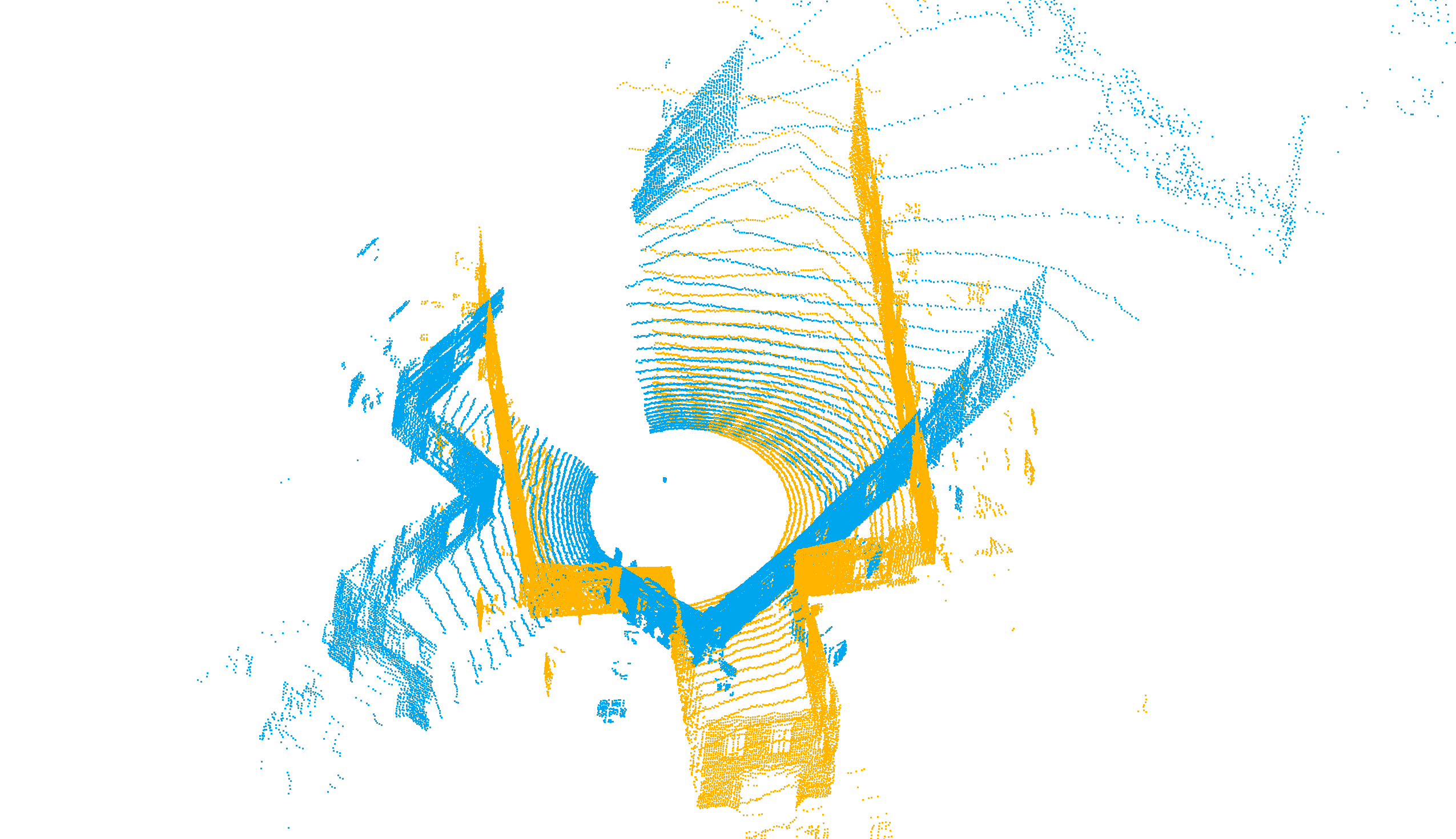}
    \end{subfigure} &
    \begin{subfigure}[b]{\qualitativewidth\textwidth}
        \includegraphics[width=1.0\textwidth]{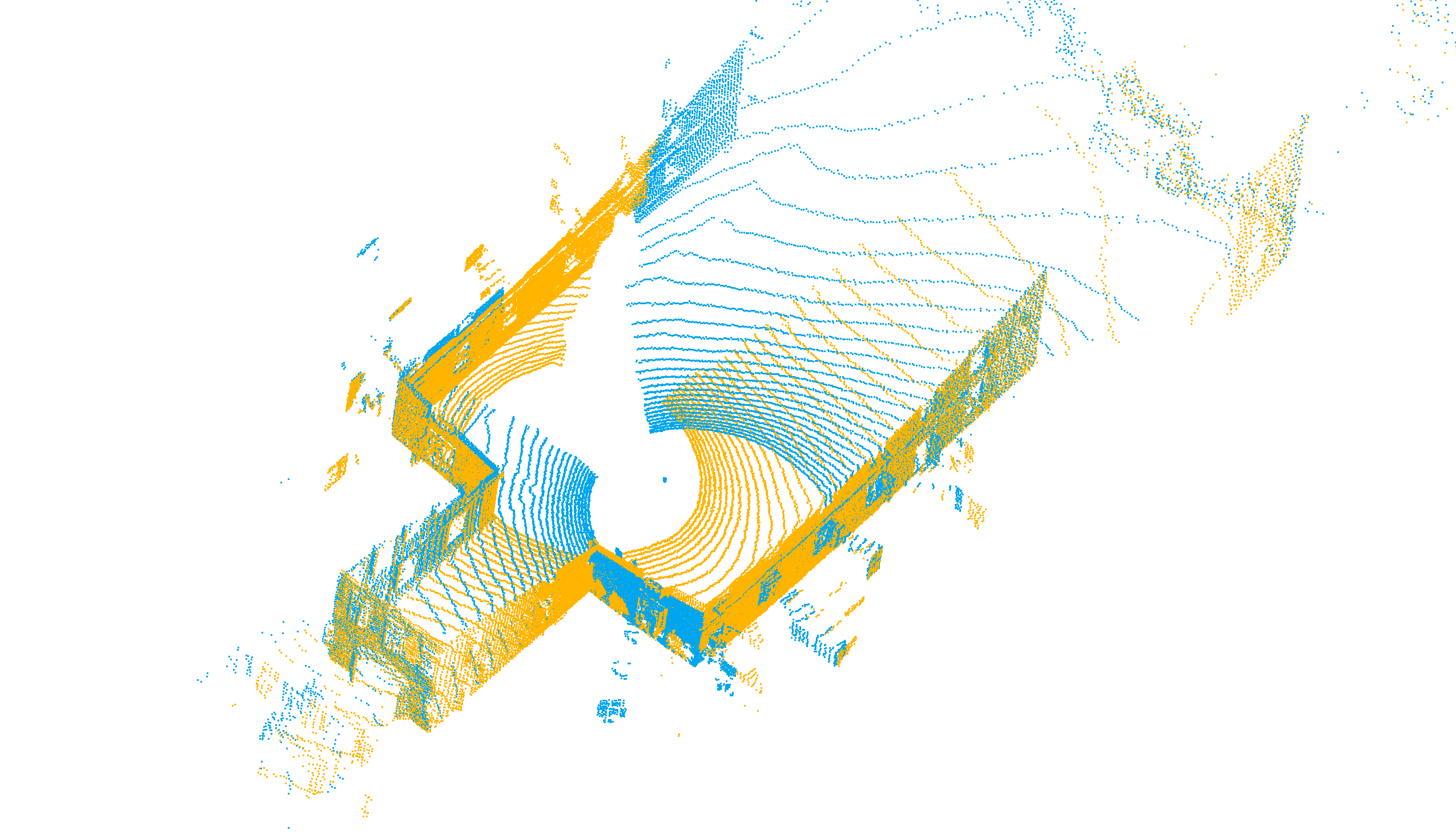}
    \end{subfigure} &
    \begin{subfigure}[b]{\qualitativewidth\textwidth}
        \includegraphics[width=1.0\textwidth]{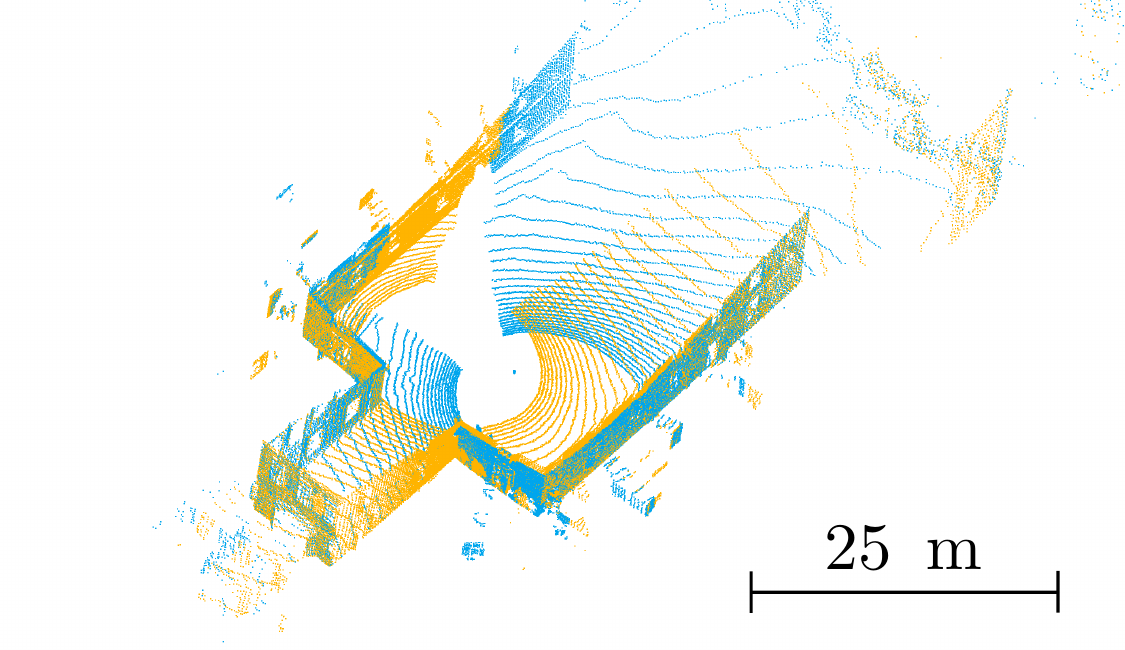}
    \end{subfigure} \\
    &
    \begin{subfigure}[b]{\qualitativewidth\textwidth}
        \includegraphics[width=1.0\textwidth]{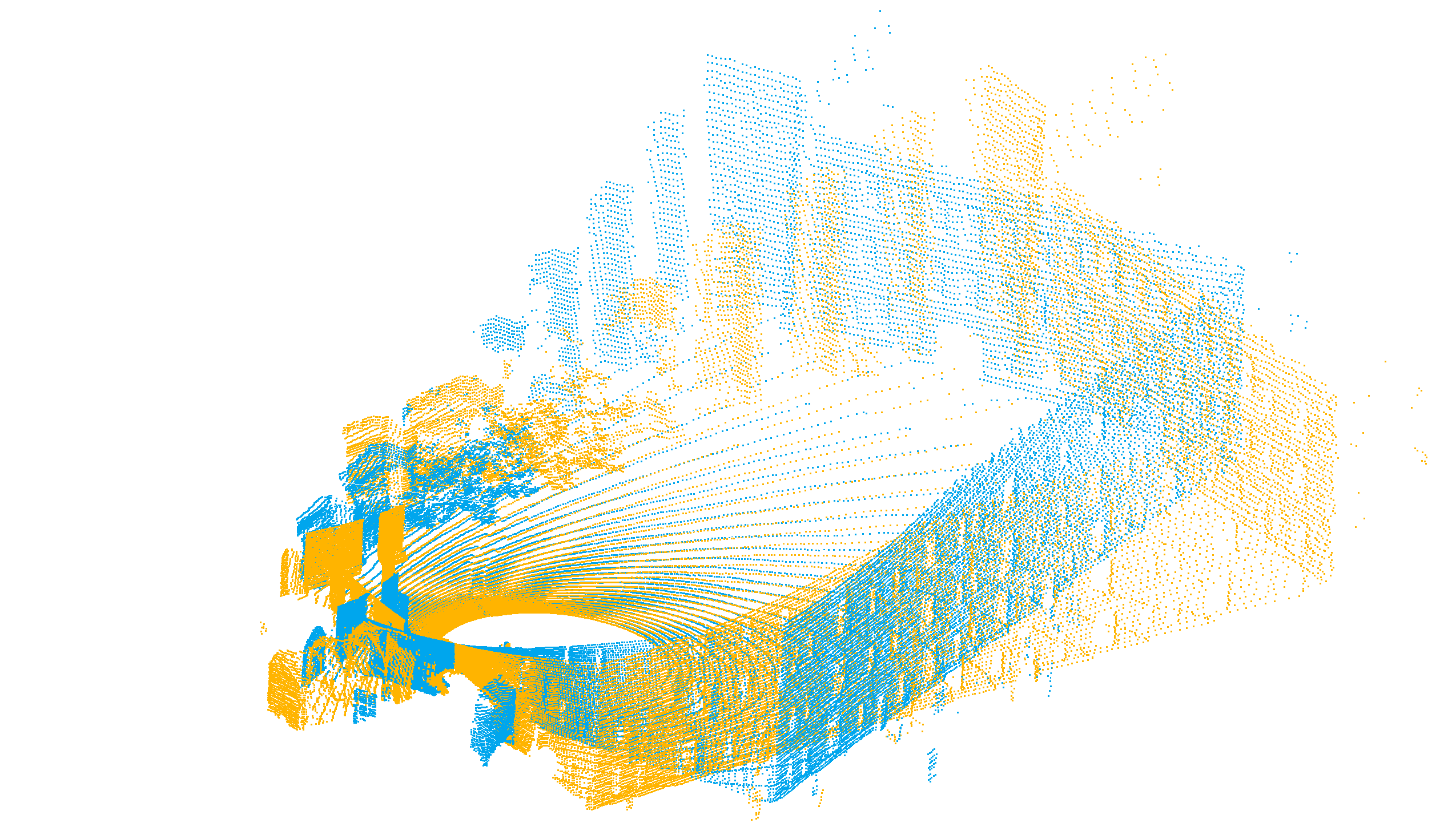}
    \end{subfigure} &
    \begin{subfigure}[b]{\qualitativewidth\textwidth}
        \includegraphics[width=1.0\textwidth]{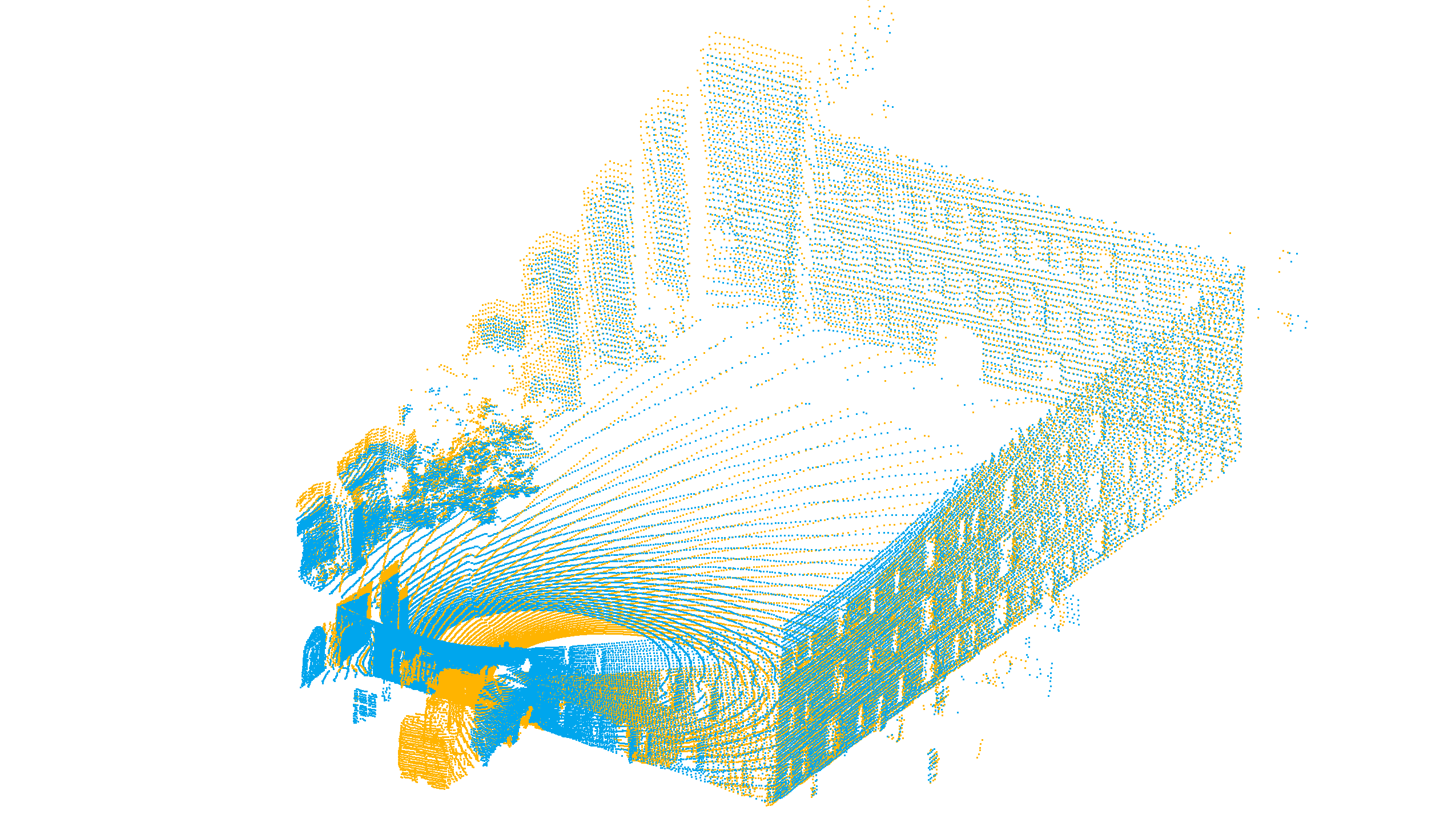}
    \end{subfigure} &
    \begin{subfigure}[b]{\qualitativewidth\textwidth}
        \includegraphics[width=1.0\textwidth]{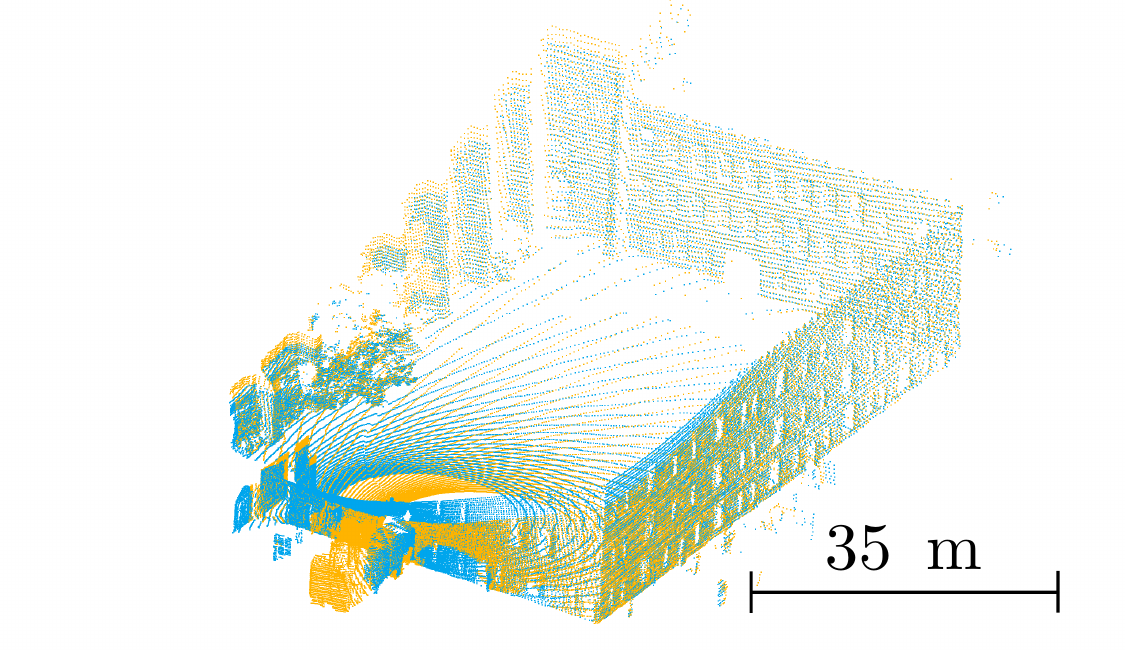}
    \end{subfigure} \\
    \\ \midrule
    \multirow{2}{*}{\makebox[0pt][r]{\rotatebox{90}{\KAIST}\hspace{-0.1cm}}} &
    \begin{subfigure}[b]{\qualitativewidth\textwidth}
        \includegraphics[width=1.0\textwidth]{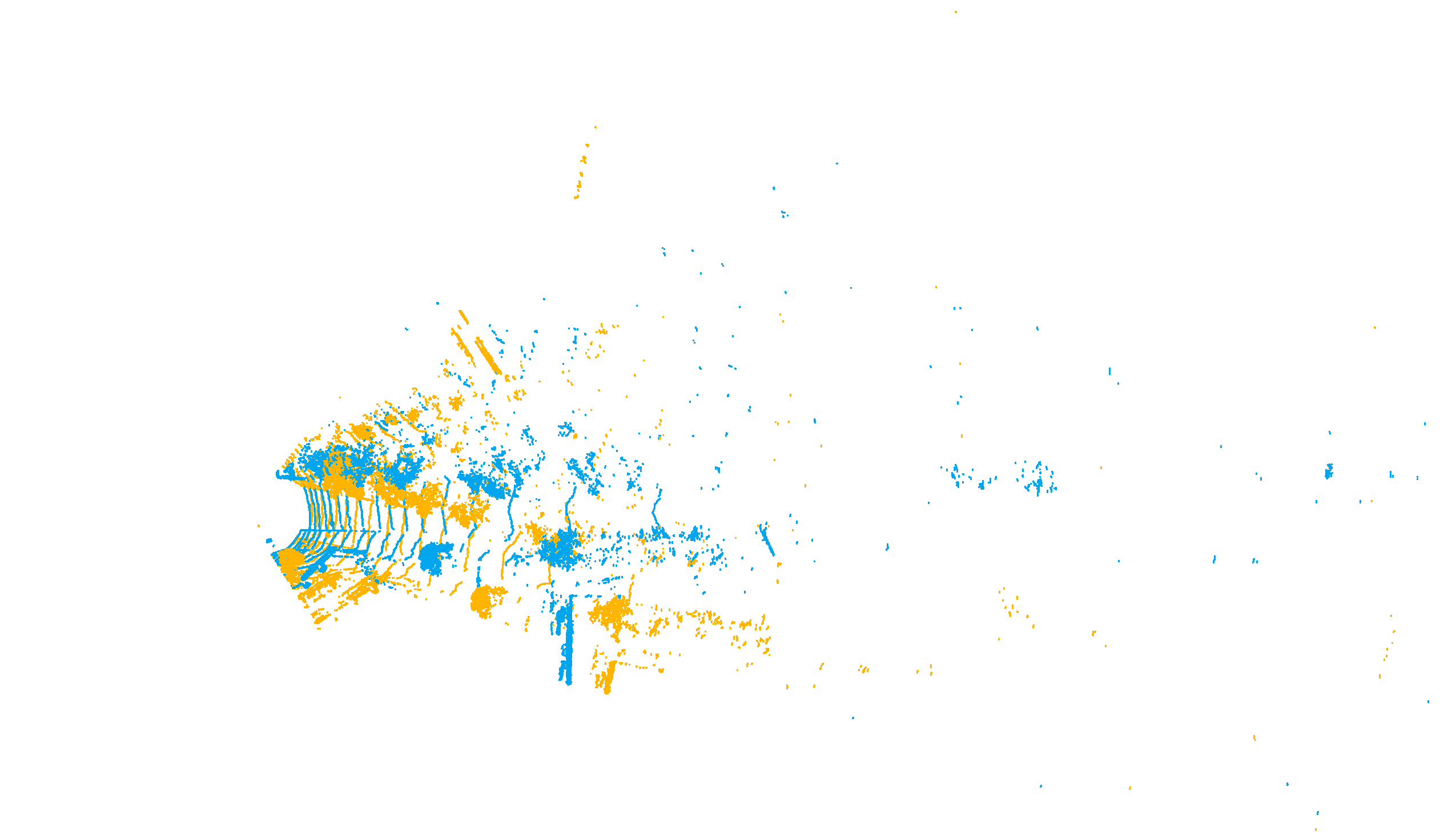}
    \end{subfigure} &
    \begin{subfigure}[b]{\qualitativewidth\textwidth}
        \includegraphics[width=1.0\textwidth]{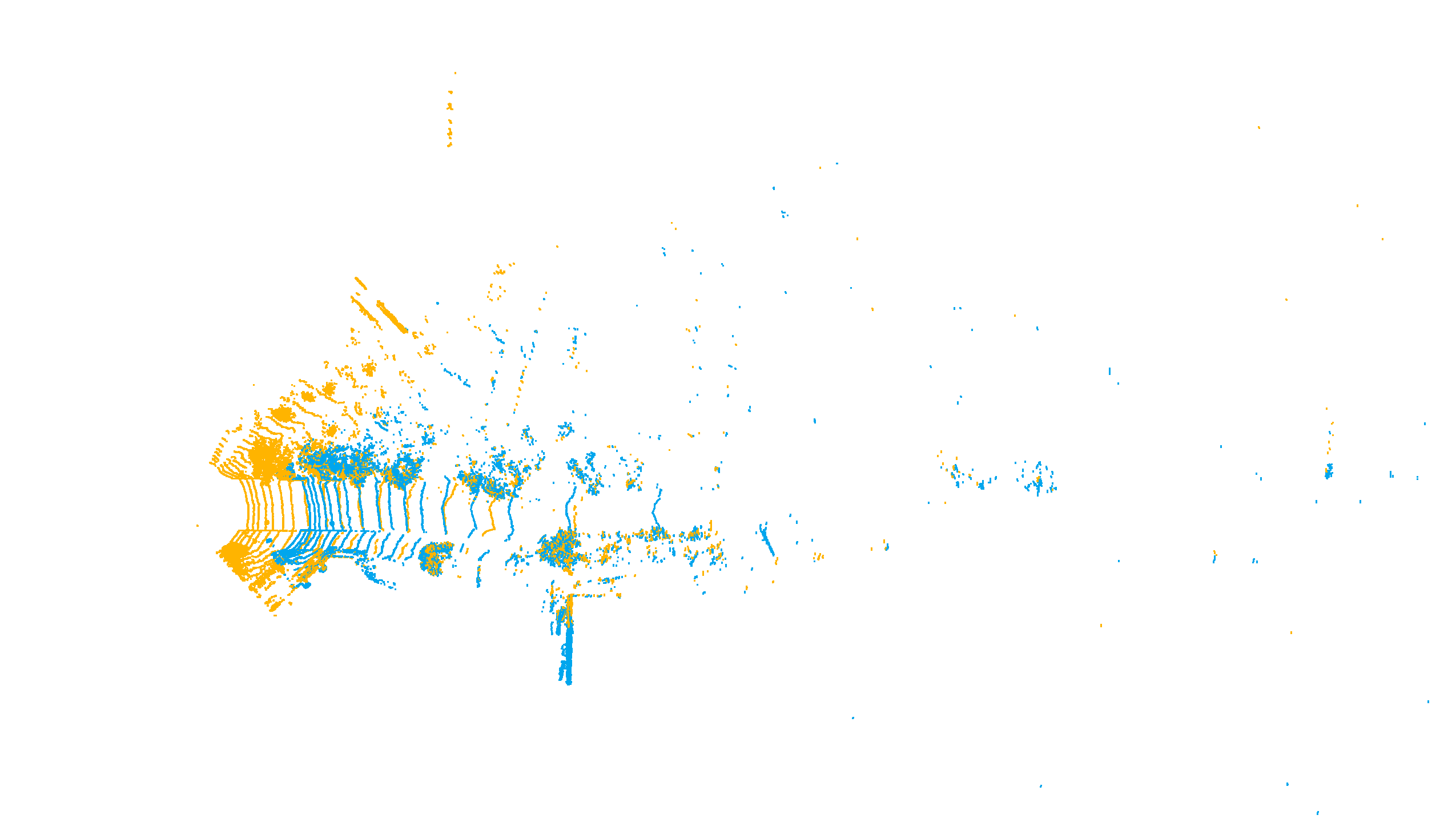}
    \end{subfigure} &
    \begin{subfigure}[b]{\qualitativewidth\textwidth}
        \includegraphics[width=1.0\textwidth]{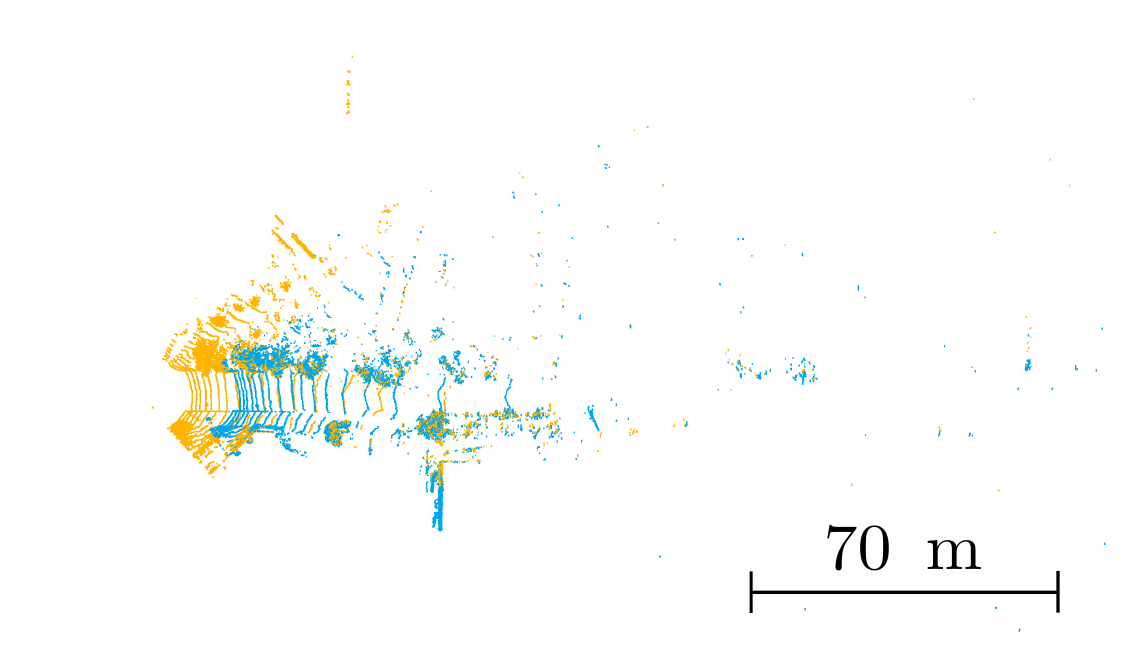}
    \end{subfigure} \\
    &
    \begin{subfigure}[b]{\qualitativewidth\textwidth}
        \includegraphics[width=1.0\textwidth]{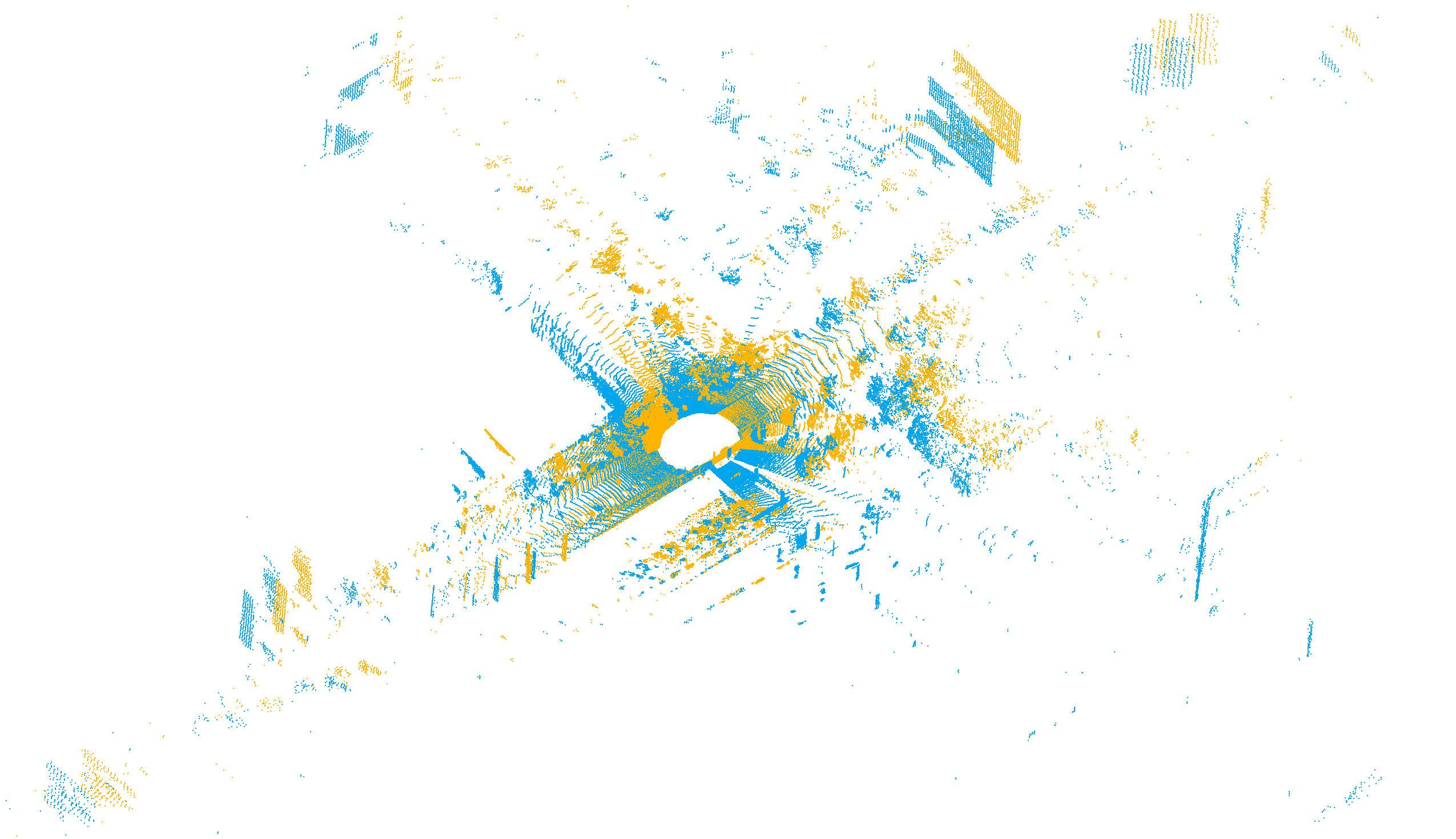}
    \end{subfigure} &
    \begin{subfigure}[b]{\qualitativewidth\textwidth}
        \includegraphics[width=1.0\textwidth]{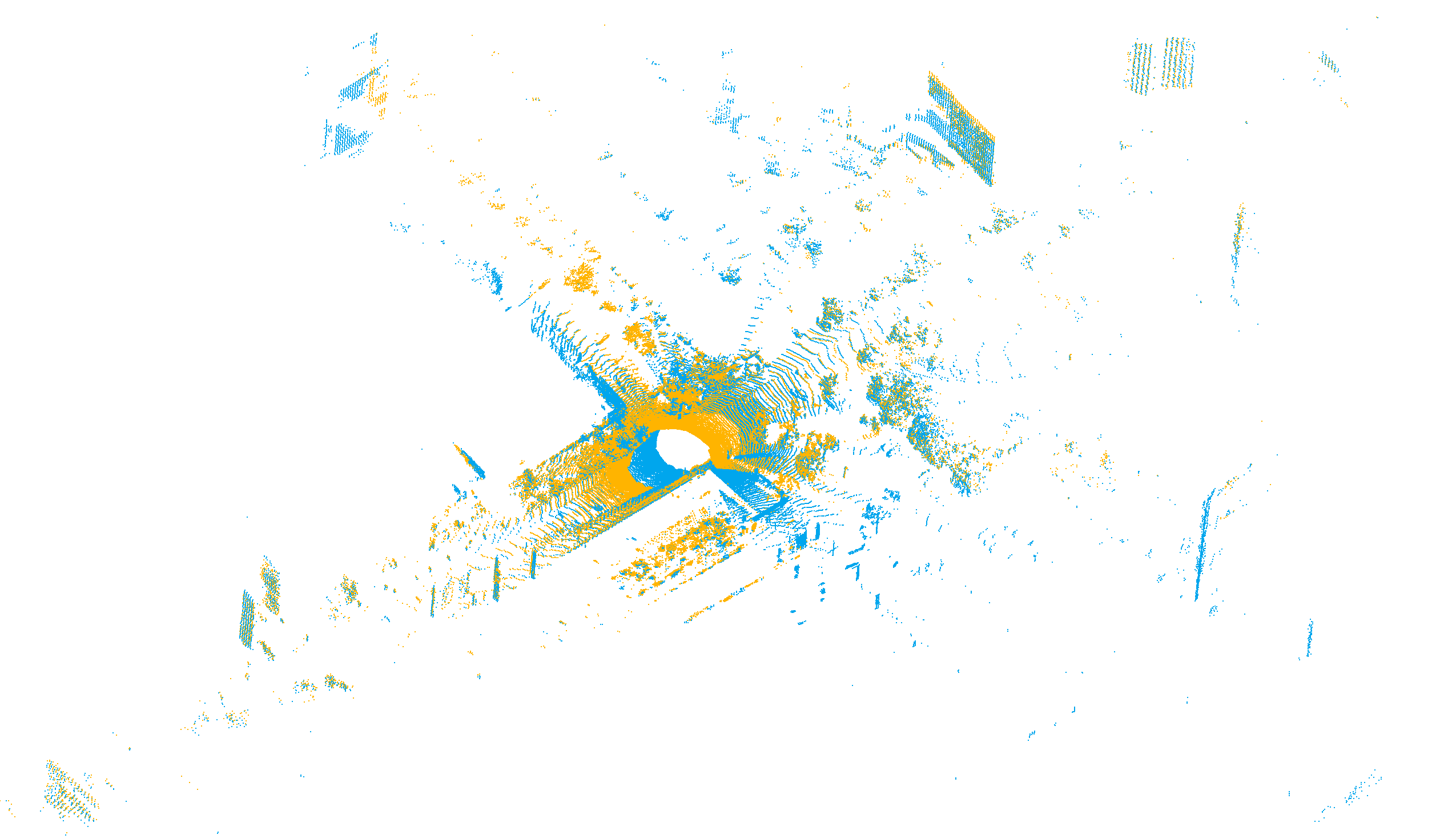}
    \end{subfigure} &
    \begin{subfigure}[b]{\qualitativewidth\textwidth}
        \includegraphics[width=1.0\textwidth]{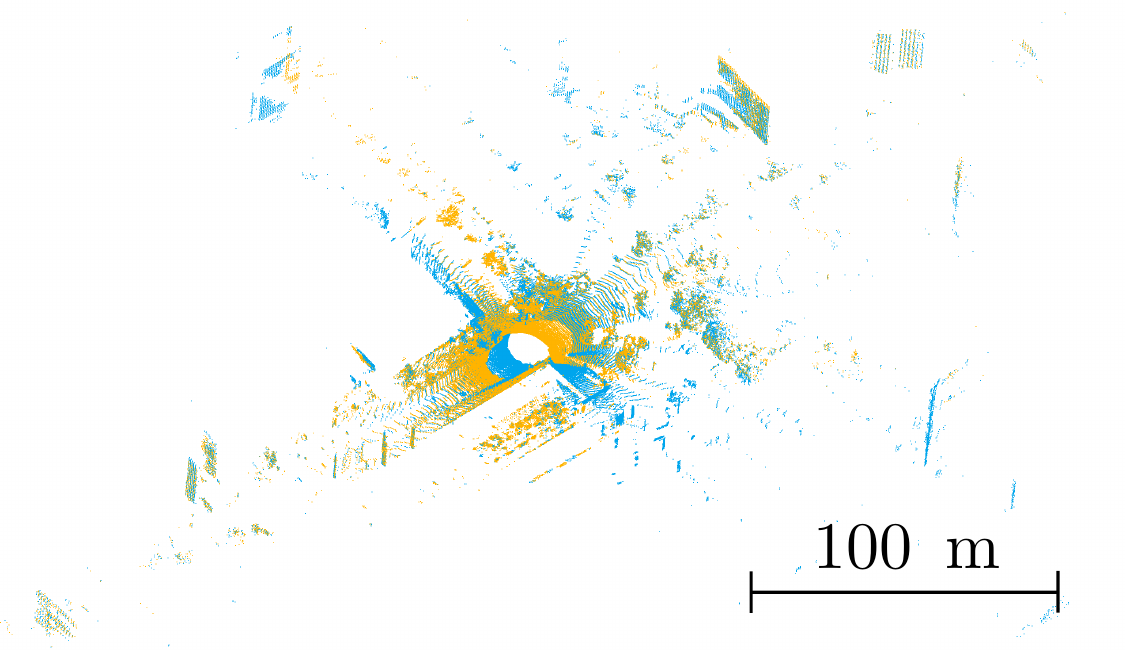}
    \end{subfigure} \\
    \\ \midrule \\ 
    \multirow{2}{*}{\makebox[0pt][r]{\rotatebox{90}{\MIT}\hspace{-0.1cm}}} &
    \begin{subfigure}[b]{\qualitativewidth\textwidth}
        \includegraphics[width=1.0\textwidth]{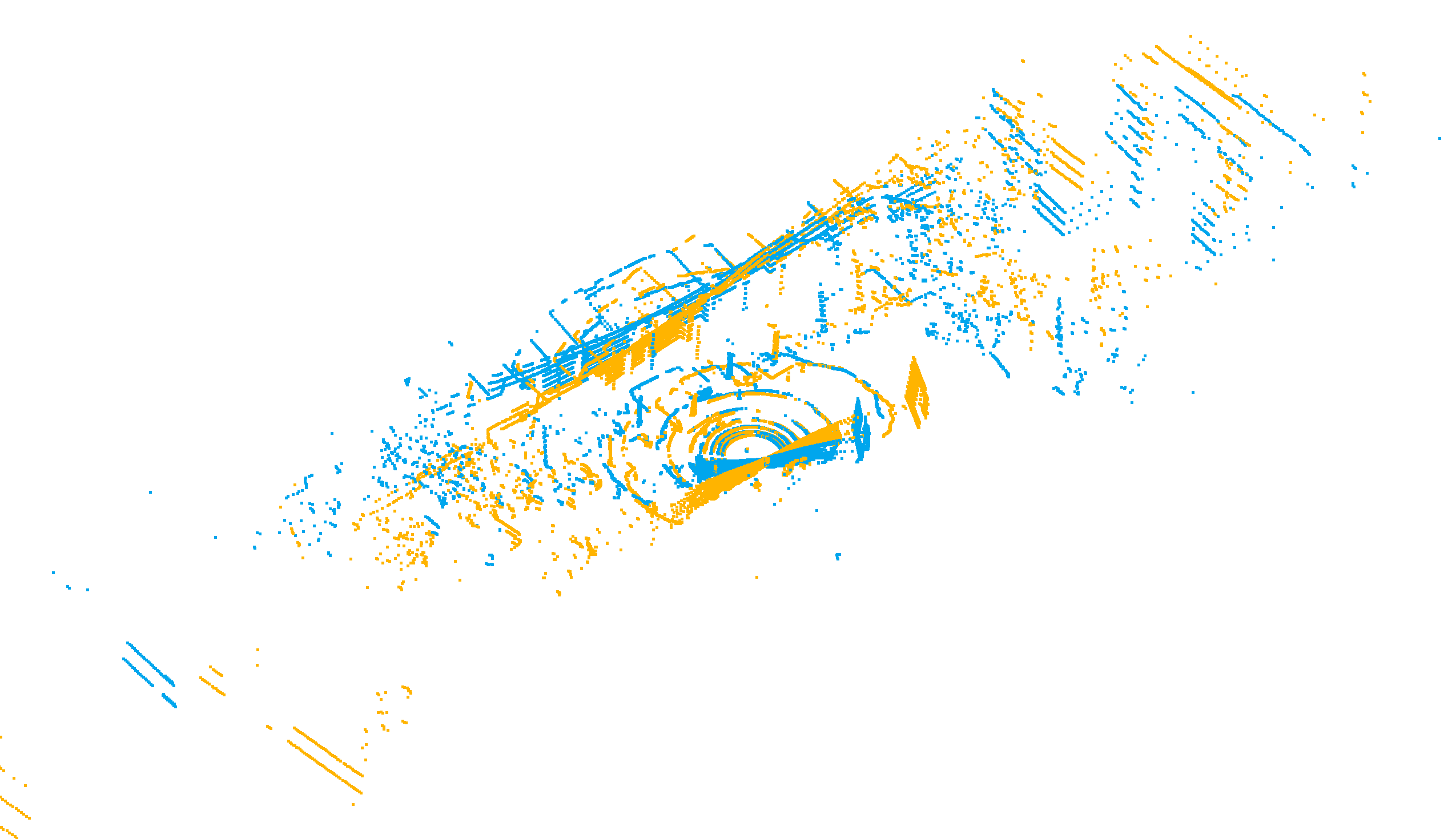}
    \end{subfigure} &
    \begin{subfigure}[b]{\qualitativewidth\textwidth}
        \includegraphics[width=1.0\textwidth]{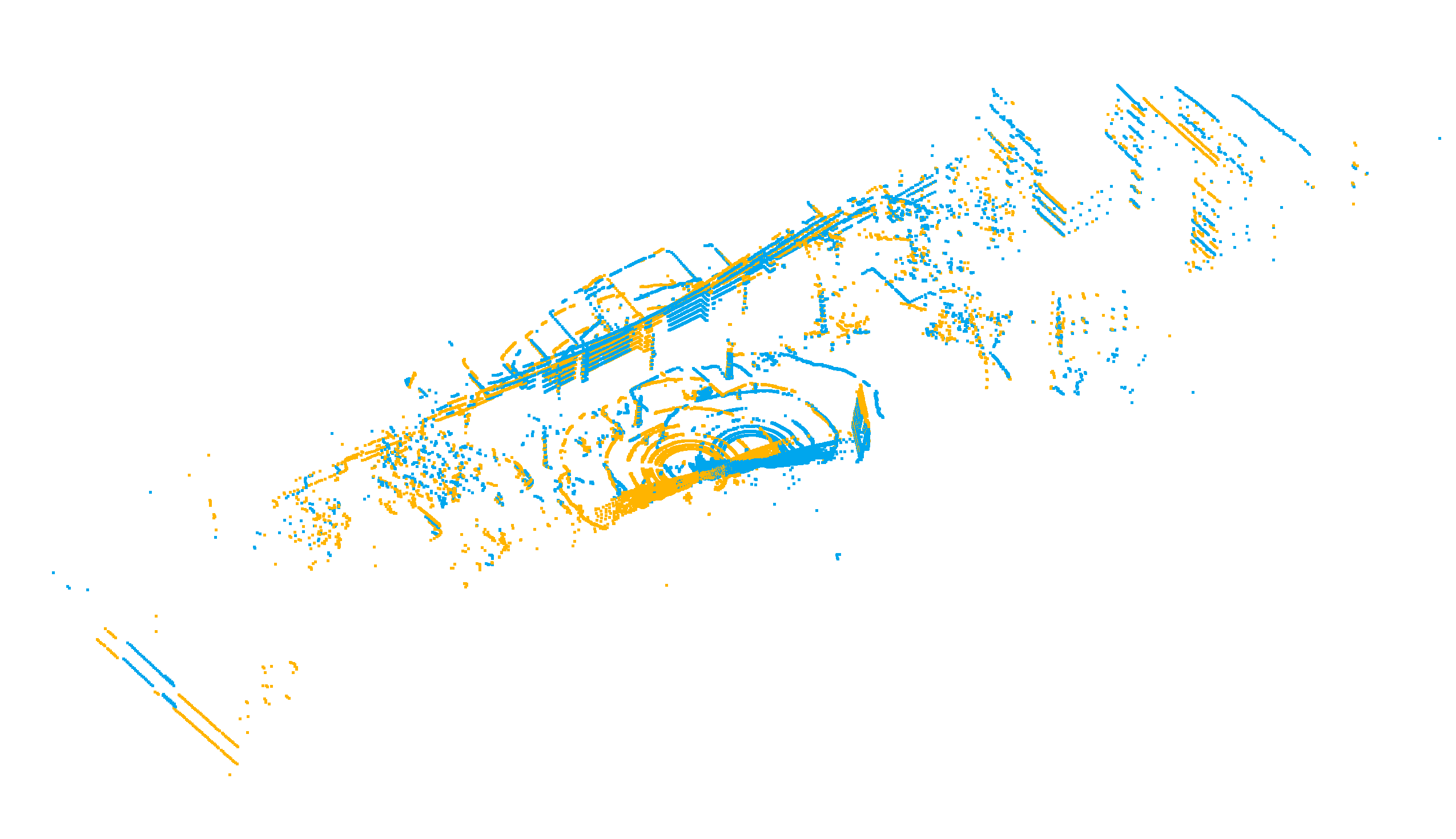}
    \end{subfigure} &
    \begin{subfigure}[b]{\qualitativewidth\textwidth}
        \includegraphics[width=1.0\textwidth]{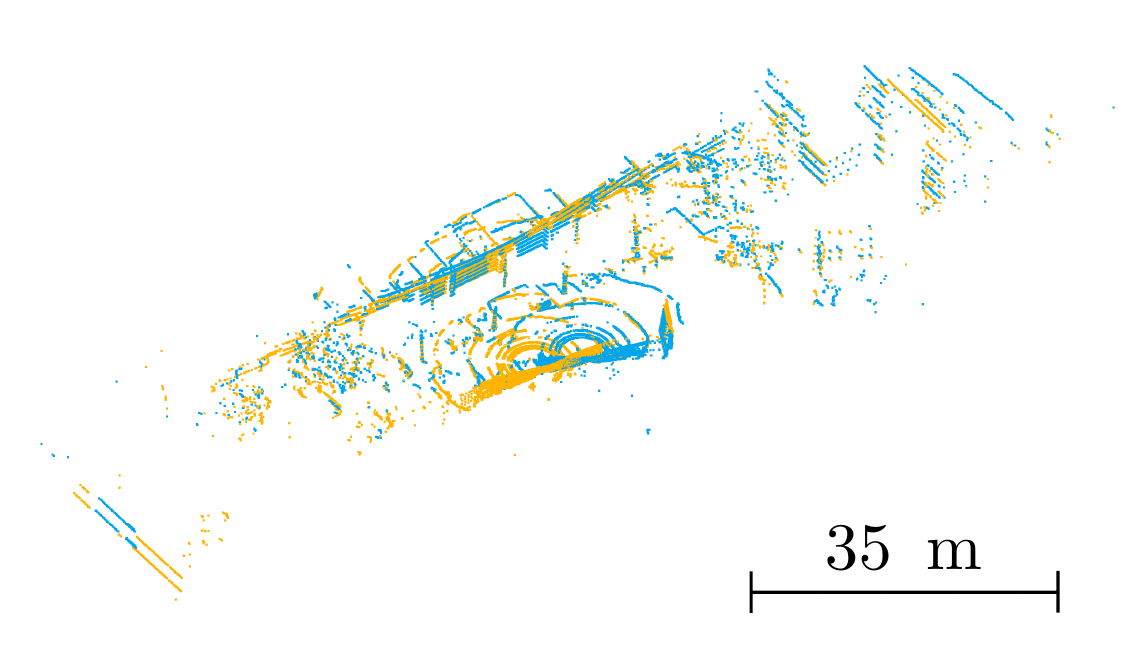}
    \end{subfigure} \\
    &
    \begin{subfigure}[b]{\qualitativewidth\textwidth}
        \includegraphics[width=1.0\textwidth]{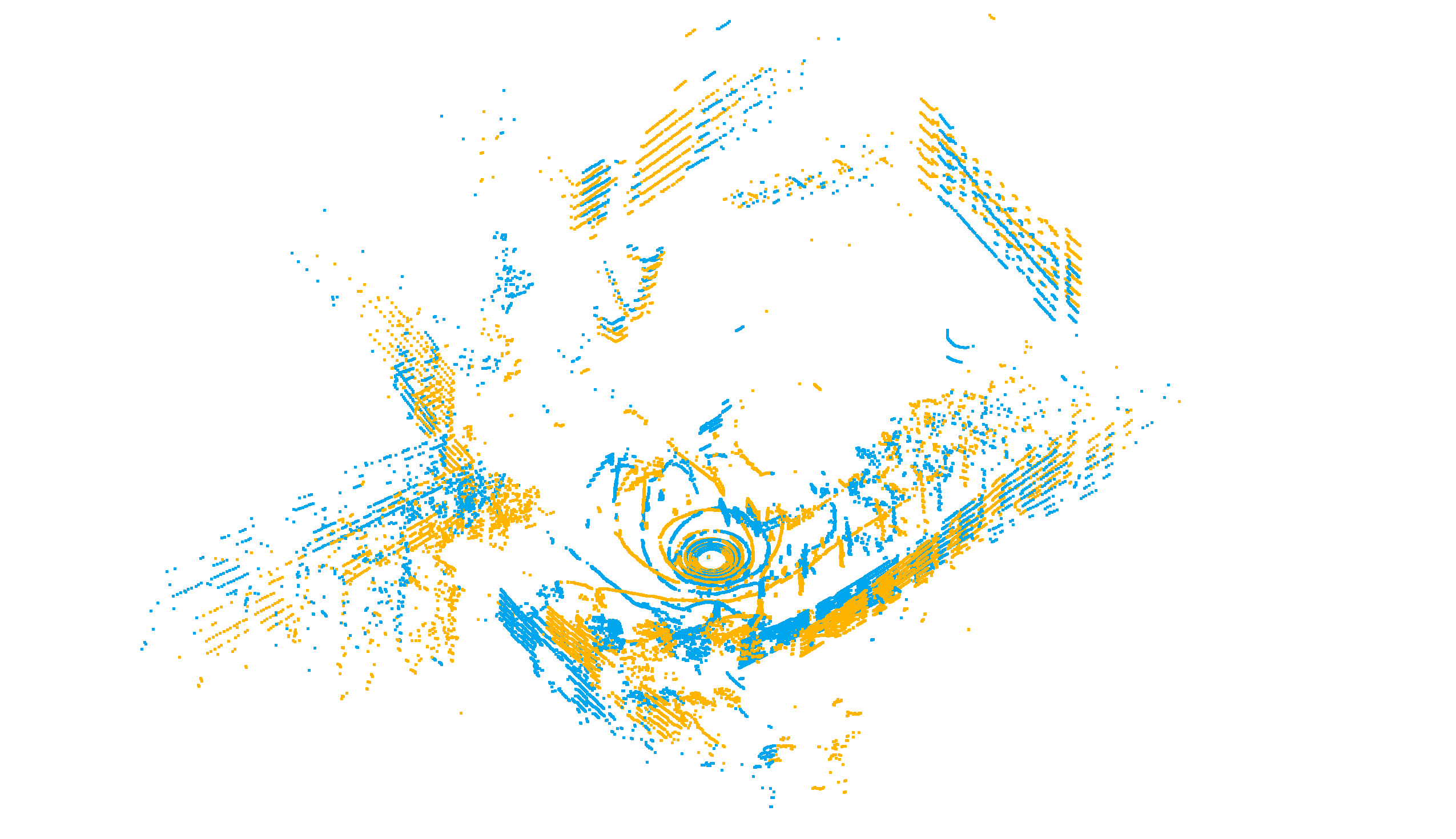}
        \caption*{(a) Source and target}
    \end{subfigure} &
    \begin{subfigure}[b]{\qualitativewidth\textwidth}
        \includegraphics[width=1.0\textwidth]{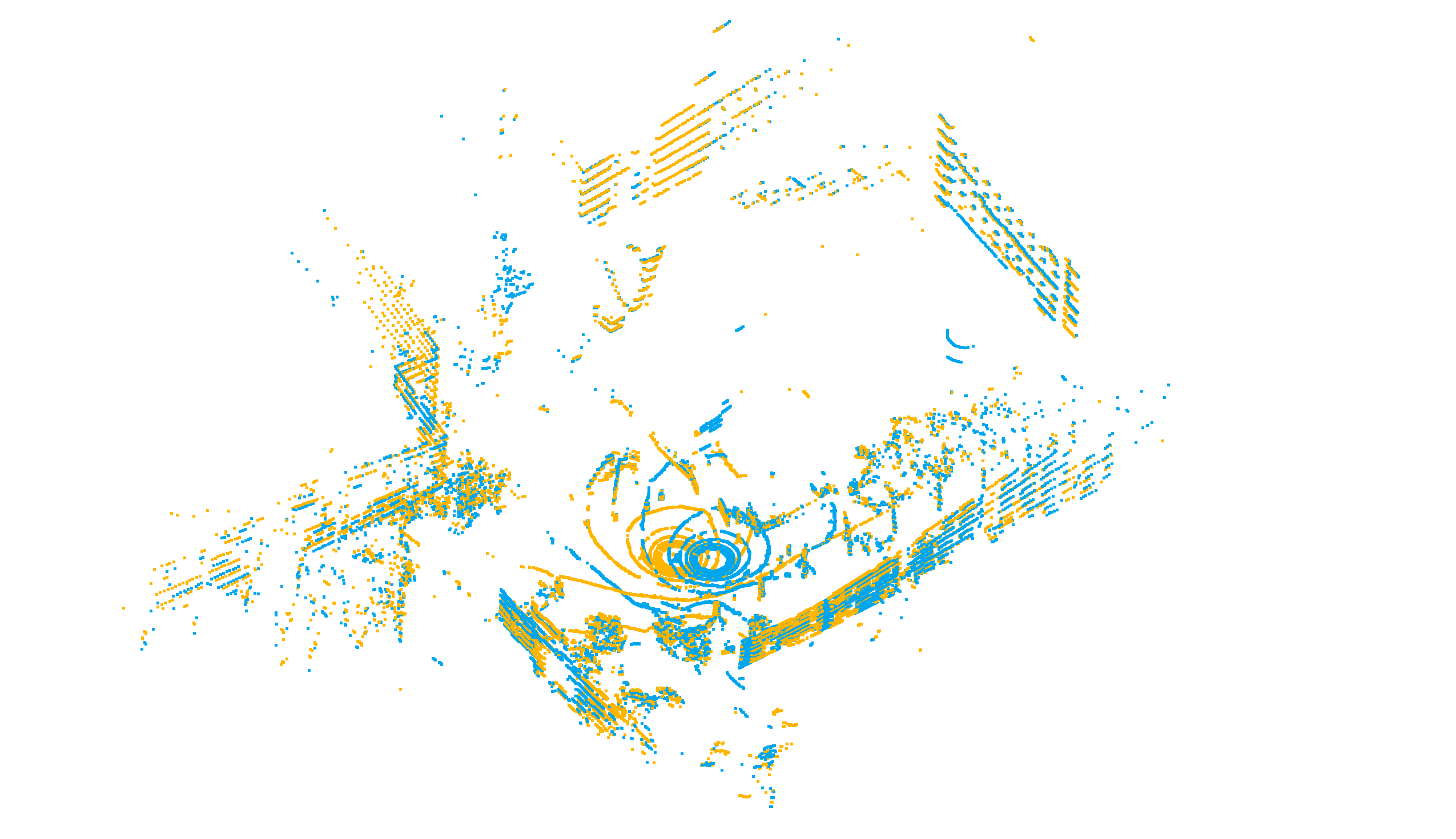}
        \caption*{(b) BUFFER-X (Ours)}
    \end{subfigure} &
    \begin{subfigure}[b]{\qualitativewidth\textwidth}
        \includegraphics[width=1.0\textwidth]{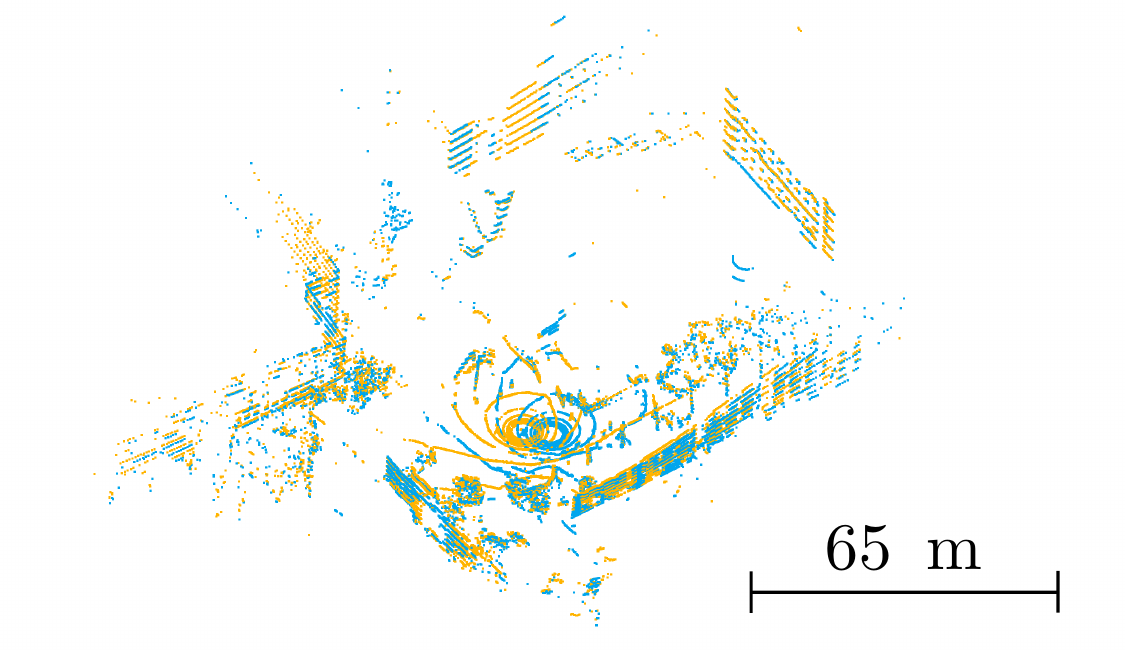}
        \caption*{(c) Ground truth}
    \end{subfigure} \\
\end{tabular}
    \setlength{\arrayrulewidth}{0.4pt}  
    \arrayrulecolor{black}  
    \caption{Qualitative results on outdoor point cloud registration: (T-B): \Oxford, \KAIST, and {\MIT} sequences. (a)~Input source (yellow) and target (blue) point clouds before registration. (b)~Registration results obtained using our BUFFER-X, trained only on {\ThreeDMatch}.  (c)~Ground truth alignment. Visualization demonstrates that BUFFER-X achieves accurate alignment, closely matching the ground truth.}
    \label{fig:viz_outdoor}
\end{figure*}

Therefore, these results further support the zero-shot registration capability of our BUFFER-X, regardless of the environment, sensor type, acquisition setup, or range.

\end{document}